\documentclass{article}
\usepackage[dvips]{graphicx}

\usepackage{a4wide}    
\usepackage{amssymb,amsmath,array}

\usepackage[utf8]{inputenc}
\usepackage[T1]{fontenc}
\usepackage{tabularx}

\usepackage{amsfonts}
\usepackage{bm}
\usepackage{dsfont}
\usepackage{adjustbox}
\usepackage{array}
\usepackage{booktabs}  
\usepackage{wrapfig}  
\usepackage{cancel}  
\usepackage{algorithm}
\usepackage{algpseudocode}
\usepackage{hyperref}
\hypersetup{
	colorlinks=true,
	linkcolor=blue,
	filecolor=magenta,      
	urlcolor=cyan,
}
\title{Alpha Entropy Search for New Information-based Bayesian Optimization}

\date{}

\author{
  Daniel Fern\'andez-S\'anchez\\
  Universidad Aut\'onoma de Madrid\\
  Francisco Tom\'as y Valiente 11\\
  28049, Madrid, Spain\\
  \texttt{daniel.fernandezs@uam.es}
  \and
  Eduardo C. Garrido-Merch\'an\\
  Universidad Pontificia Comillas\\
  Alberto Aguilera 23\\
  28015, Madrid, Spain\\
  \texttt{ecgarrido@comillas.edu}
  \and
  Daniel Hern\'andez-Lobato\\
  Universidad Aut\'onoma de Madrid\\
  Francisco Tom\'as y Valiente 11\\
  28049, Madrid, Spain\\
  \texttt{daniel.hernandez@uam.es}
}

\begin{document}

\maketitle

\begin{abstract}
Bayesian optimization (BO) methods based on information theory have obtained state-of-the-art results in several tasks.
These techniques heavily rely on the Kullback-Leibler (KL) divergence to compute the acquisition function.
In this work, we introduce a novel information-based class of acquisition functions for BO called Alpha Entropy Search (AES). 
AES is based on the alpha-divergence, that generalizes the KL divergence. Iteratively, AES selects the next evaluation point as the 
one whose associated target value has the highest level of the dependency with respect to the location and associated 
value of the global maximum of the optimization problem. Dependency is measured in terms of the alpha-divergence, as an
alternative to the KL divergence. Intuitively, this favors the evaluation of the objective 
function at the most informative points about the global maximum. 
The alpha-divergence has a free parameter $\alpha$, which determines the behavior of the divergence, trading-off evaluating 
differences between distributions at a single mode, and evaluating differences globally. Therefore, different values 
of $\alpha$ result in different acquisition functions. AES acquisition lacks a closed-form expression. However, 
we propose an efficient and accurate approximation using a truncated Gaussian distribution. In practice, the 
value of $\alpha$ can be chosen by the practitioner, but here we suggest to use a combination of acquisition 
functions obtained by simultaneously considering a range of values of $\alpha$. We provide an implementation of 
AES in BOTorch and we evaluate its performance in both synthetic, benchmark and real-world experiments involving 
the tuning of the hyper-parameters of a deep neural network. These experiments show that the performance of AES 
is competitive with respect to other information-based acquisition functions such as Joint Entropy Search, Max-Value
Entropy Search or Predictive Entropy Search.
\end{abstract}

\section{Introduction}

Bayesian optimization (BO) includes a set of methods that have been successfully used for the optimization of black-box functions, and most concretely for the problem of tuning the hyper-parameters of machine learning models \cite{snoek2012practical}. In particular, a black-box function $f(\cdot)$ is characterized by having an unknown analytical expression and being costly to evaluate in computational or economical terms. Besides this, we also consider that the evaluation of $f(\cdot)$  may be corrupted by noise. Formally, the optimization scenario we consider can be defined as trying to find:
\begin{align}
\mathbf{x}^\star = \arg\max_{\mathbf{x} \in \mathcal{X}} f(\mathbf{x})\,,
\end{align}
where $f(\cdot)$ is the objective, and $\mathbf{x}^\star$ is the global optima in the considered bounded input space $\mathcal{X} \subset \mathds{R}^d$. We also 
denote $y^\star$ as the associated optimal objective value. Namely, $y^\star = f(\mathbf{x}^\star)$. 
Importantly, we also assume that the evaluation of $f(\cdot)$ may be contaminated by Gaussian random noise. 
That is, instead of observing directly $f(\mathbf{x})$, we observe $y = f(\mathbf{x}) + \epsilon$, where $\epsilon \sim \mathcal{N}(0, \sigma^2)$. 

Since the objective is assumed to be very expensive to evaluate, we would like to use as few evaluations as possible to estimate
$\mathbf{x}^\star$. BO methods are very successful in this task \cite{brochu2010tutorial,shahriari2015taking,garnett2023bayesian}. To tackle this scenario, given a small number of initial evaluations, 
they focus on modeling the black-box function $f(\cdot)$ with a probabilistic surrogate model in the input space $\mathcal{X}$. This model is 
typically a Gaussian Process (GP),  which outputs a predictive distribution for $f(\cdot)$, capturing the potential values 
of the objective in regions of the input space that have not been explored so far \cite{williams2006gaussian}. Using this model, BO methods guide the search for the optimum and make intelligent decisions
about what point should be evaluated next at each iteration. For this, they use an acquisition function $a(\mathbf{x})$ that measures
the expected utility of performing an evaluation at $\mathbf{x}$ with the goal of solving the optimization problem \cite{shahriari2015taking}. 
The next evaluation is simply the maximizer of the acquisition function $a(\mathbf{x})$.
Importantly, the probabilistic model and the acquisition function are very cheap to evaluate and maximize, respectively, since they
do not imply evaluating the actual objective. Therefore, the over-head that the process described introduces can be considered negligible.
After enough evaluations of the objective are performed, the best observation is returned as the global optimizer of $f(\cdot)$, in a noiseless setting. 
In a noisy setting, a similar recommendation is made, but the probabilistic model is used first to remove the noise from the evaluations performed.

There is a plethora of different optimization scenarios, which are particular cases of the one described before, where BO methods can 
be applied in practice in science or engineering. For example, BO has successfully been used in energy for robust ocean 
wave features prediction \cite{cornejo2018bayesian}; in chemistry, for the discovery of energy storage molecular materials 
\cite{agarwal2021discovery}; in robotics, for a better design of the wing shape of an unmanned aerial vehicle \cite{martinez2017bayesian}; in 
finance, for environmental, social and governance sustainable portfolio optimization \cite{garrido2023bayesian}; or even as 
a way to suggest a better dessert, optimizing chocolate chip cookies \cite{solnik2017bayesian}. Importantly, BO also gives superior 
results to other optimization methods that do not rely on a model to guide the search for the optimum, such 
as meta-heuristics or genetic algorithms \cite{lobato16_accelerator}. 

A critical and important part of any BO method is the acquisition function. In this regard, in the BO literature, 
acquisition functions based on information theory have delivered state-of-the-art results by efficiently guiding 
the search for $\mathbf{x}^\star$ \cite{hernandez2014predictive,villemonteix2009informational,hennig2012entropy}. 
These strategies choose the next evaluation point as the one that results in the highest 
expected decrease in the entropy of the global optimum $\mathbf{x}^\star$. The global optimum
can be regarded as a random variable since there is uncertainty about the potential values of the 
objective $f(\cdot)$. These potential values are modeled by the GP. Among information-based strategies, the recently proposed 
Joint Entropy Search (JES), obtains state-of-the-art results \cite{hvarfner2022joint,tu2022joint}. 
The JES acquisition function $a(\mathbf{x})$ measures the expected reduction in the differential entropy 
of both $\mathbf{x}^\star$ and $y^\star$, \emph{i.e.}, $\{\mathbf{x}^\star,y^\star\}$ after observing $y$ at 
$\mathbf{x}$. Importantly, this expected reduction in the differential entropy
can be shown to be equal to the mutual information between $\{\mathbf{x}^\star,y^\star\}$ and $y$, where the distribution of 
$y$, \emph{i.e.}, the noisy objective value at the candidate point $\mathbf{x}$, is also given by the GP \cite{hvarfner2022joint}.
The mutual information is just the  Kullback-Leibler (KL) divergence between $p(\{\mathbf{x}^\star,y^\star\},y)$
and $p(\{\mathbf{x}^\star, y^\star\})p(y)$, \emph{i.e.}, the same distribution, but where independence
between $\{\mathbf{x}^\star, y^\star\}$ and $y$ is assumed. The KL divergence is always non-negative and equal to
zero only when the two distributions are equal. Thus, JES simply chooses the next point $\mathbf{x}$ to evaluate as the 
one at which there is a higher level of dependency between $\{\mathbf{x}^\star, y^\star\}$ and $y$, as measured by the KL 
divergence. Observing $y$ at such an $\mathbf{x}$ will provide more knowledge about the
potential values of $\{\mathbf{x}^\star, y^\star\}$, due to the strong dependencies, and is expected to help the most to solve 
the optimization problem.

As an alternative to the KL divergence, we consider in this paper other methods to estimate the level of dependency
between $\{\mathbf{x}^\star, y^\star\}$ and $y$. More precisely, we consider a generalization of the KL divergence
to measure how similar $p(\{\mathbf{x}^\star,y^\star\},y)$ is to $p(\{\mathbf{x}^\star, y^\star\})p(y)$,
the $\alpha$-divergence \cite{amari1985}. The $\alpha$-divergence includes a parameter $\alpha$
that has an effect in the behavior of the divergence, trading-off evaluating differences between each 
distribution at a single mode, and evaluating differences globally \cite{minka2005divergence}.
We denote the resulting acquisition function as Alpha Entropy Search (AES). A difficulty is, however, that the closed-form expression of the 
$\alpha$-divergence is intractable. To overcome this, we describe here a simple and efficient method to approximate its 
value and hence the corresponding acquisition function. Of course, changing $\alpha$ results in different acquisition functions.
Notwithstanding, empirically, we did not observe a value for $\alpha$ that gives over-all good results. In consequence, we suggest to 
consider simultaneously a range of values for $\alpha$, resulting in a weighted combination of acquisition functions.
We have evaluated this combination across several experiments, including synthetic, benchmark, and real-world problems
related to the tuning of the hyper-parameters of deep neural networks. These results show that, in general, the proposed
method based on the $\alpha$-divergence, AES,  gives similar or better results than other information-based BO acquisition functions.

The organization of the paper is as follows: first, we include a section where we give fundamental details 
about information-based BO.  Next, another section explains the analytical details and 
the methodology of our proposed approach, the AES acquisition function. We continue with a section about related work,
where we emphasize the similarities and differences between our proposed method and other related methods that have 
been published in the BO literature. Afterward, we include a section where we give empirical evidence of the performance of AES
with respect to other information-based acquisition functions in synthetic, benchmark and real-world experiments. Finally, we end the manuscript 
with a section summarizing the conclusions. 

\section{Information-based Bayesian Optimization}

For clarity, we begin this section by illustrating the fundamentals of information-based BO, to further introduce the proposed acquisition 
function, AES, in the next section.

\begin{figure}[htb!]
  \centering
  \includegraphics[width=0.76\textwidth]{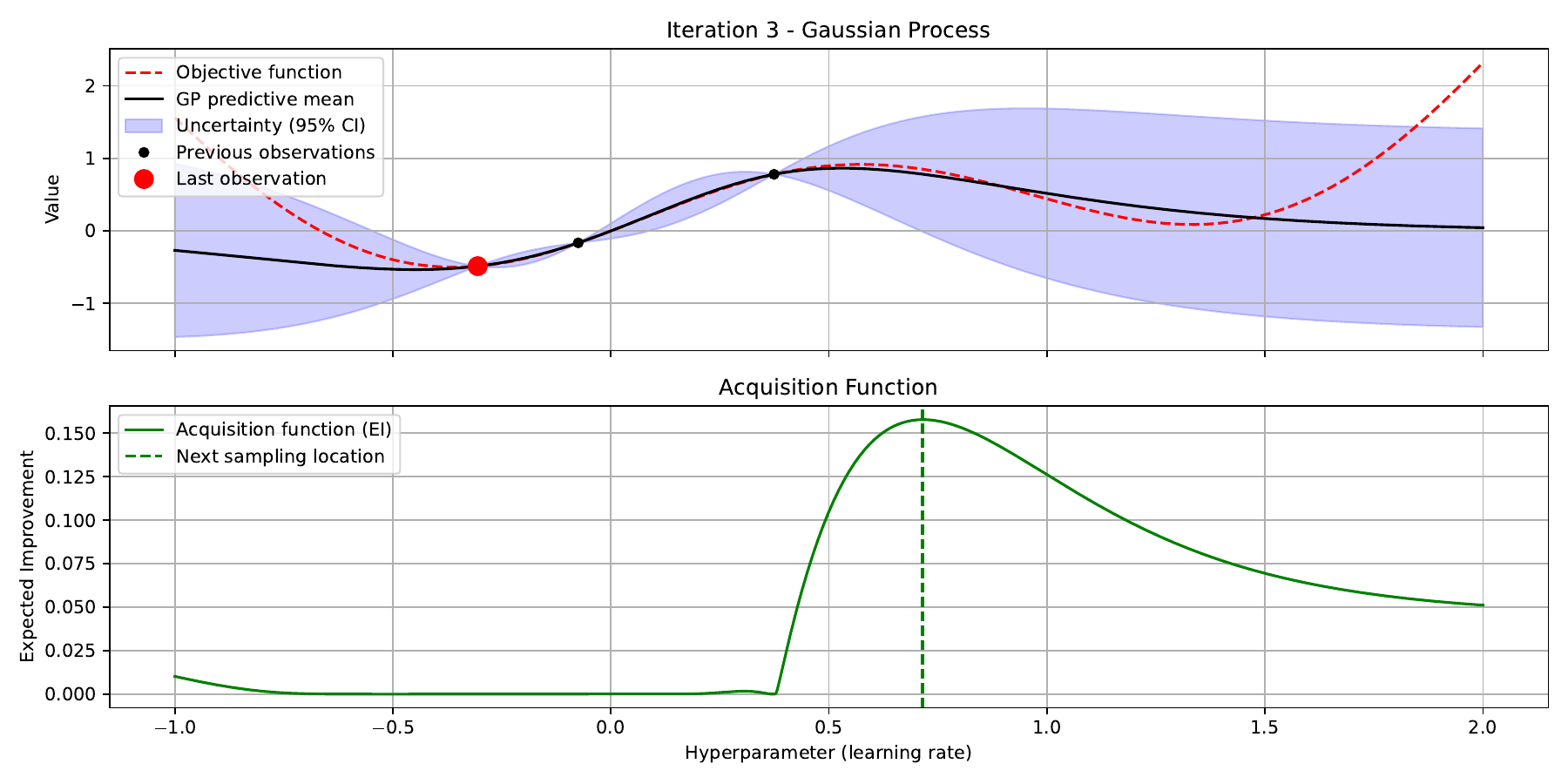}\par
  \includegraphics[width=0.76\textwidth]{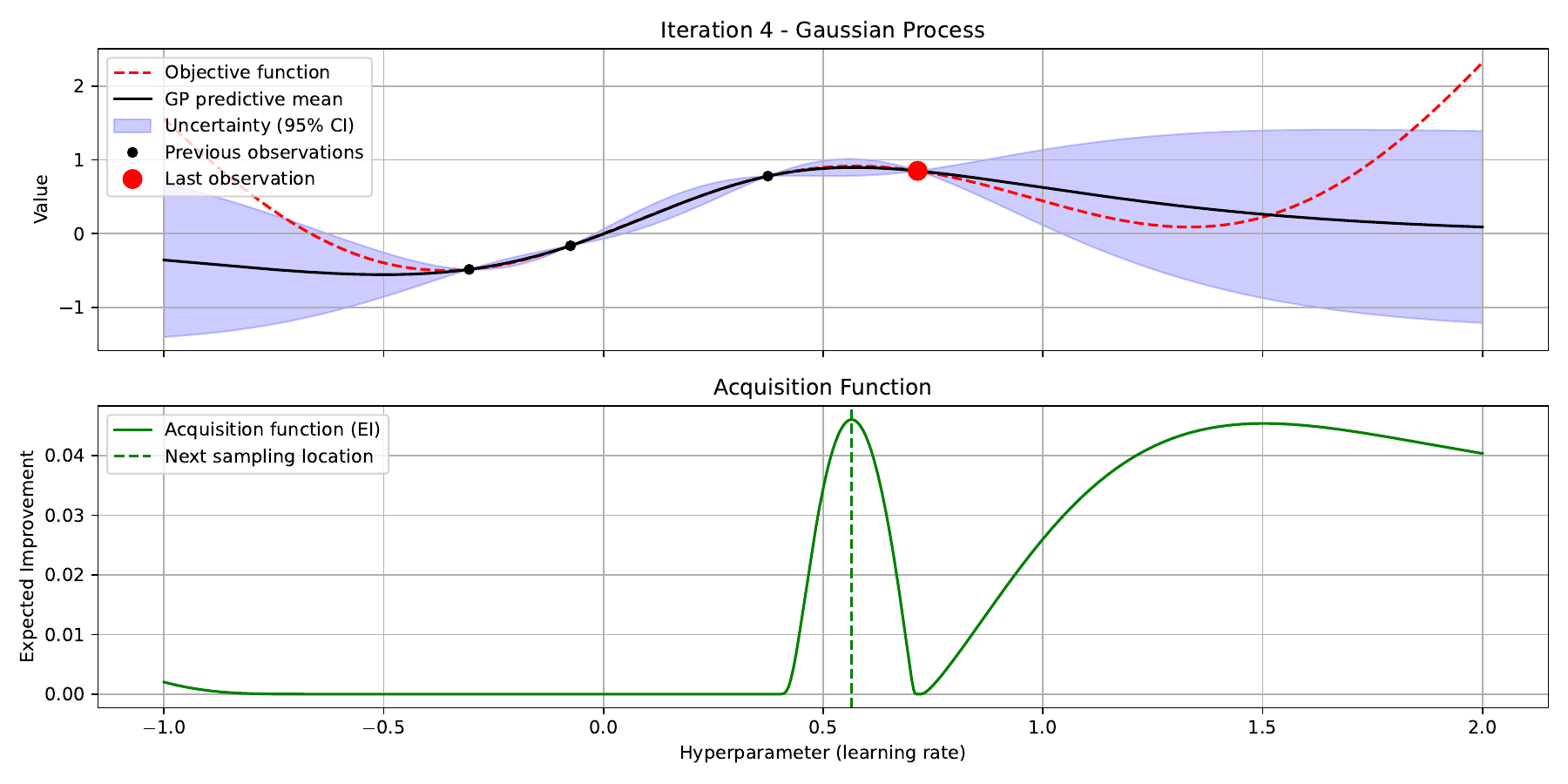}\par 
  \includegraphics[width=0.76\textwidth]{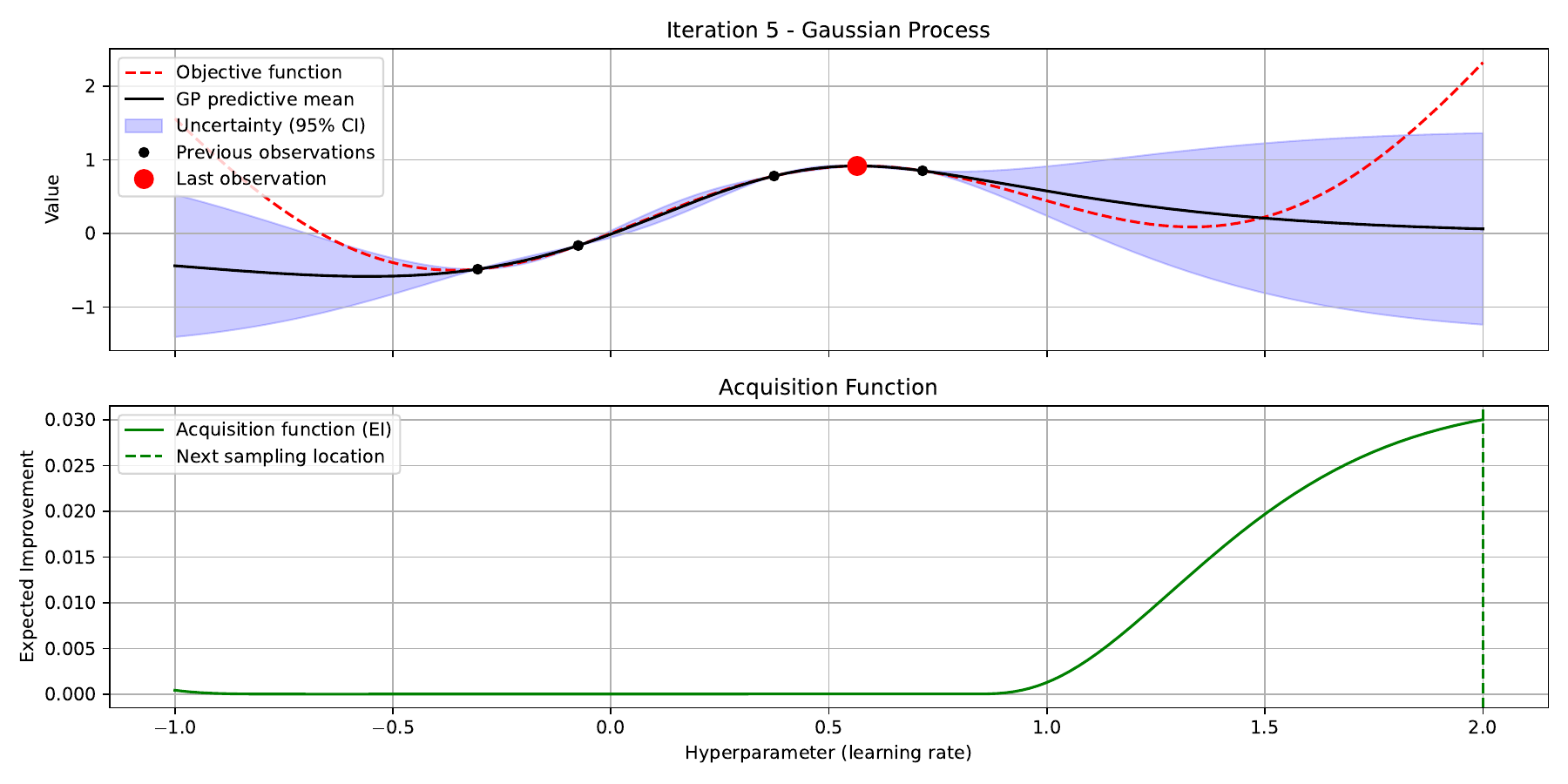}
	\caption{GP fit of the objective function (top of images) and the associated acquisition function (Expected Improvement \cite{garnett2023bayesian}) 
	built using the predictive distribution of the GP (down). Black points are observations are the red point is the last evaluation performed, given by the maximizer of 
	the acquisition function in the previous iteration. We can see the BO process as iterations are carried out (from t=3 to t=5) and how the GP and 
	the associated acquisition function guides the search for the optimum. }
  \label{fig:gp_af}
\end{figure}

We begin with a short description of the vanilla BO algorithm. As we have briefly described in the introduction, BO is a class of methods that 
optimize black-box functions \cite{shahriari2015taking}. In particular, the algorithm receives as an input an initial dataset 
$\mathcal{D}_0 = {(\mathbf{x}_i, y_i)}_{i=1}^{n_0}$, where each $\mathbf{x}_i$ represents the initial input 
space points in $\mathcal{X}$ and $y_i$ denotes the associated noisy observation of the objective value at $\mathbf{x}_i$. 
This dataset is chosen at random from $\mathcal{X}$.
Then, a probabilistic surrogate model, such as a Gaussian process (GP) fits the initial points in $\mathcal{D}_0$ to model the underlying objective 
function $f(\cdot)$ \cite{williams2006gaussian}. Afterward, we enter a loop where, at each iteration $t$, the algorithm selects the next candidate point to evaluate $\mathbf{x}_t$ by maximizing 
an acquisition function $a(\mathbf{x})$. This acquisition function balances exploration and exploitation of the objective values \cite{garnett2023bayesian}. 
The acquisition function uses the posterior predictive distribution of the GP for the values of $f(\cdot)$ at each $\mathbf{x}$, 
\emph{i.e.}, $p(f(\mathbf{x})|\mathbf{x},\mathcal{D}_0)$, to estimate the expected utility of performing an evaluation of the objective at $\mathbf{x}$. 
Then, the objective function is evaluated at the maximizer of the acquisition. That is, $\mathbf{x}_t=\text{arg max}_{\mathbf{x}\in\mathcal{X}}\, a(\mathbf{x})$. 
This retrieves the noisy objective value at $\mathbf{x}_t$. Namely, $y_t = f(\mathbf{x}_t) + \epsilon_t$, where $\epsilon_t$ is assumed to be Gaussian noise. 
Next, this new observation is added to the dataset of observed points, $\mathcal{D}_t = \mathcal{D}_{t-1} \cup {(\mathbf{x}_t, y_t)}$, and the GP 
model is updated with $\mathcal{D}_t$. We illustrate these steps in Figure \ref{fig:gp_af}. The process iterates for 
a predefined number of steps or budget $T$, and when the execution finishes, then it returns the point $\mathbf{x}_t$ with best associated value 
$y_t$ in the noiseless setting. In the noisy setting, we may use the GP to remove the noise around each $y_t$ first. We summarize the steps of the 
described BO algorithm in Algorithm \ref{alg:bayesian_optimization}. Importantly, the GP and the acquisition $a(\cdot)$ are very cheap to evaluate and 
maximize, respectively, since they do not imply evaluating the actual objective. Thus, the extra time that the process described introduces can be 
considered negligible compared with the cost of evaluating the objective each time, which is assumed to be extremely expensive.
Because of the intelligent decisions made when choosing each evaluation point, BO methods often perform much better than
a random exploration of the input space \cite{snoek2012practical}.

When a GP is used as the underlying probabilistic model in BO, it is assumed that the objective function 
$f(\cdot)$ is a sample from such a GP. As a consequence, at any location $\mathbf{x} \in \mathcal{X}$, 
the distribution of the potential values of the latent function $f(\cdot)$ at $\mathbf{x}$, conditioned on the current observations 
$\mathcal{D}_{t-1}=\{(\mathbf{x}_i,y_i)\}_{i=1}^n$, \emph{i.e.}, $p(y|\mathbf{x},\mathcal{D}_{t-1})$, 
is Gaussian, with mean $\mu_n(\mathbf{x})$ and variance $v_n(\mathbf{x})$
\begin{align}
	\mu_n(\mathbf{x}) & = \mathbf{k}_n(\mathbf{x})^T (\mathbf{K}_n + \sigma^2 \mathbf{I})^{-1} \mathbf{y}_n\,, \label{eq:gp_mean}\\
	v_n(\mathbf{x}) & = k(\mathbf{x}, \mathbf{x}) - \mathbf{k}_n(\mathbf{x})^T (\mathbf{K}_n + \sigma^2 I)^{-1} \mathbf{k}_n(\mathbf{x}) \label{eq:gp_var}\,, 
\end{align}
where $\mathbf{k}_n(\mathbf{x})$ is a vector of cross-covariance terms between $f(\mathbf{x})$ and $\{f(\mathbf{x}_1), \ldots, f(\mathbf{x}_n)\}$;
$\mathbf{K}_n$ is the $n \times n$ covariance matrix between each $\{f(\mathbf{x}_1), \ldots, f(\mathbf{x}_n)\}$;
$k(\mathbf{x}, \mathbf{x})$ is the prior variance of $f(\mathbf{x})$; and $\sigma^2$ is the variance of the noise.
See \cite{williams2006gaussian} for further details. All these covariance values are computed in terms of a 
covariance function $k(\cdot,\cdot)$, which in the case of BO is often set to be the Mat\'ern covariance function \cite{snoek2012practical}.
All GP hyper-parameters such as length-scales, variance of the noise, etc. are tuned, \emph{e.g.}, by maximizing the 
marginal likelihood \cite{williams2006gaussian}.
  
\begin{algorithm}
\caption{Bayesian Optimization Vanilla Algorithm}
\label{alg:bayesian_optimization}
\begin{algorithmic}[1]
	\State \textbf{Input:} Initial dataset $\mathcal{D}_0 = \{(\mathbf{x}_i, y_i)\}_{i=1}^{n_0}$, GP prior $p(f)$, 
	acquisition function $a(\cdot)$, number of iterations $T$
\State Fit Gaussian process model to $\mathcal{D}_0$
\For {$t = 1$ to $T$}
	\State Select next query point $\mathbf{x}_t = \arg\max_{\mathbf{x}\in\mathcal{X}} a(\mathbf{x})$
	\State Evaluate the objective function $y_t = f(\mathbf{x}_t) + \epsilon_t$, where $\epsilon_t \sim \mathcal{N}(0, \sigma^2)$
	\State Augment the dataset $\mathcal{D}_t = \mathcal{D}_{t-1} \cup \{(\mathbf{x}_t, y_t)\}$
    \State Update Gaussian process model with $\mathcal{D}_t$
\EndFor
	\State \textbf{Output:} The best point found $\mathbf{x}_i$, where $i = \arg\max  y_i$, in the noiseless setting.
\end{algorithmic}
\end{algorithm}

Information-based BO methods use concepts from information theory to estimate the acquisition 
function $a(\mathbf{x})$ \cite{hernandez2014predictive,villemonteix2009informational,hennig2012entropy,hvarfner2022joint,tu2022joint}. Concretely, 
they use the notion of information gain to guide the selection of the next query point $\mathbf{x}_t$ to reduce the 
most the uncertainty about the objective global maximum. Uncertainty can be measured in terms of the entropy. 
In other words, one seeks to suggest the point whose evaluation maximizes the expected reduction of the entropy about some random 
variable that is related to the extremum, which, \emph{e.g.}, can be the location of the optimum in the input space 
$\mathbf{x}^\star$ or its associated value $y^\star$. These quantities can be regarded as random variables since the GP imposes a probability
distribution on the objective values, and hence on the global optimum of the objective.
The logic behind the process is to iteratively minimize the entropy of these quantities through 
evaluations that maximize the expected reduction of their entropy. Low entropy means a 
better knowledge of the values that the random variable can take. 
The current information about $\{\mathbf{x}^\star,y^\star\}$ can be 
estimated in terms of the negative differential entropy of the distribution 
$p(\{\mathbf{x}^\star,y^\star\}|\mathcal{D}_{t-1})$, where $\mathcal{D}_{t-1}$ is the dataset of
observations and evaluations performed so far. Such a distribution is fully specified by the GP model.
Specifically, in Joint Entropy Search (JES) one selects $\mathbf{x}_t$ by maximizing the following expression \cite{hvarfner2022joint}:
\begin{align} \label{eq:JES}
	a(\mathbf{x}) & = H[p(\{\mathbf{x}^\star,y^\star\}|\mathcal{D}_{t-1})] - 
	\mathds{E}_{p(y|\mathcal{D}_{t-1}, \mathbf{x})}[H[p(\{\mathbf{x}^\star,y^\star\}|\mathcal{D}_{t-1} \cup \{(\mathbf{x}, y)\})]]\,,
\end{align}
where $H[\cdot]$ denotes the differential entropy of the associated probability distribution; 
the first term in (\ref{eq:JES}) is just the entropy of the solution of the optimization problem
at the current iteration; and the second term in (\ref{eq:JES}) is the expected entropy after observing $y$ at $\mathbf{x}$.
Of course, we do not know the actual value of $y$. However, as described previously, the GP gives a predictive distribution 
for its values given the observed data. Namely, $p(y|\mathcal{D}_{t-1}, \mathbf{x})$, 
which is Gaussian with mean given by (\ref{eq:gp_mean}) and variance given by (\ref{eq:gp_var}) plus the noise variance $\sigma^2$, to
account for the fact that $y$ is a potential noisy version of $f(\mathbf{x})$.

A limitation of the approach described is that (\ref{eq:JES}) is too complicated to be evaluated in closed-form, which makes difficult 
its practical use. To overcome this difficulty, in \cite{hvarfner2022joint} it is suggested to use the fact that (\ref{eq:JES}) is the mutual 
information between $\{\mathbf{x}^\star,y^\star\}$ and $y$, $I(\{\mathbf{x}^\star,y^\star\};y)$. The mutual information is symmetric
and one can swap the roles between $\{\mathbf{x}^\star,y^\star\}$ and $y$ to obtain an alternative but equivalent expression 
to (\ref{eq:JES}) \cite{hernandez2014predictive}:
\begin{align}
	\label{eq:JES_simplified}
	a(\mathbf{x}) & = H[p(y|\mathcal{D}_{t-1},\mathbf{x})] -
	\mathds{E}_{p(\{\mathbf{x}^\star,y^\star\}|\mathcal{D}_{t-1})}[H[p(y|\{\mathbf{x}^\star,y^\star\},\mathcal{D}_{t-1},\mathbf{x})]]\,,
\end{align}
where now the first term in (\ref{eq:JES_simplified}) is just the entropy of the predictive distribution at $\mathbf{x}$ given the observed data; 
the expectation in (\ref{eq:JES_simplified}) can be simply approximated by Monte Carlo by sampling
$\{\mathbf{x}^\star,y^\star\}$ given the current observations; and $H[p(y|\{\mathbf{x}^\star,y^\star\},\mathcal{D}_{t-1},\mathbf{x})]$ 
is the entropy of the predictive distribution at $\mathbf{x}$ given that $\{\mathbf{x}^\star,y^\star\}$ is the solution
to the optimization problem. 

Approximately sampling $\{\mathbf{x}^\star,y^\star\}$ from the GP predictive distribution is tractable.
This can be done using a random Fourier feature approximation of the GP to sample functions from the GP posterior \cite{hernandez2014predictive}. 
These functional samples can then be optimized to obtain an approximate sample of $\{\mathbf{x}^\star,y^\star\}$.
The distribution $p(y|\{\mathbf{x}^\star,y^\star\},\mathcal{D}_{t-1},\mathbf{x})$ is intractable.
However, it can be approximated by a Gaussian distribution by making use of a truncated Gaussian to 
estimate the distribution of $f(\mathbf{x})$ given that $f(\mathbf{x}) < y^\star$ \cite{hvarfner2022joint}. 
This is precisely the approach followed in the BO framework BOTorch to evaluate the JES acquisition function \cite{balandat2020botorch}.

Finally, besides JES, there are other information-based strategies suggested in the literature: entropy search 
\cite{villemonteix2009informational,hennig2012entropy}, the first information-based BO method, and predictive entropy search \cite{hernandez2014predictive},
which only consider the entropy of $\mathbf{x}^\star$; and max value entropy search \cite{wang2017max}, which only considers the 
entropy of $y^\star$. JES has shown similar or better results than these strategies \cite{hvarfner2022joint}.

\section{Alpha Entropy Search} \label{SEC:RES}

As described in the previous section, the JES acquisition in (\ref{eq:JES_simplified}) is the mutual information between $\{\mathbf{x}^\star,y^\star\}$
and $y$, denoted $I(\{\mathbf{x}^\star,y^\star\};y)$, given the current observations collected so far $\mathcal{D}_{t-1}$. 
It is well known that the mutual information can be expressed in terms of the Kullback-Leibler (KL) divergence between
probability distributions. To illustrate this, consider two distributions $p(\mathbf{x})$ and $q(\mathbf{x})$. The KL-divergence between them is:
\begin{align} \label{EQ:KL}
	\text{KL}(p(\mathbf{x}) \| q(\mathbf{x})) & = \int p(\mathbf{x}) \log \frac{p(\mathbf{x})}{q(\mathbf{x})} \, d\mathbf{x} \,,
\end{align}
which is non-symmetric, non-negative, and equal to zero only when $p(\mathbf{x})=q(\mathbf{x})$. Thus, by taking a look at (\ref{EQ:KL})
we can see that we see that we can write (\ref{eq:JES_simplified}) as follows:
\begin{align}
	a(\mathbf{x}) &=  H[p(y|\mathcal{D}_{t-1},\mathbf{x})] -
	\mathds{E}_{p(\{\mathbf{x}^\star,y^\star\}|\mathcal{D}_{t-1})}[H[p(y|\{\mathbf{x}^\star,y^\star\},\mathcal{D}_{t-1},\mathbf{x})]] \nonumber \\
	& = I(\{\mathbf{x}^\star,y^\star\};y) = I(y;\{\mathbf{x}^\star,y^\star\}) \nonumber \\
	&= \text{KL}(p(\{\mathbf{x}^\star, y^\star\},y|\mathcal{D}_{t-1},\mathbf{x})||p(\{\mathbf{x}^\star,y^\star\}|\mathcal{D}_{t-1})p(y|\mathcal{D}_{t-1}, \mathbf{x}))\,. 
    \label{EQ:KL_PYX_PYPX}
\end{align}
Appendix \ref{SEC:AP1} shows the details of this identity. Note that (\ref{EQ:KL_PYX_PYPX}) is just the KL-divergence between a joint probability distribution
$p(\{\mathbf{x}^\star, y^\star\},y|\mathcal{D}_{t-1},\mathbf{x})$ and the corresponding factorizing distribution $p(\{\mathbf{x}^\star,y^\star\}|\mathcal{D}_{t-1})p(y|\mathcal{D}_{t-1}, \mathbf{x})$
that assumes independence between $y$ and $\{\mathbf{x}^\star, y^\star\}$.  This allows to interpret the JES acquisition in the following way. Specifically,
(\ref{EQ:KL_PYX_PYPX}) measures the level of dependency between $y$ and $\{\mathbf{x}^\star,y^\star\}$ at $\mathbf{x}$
in terms of the KL-divergence. The next point to evaluate is thus the one maximizing this level of dependency. The idea is that
because of the strong dependencies between $y$ and $\{\mathbf{x}^\star,y^\star\}$ at $\mathbf{x}$, observing $y$ will give a lot 
of information about the potential values of $\{\mathbf{x}^\star,y^\star\}$, \emph{i.e.}, the solution of the optimization problem, and will help the most to solve it.
In this work, we conjecture that other methods to estimate the level of dependencies between $\{\mathbf{x}^\star, y^\star\}$ and $y$ may provide better optimization 
results in some scenarios. With this idea in mind, we propose to consider a different divergence between probability distributions. Namely, the $\alpha$-divergence, which is described next.

\subsection{Amari's  $\alpha$-divergence} \label{SUB:ADIV}

\begin{figure}[htb!]
  \centering
  \includegraphics[width=1.0\textwidth]{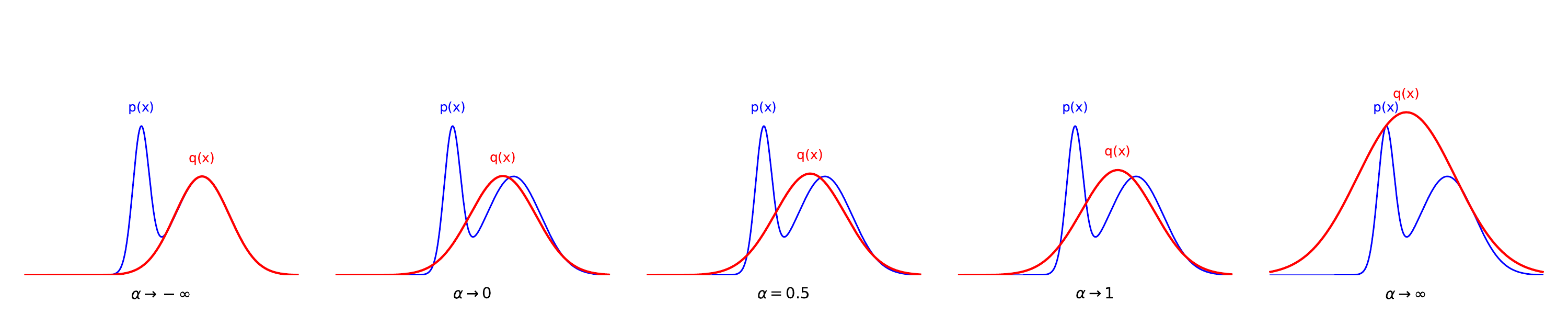}
  \caption{The Gaussian distribution $q(x)$ is fitted to $p(x)$ by minimizing Amari’s divergence
  with different values of $\alpha$. When $\alpha \rightarrow -\infty$, $\alpha$
  tries to match one mode of $p(x)$, and as $\alpha$ increases, $q(x)$ starts covering
  more of the entire distribution. Finally, when $\alpha \rightarrow \infty$,
  $q(x)$ covers $p(x)$ entirely. Reproduced from \cite{minka2005divergence}.}
  \label{FIG:AMARI-DIV}
\end{figure}

The KL-divergence \eqref{EQ:KL} can be generalized by replacing the natural
logarithm with the $\alpha$-logarithm and multiplying it by $\alpha^{-1}$ \cite{cichocki2010families}.
The $\alpha$-logarithm (also known as the Tsallis logarithm or the q-logarithm) is defined as:
\begin{align} \label{EQ:LOG-R}
    \log_{\alpha}(x) = \frac{x^{1-\alpha} - 1}{1 - \alpha}\,,
\end{align}
with $\alpha \in \mathds{R} \setminus \{1\}$ and such that $\log_{\alpha}(x) \rightarrow \log (x)$ when $\alpha \rightarrow 1$. 
This substitution leads to Amari's $\alpha$-divergence between probability distributions $p(\mathbf{x})$ and $q(\mathbf{x})$ \cite{amari1985},
which is defined as:
\begin{align} \label{EQ:DIV-A}
	D_{\alpha} (p(\mathbf{x})||q(\mathbf{x}))
     =
     \frac{1}{(1 - \alpha) \alpha}
     \left(
	1 - \int q(\mathbf{x})^{1-\alpha} p(\mathbf{x})^\alpha
	d\mathbf{x}
    \right)\,,
\end{align}
for $\alpha \in \mathds{R} \setminus \{0,1\}$.
This divergence is also non-negative and only equal to zero when the two distributions coincide.
Amari's divergence is parameterized by $\alpha$, which adjusts the sensitivity
to different regions of the probability distributions, allowing us to control
the emphasis of the divergence on specific differences between $p(\mathbf{x})$ and $q(\mathbf{x})$.
In Figure \ref{FIG:AMARI-DIV}, we illustrate how varying $\alpha$ affects the behavior of Amari's divergence. 
The figure show the results obtained when minimizing $D_{\alpha} (p(x)||q(x))$
when $p(x)$ is a multi-modal distribution and $q(x)$ is a Gaussian distribution.
For better visualization purposes, we have considered a version of the $\alpha$-divergence
that applies to un-normalized distributions \cite{minka2005divergence}.
The behavior of the $\alpha$-divergence with respect to $\alpha$ can be summarized as:
\begin{itemize}
    \item \textbf{$\alpha \rightarrow - \infty$}: The divergence emphasizes 
	capturing one of the modes of $p(x)$.
    \item \textbf{$\alpha \rightarrow 0$}: The divergence converges to the
    reversed Kullback-Leibler divergence $D_{\text{KL}}(q(x)\,\|\, p(x))$:
    \[
    \lim_{\alpha \rightarrow 0} D_{\alpha}(p(x)\,\|\, q(x)) = D_{\text{KL}}(q(x)\,\|\, p(x))
    \]
    At this point, $q(x)$ starts to capture more of the narrow mode of $p(x)$.
    
    \item \textbf{$\alpha = 0.5$}: $q(x)$ captures slightly more of the narrow mode of
    $p(x)$ compared to when $\alpha \rightarrow 0$. In this case the $\alpha$-divergence
		is equal to the Hellinger distance:
    \[
	    D_{0.5}(p(x)\,\|\, q(x)) = 2 \int \left(\sqrt{p(x)} - \sqrt{q(x)} \right)^2 d x
    \]
    
    \item \textbf{$\alpha \rightarrow 1$}: $q(x)$ captures even more of the narrow mode of
    $p(x)$ compared to $\alpha = 0.5$. The divergence converges to the
    direct Kullback-Leibler divergence $D_{\text{KL}}(p(x)\,\|\, x(x))$:
       \[
    \lim_{\alpha \rightarrow 1} D_{\alpha}(p(x)\,\|\, q(x)) = D_{\text{KL}}(p(x)\,\|\, q(x))
    \]
    \item \textbf{$\alpha \rightarrow \infty$}: In this limit, the divergence encourages
    $q(x)$ to cover the full distribution $p(x)$.
\end{itemize}
Thus, by adjusting $\alpha$, we can control how the divergence trades-off evaluating differences 
between the distributions at a single mode, and evaluating differences between them globally,
as illustrated by Figure \ref{FIG:AMARI-DIV}.

In most practical applications of $\alpha$-divergences for approximate
inference only values of $\alpha$ in the interval $(0,1)$ are considered, allowing
to interpolate between the two KL-divergences described above \cite{hernandez2016black,rodriguez2022adversarial}. 
Therefore, in this work we only consider such a range of values for $\alpha$. Note that considering 
different values of $\alpha$ will result in different acquisition functions for BO.

We expect that by replacing the KL-divergence of JES with a more general divergence,
such as Amari's $\alpha$-divergence, we will be able to adjust $\alpha$ and modulate the 
weight given to the discrepancies between distributions across different regions.
Specifically, different values of $\alpha$ will amplify or down-weight differences across different areas of 
the input space, leading to a hopefully better way, in some scenarios, of exploring the objective 
function towards its optimum. This substitution gives us the following acquisition function:
\begin{align}
	a_{\text{AES}}(\mathbf{x})
	&= D_{\alpha}(p(y,\{y^\star, \mathbf{x}^\star\}|\mathcal{D}_{t-1}, \mathbf{x})||p(\{y^\star, \mathbf{x}^\star\}|\mathcal{D}_{t-1}, \mathbf{x})p(y|\mathcal{D}_{t-1}, \mathbf{x})) \nonumber \\
    &= \textstyle
    \frac{1}{(1 - \alpha) \alpha}
    \left( 
	1 - \mathds{E}_{p(\{y^\star, \mathbf{x}^\star\}|\mathcal{D}_{t-1})}
            \left[
                \int 
		p(y|\mathcal{D}_{t-1}, \mathbf{x})
                \left(
		    \frac{p(y|\mathcal{D}_{t-1},\mathbf{x},\{y^\star, \mathbf{x}^\star\})}{p(y|\mathcal{D}_{t-1}, \mathbf{x})}
                \right)^\alpha dy 
            \right]
    \right)\,.
\label{EQ:aES_acq}
\end{align} 
Appendix \ref{SEC:AP2} shows the detailed derivation. As it is common with other information-based BO methods,
this expression is analytically intractable and requires approximation.
Specifically, neither the expectation in (\ref{EQ:aES_acq}) nor the conditional distribution
$p(y|\mathcal{D}_{t-1},\mathbf{x},\{y^\star, \mathbf{x}^\star\})$ can be computed in closed-form.  
Therefore, in the next section, we outline the specific approximations employed to evaluate 
and optimize the resulting acquisition function.

\subsection{Approximating the Conditional Distribution and the Expectation} \label{SUB:COMPDIV}

As described previously, the conditional predictive distribution 
$p(y|\mathcal{D}_{t-1},\mathbf{x},\{y^\star, \mathbf{x}^\star\})$ 
is intractable. This distribution describes the potential values that $y$ may take at $\mathbf{x}$
given the data observed so far $\mathcal{D}_{t-1}$ and that $\{y^\star, \mathbf{x}^\star\}$ is the 
solution of the optimization problem. 
Here, we adopt a similar approach as that of Joint Entropy Search (JES) \cite{hvarfner2022joint} 
to approximate such a conditional predictive distribution. For this, we incorporate $\{y^\star, \mathbf{x}^\star\}$ as extra data
and use a truncated Gaussian distribution. 
Specifically, if $\{y^\star, \mathbf{x}^\star\}$ is the solution of the optimization problem, $f(\mathbf{x})$ 
cannot take larger values than $y^\star$, and we know that $f(\mathbf{x}^\star)=y^\star$ must be fulfilled. 
This last condition can be satisfied by incorporating the pair $(\mathbf{x}^\star, y^\star)$ to the observed data 
$\mathcal{D}_{t-1}$, without observational noise. To guarantee that $f(\mathbf{x}) < y^\star$, we consider the Gaussian unconditional predictive
distribution $p(f(\mathbf{x})|\mathcal{D}_{t-1},\mathbf{x})=\mathcal{N}(f(\mathbf{x})|m(\mathbf{x}), v(\mathbf{x}))$,
with the mean and the variance respectively given by (\ref{eq:gp_mean}) and (\ref{eq:gp_var}), and 
truncate to zero the density above $y^\star$. The mean $m_\text{tr}(\mathbf{x})$ and variance 
$v_\text{tr}(\mathbf{x})$ of such a truncated Gaussian distribution are given by:
\begin{align}
	m_\text{tr}(\mathbf{x}) & =  m(\mathbf{x}) - \sqrt{v(\mathbf{x})} \frac{\phi(\beta)}{\Phi(\beta)}\,, \label{eq:gp_mean_tr}\\
	v_\text{tr}(\mathbf{x}) & = v(\mathbf{x}) \left[1 - \beta \frac{\phi(\beta)}{\Phi(\beta)} - \left(\frac{\phi(\beta)}{\Phi(\beta)} \right)^2 \right] \,,
	\label{eq:gp_var_tr}
\end{align}
where $\phi(\cdot)$ and $\Phi(\cdot)$ are respectively the p.d.f. and c.d.f of a standard Gaussian, and 
$\beta = (y^\star - m(\mathbf{x})) / \sqrt{v(\mathbf{x})}$.  
Taking into account the noise yields an extended skew distribution \cite{nguyen2022rectified} which
does not have a tractable density. Therefore, to account for the noise and compute the conditional predictive distribution of $y$, 
we approximate the truncated Gaussian with a Gaussian distribution, as in \cite{hvarfner2022joint}. 
Namely, $p(f(\mathbf{x})|\mathcal{D}_{t-1},\mathbf{x}, \{y^\star, \mathbf{x}^\star\})\approx \mathcal{N}(f(\mathbf{x})|m_\text{tr}(\mathbf{x}), v_\text{tr}(\mathbf{x}))$.
This results in a Gaussian approximation of the conditional distribution of $y$. That is,
$p(y|\mathcal{D}_{t-1},\mathbf{x}, \{y^\star, \mathbf{x}^\star\})\approx \mathcal{N}(y|m_\text{tr}(\mathbf{x}), v_\text{tr}(\mathbf{x}) + \sigma^2)$,
where $\sigma^2$ is the variance of the noise.

After approximating the conditional distribution, we need to evaluate the integral in (\ref{EQ:aES_acq}) w.r.t. $y$,
so that we can estimate the AES acquisition at $\mathbf{x}$.
This integral involves a product and a ratio between the predictive distribution
conditioned to the problem's solution and the unconditioned distribution, to the power of $\alpha$. 
Given that both distributions are Gaussian (after the approximations described above), we can evaluate
the integral in closed form using the exponential form of the Gaussian distribution:
\begin{align}
	\int 
		p(y|\mathcal{D}_{t-1}, \mathbf{x})
                \left(
		    \frac{p(y|\mathcal{D}_{t-1},\mathbf{x},\{y^\star, \mathbf{x}^\star\})}{p(y|\mathcal{D}_{t-1}, \mathbf{x})}
                \right)^\alpha dy 
	& = 
	\exp\left\{ (\alpha - 1) g(\bm{\eta}) -\alpha g(\bm{\eta}^\star)\right. \nonumber \\
	& \quad \left. + g((1 - \alpha)\bm{\eta} + \alpha \bm{\eta}^\star) \right\}\,,
\end{align} 
where $g(\bm{\eta})$ is the log-normalizer of a Gaussian with natural parameters $\bm{\eta}$,
$\bm{\eta}$ are the natural parameters of $p(y|\mathcal{D}_{t-1}, \mathbf{x})$, and
$\bm{\eta}^\star$ are the natural parameters of the Gaussian approximation of $p(y|\mathcal{D}_{t-1},\mathbf{x},\{y^\star, \mathbf{x}^\star\})$.
Specifically,
\begin{align}
	g({\bm \eta}) & = 0.5 \log ( 2 \pi) - 0.5 \log \eta_2 + 0.5 \frac{\eta_1^2}{\eta_2}\,, &
	\bm{\eta} & = \left(\frac{m(\mathbf{x})}{v(\mathbf{x})+\sigma^2}, \frac{1}{v(\mathbf{x}) + \sigma^2} \right)^\text{T}\,, \nonumber \\
	\bm{\eta}^\star & = \left(\frac{m_\text{tr}(\mathbf{x})}{v_\text{tr}(\mathbf{x})+\sigma^2}, \frac{1}{v_\text{tr}(\mathbf{x}) + \sigma^2} \right)^\text{T}\,,
\end{align}
where $m(\mathbf{x})$ and $v(\mathbf{x})$ are given by (\ref{eq:gp_mean}) and (\ref{eq:gp_var}), respectively, and where $m_\text{tr}(\mathbf{x})$ and 
$v_\text{tr}(\mathbf{x})$ are given by (\ref{eq:gp_mean_tr}) and (\ref{eq:gp_var_tr}), respectively. 
See Appendix \ref{SEC:AP3} for further details.

Furthermore, to approximate the expectation in \eqref{EQ:aES_acq}, we generate
samples of pairs of optimal locations and optimal values $\{\mathbf{x}^\star, y^\star\}$
from $p(\{y^\star, \mathbf{x}^\star\}|\mathcal{D}_{t-1})$ using a random features approximation of the GP \cite{rahimi2007random}, as in \cite{hernandez2014predictive,hvarfner2022joint}.
Such a random features approximation allows to sample functions from the GP posterior. These functions can be optimized to obtain an approximate
sample of $\{\mathbf{x}^\star, y^\star\}$. This is a common and efficient method often used in BO methods to sample from the posterior distribution 
and obtain optimal values. Given the samples, the expectation w.r.t. $p(\{y^\star, \mathbf{x}^\star\}|\mathcal{D}_{t-1})$ in 
\eqref{EQ:aES_acq} is evaluated employing a Monte Carlo approach. 

Using the approximations described in this section, the approximate acquisition function of AES is given by:
\begin{align}
	\tilde{a}_{\text{AES}}(\mathbf{x};\alpha)
    &= 
    \frac{1}{(1 - \alpha) \alpha}
    \left( 
	1 - \frac{1}{S} \sum_{s=1}^S
	\exp\left\{ (\alpha - 1) g(\bm{\eta}) -\alpha g(\bm{\eta}_s^\star) + g((1 - \alpha)\bm{\eta} + \alpha \bm{\eta}_s^\star) \right\}
	\right)\,,
\label{EQ:aES_acq_approx}
\end{align}
where we have considered $S$ samples of $\{y^\star, \mathbf{x}^\star\}$ to approximate the expectation 
in (\ref{EQ:aES_acq}) and $\bm{\eta}_s^\star$ are the natural parameters of the approximate conditional 
distribution $p(y|\mathcal{D}_{t-1},\mathbf{x},\{y^\star_s, \mathbf{x}_s^\star\})$, 
for the $s$-th sample of $\{\mathbf{x}^\star, y^\star\}$, denoted $\{y^\star_s, \mathbf{x}_s^\star\}$. 
In (\ref{EQ:aES_acq_approx}), $\alpha$ is a free parameter that will result in different acquisition functions.

When $S \rightarrow \infty$ and $\alpha \rightarrow 1$ one should expect that $\tilde{a}_{\text{AES}}(\mathbf{x}) \rightarrow \tilde{a}_\text{JES}(\mathbf{x})$, 
where $\tilde{a}_\text{JES}(\mathbf{x})$ is given by (\ref{EQ:KL_PYX_PYPX}), when a similar approximation is employed to that of AES.  
That is, $p(y|\mathcal{D}_{t-1},\mathbf{x},\{y^\star, \mathbf{x}^\star\})$  is also approximated using a truncated Gaussian distribution
and the expectation is approximated by Monte Carlo. However, this need not be the case. 
The reason for this is that when $S \rightarrow \infty$, $\tilde{a}_{\text{AES}}(\mathbf{x})\rightarrow 
\text{KL}(\tilde{p}(y,\{y^\star, \mathbf{x}^\star\}|\mathcal{D}_{t-1}, \mathbf{x})||p(\{y^\star, \mathbf{x}^\star\}|\mathcal{D}_{t-1}, \mathbf{x})p(y|\mathcal{D}_{t-1}, \mathbf{x}))$,
where $\tilde{p}(y,\{y^\star, \mathbf{x}^\star\}|\mathcal{D}_{t-1}, \mathbf{x})$ is an approximate joint distribution
that results from the ratio between the approximate conditional predictive distribution 
$\mathcal{N}(y|m_\text{tr}(\mathbf{x}),v_\text{tr}(\mathbf{x})+\sigma^2)\approx p(y|\mathcal{D}_{t-1},\mathbf{x},\{y^\star, \mathbf{x}^\star\})$ 
and the unconditioned predictive distribution $p(y|\mathcal{D}_{t-1},\mathbf{x})$.
By contrast, in general, when it is fulfilled that $S \rightarrow \infty$
we have that 
$\tilde{a}_{\text{JES}}(\mathbf{x})\nrightarrow \text{KL}(\tilde{p}(\{\mathbf{x}^\star, y^\star\},y|\mathcal{D}_{t-1},\mathbf{x})||p(\{\mathbf{x}^\star,y^\star\}|\mathcal{D}_{t-1})p(y|\mathcal{D}_{t-1}, \mathbf{x}))$.
The reason is that it is possible to show that in the JES approximate acquisition function, sometimes the exact 
joint distribution is used $p(\{\mathbf{x}^\star, y^\star\},y|\mathcal{D}_{t-1},\mathbf{x})$, while other times the approximate 
joint distribution $\tilde{p}(\{\mathbf{x}^\star, y^\star\},y|\mathcal{D}_{t-1},\mathbf{x})$ is used instead. Appendix   \ref{SEC:AP4} shows the 
details. Summing up, (\ref{EQ:aES_acq_approx}) results in a different approximation to that of JES when $\alpha \rightarrow 1$. 
However, the differences between both approximations are small according to our experiments. 

\subsection{An Ensemble of Acquisition Functions} \label{SUB:RESACQ}

The acquisition function of AES, given in (\ref{EQ:aES_acq_approx}), depends on the parameter $\alpha$.
The selection of a value for such a parameter is non-trivial, as it affects the divergence, 
and hence the acquisition function. Specifically, different values of $\alpha$ may yield better or 
worse optimization performance, depending on the particular optimization problem. Moreover, in our experiments, 
we did not observe a particular value for $\alpha$ that generally performs better than the JES 
acquisition function. 

Motivated by the aforementioned result and by the fact we would like to avoid choosing specific
values for $\alpha$, we investigate here the possibility of considering simultaneously a range of 
values for $\alpha \in (0,1)$. The result is an ensemble of acquisition functions, whose optimization 
results are expected to be better than those of the JES acquisition function.
In particular, we consider eleven values for $\alpha$ equally spaced in the interval $(0,1)$, where 
the first value is equal to $0.001$ and the last value is equal to $0.999$. 
This range of values has been employed before in the literature of approximate 
inference and provides a good coverage of different $\alpha$ values 
\cite{villacampa2020alpha,villacampa2022alpha,rodriguez2022adversarial}.
Moreover, the value $\alpha=0.999$ results in the direct KL-divergence, which should provide
similar, but not exactly the same results (for the reasons given in the previous section) to those of the 
JES acquisition function. The value $\alpha=0.001$ is expected to result in the reversed KL-divergence.
The range of values of $\alpha$ considered interpolates between these two divergences.
Last, in our ensemble of acquisition functions, we normalized each acquisition function
by its maximum value, to give equal weight in the ensemble to each value of $\alpha$. Specifically,
the ensemble acquisition function is:
\begin{align}
	\tilde{a}_{\text{ens}}(\mathbf{x}) &= \sum_{\alpha \in \Gamma} \frac{1}{w_\alpha} \tilde{a}_\text{AES}(\mathbf{x};\alpha)\,,
\label{EQ:ensemble}
\end{align}
where $\tilde{a}_\text{AES}(\mathbf{x};\alpha)$ is given by (\ref{EQ:aES_acq_approx}), 
$\Gamma$ is a set with the 11 different values of $\alpha$ considered, and $w_\alpha = \tilde{a}_\text{AES}(\mathbf{x}_\text{max}^\alpha;\alpha)$
with $\mathbf{x}_\text{max}^\alpha = \text{arg max}_\mathbf{x} \, \tilde{a}_\text{AES}(\mathbf{x};\alpha)$.
That is, we normalize each different acquisition function by its maximum value.
Importantly, in each acquisition function we use the same $S$ samples of $\{\mathbf{x}^\star, y^\star\}$, which are generated only one time instead of
11 times. Thus, the over-head of sampling $\{\mathbf{x}^\star, y^\star\}$ in this acquisition function is the same as 
that of JES.

\begin{figure}[htb!]
  \centering
  \includegraphics[width=1.0\textwidth]{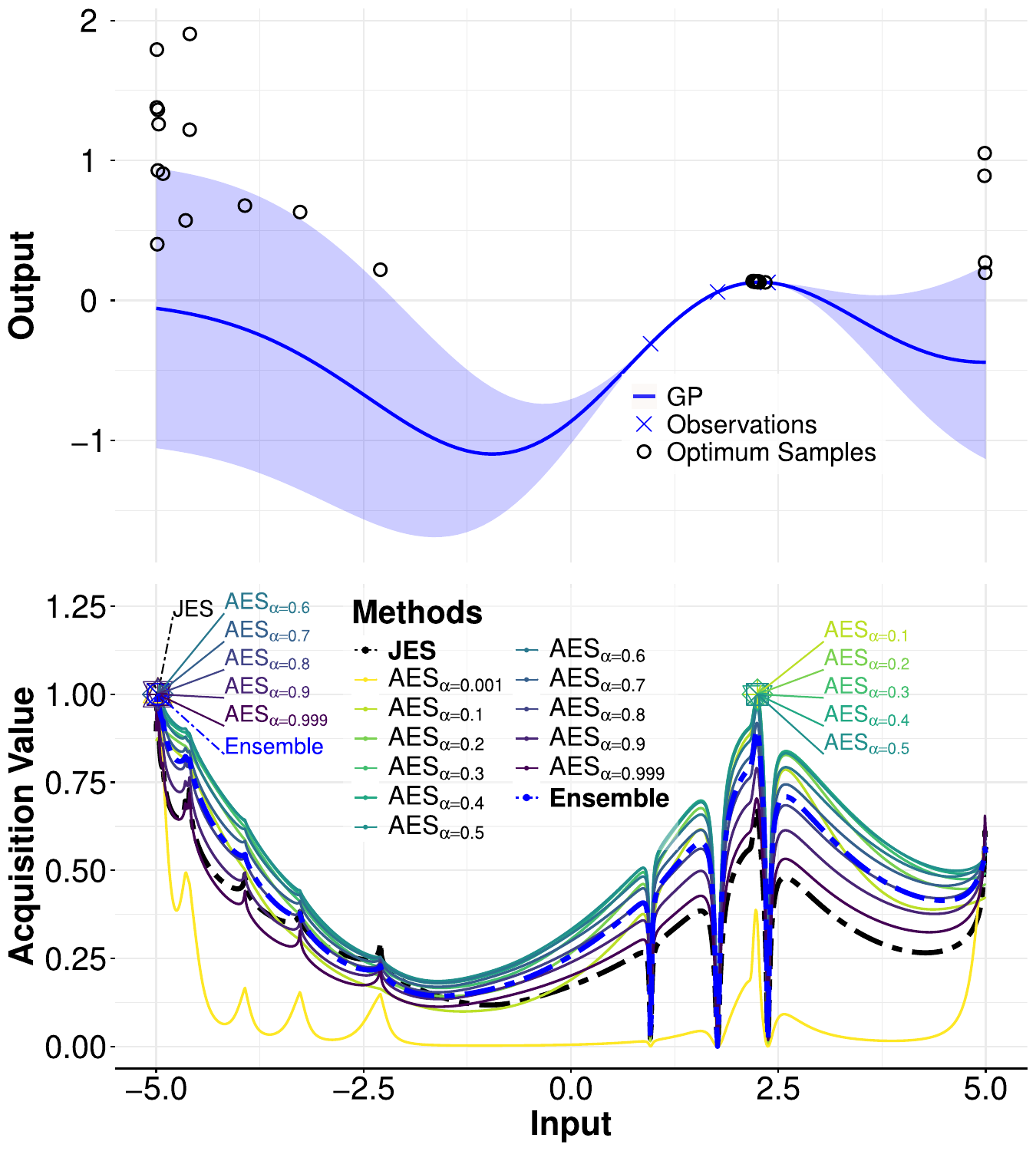}
	\caption{(bottom) Comparison of AES for different $\alpha$ values, JES and the ensemble acquisition function. We also display the maximum of each acquisition function.
	(top) Predictive distribution of the GP and generated samples of $\{\mathbf{x}^\star, y^\star\}$. The acquisition functions have been normalized so that the maximum is equal to one
	for a better visualization. Best viewed in color.} 
  \label{fig:comparison_alpha_values}
\end{figure}

Figure \ref{fig:comparison_alpha_values} (bottom) shows a comparison of the AES acquisition function for different values 
of $\alpha$, in a one dimensional problem. The acquisition function for JES is also displayed, and that of the ensemble method
described in the previous paragraph.  For each acquisition function we also display its maximum.
For the sake of visualization, each acquisition has been normalized so that its maximum is equal to $1$. 
Figure \ref{fig:comparison_alpha_values} (top) shows the 
predictive distribution of the GP and the location in the input space of the generated samples of $\{\mathbf{x}^\star, y^\star\}$. Here, we 
considered 32 samples of $\{\mathbf{x}^\star, y^\star\}$ and a noiseless setting.

Figure \ref{fig:comparison_alpha_values} shows that AES for $\alpha =0.999$
gives similar values to JES, as expected, but not exactly the same ones.
The reason for this is given in the previous section.
We also observe that the peaks, \emph{i.e.}, local maxima of JES and AES, for large values of $\alpha$, 
often occur at the locations where the samples of $\{\mathbf{x}^\star, y^\star\}$ are found.
This makes sense for JES since the entropy reduction is 
maximum there. Specifically, observing $\{\mathbf{x}^\star_s, y^\star_s\}$ reduces the predictive variance to almost zero at $\mathbf{x}_s^\star$ in the noiseless 
setting. Since AES is approximating JES for larger values of $\alpha$, a similar behavior is expected for AES in such a setting. 
Thus, both acquisition functions generate peaks at the sampled points, effectively making both methods equivalent to Thompson Sampling 
when we only consider one sample of $\{\mathbf{x}^\star_s, y^\star_s\}$ \cite{shahriari2015taking}. We observe that the number of local maxima 
is significantly reduced for AES, for smaller values of $\alpha$, and also for the ensemble acquisition function. 

The smaller number of local maxima in 
the ensemble acquisition function w.r.t. JES is statistically significant as indicated by Table \ref{tab:n_local_max}, which displays 
the average number of local maxima of JES and the ensemble method across 100 repetitions of the previous experiment, in which different $S$ 
samples are generated of $\{\mathbf{x}^\star,y^\star\}$. The average number of local maxima of AES for smaller values of $\alpha$ (not shown) is also significantly smaller than that of JES.
However, in our experiments we did not observe that such values of $\alpha$ significantly outperformed the optimization results of JES
in real-world problems. Only in synthetic problems. In practice, finding the global maximum is challenging, and optimization algorithms often 
converge to local minima. By averaging over $\alpha$, the ensemble acquisition becomes less rugged than when using 
$\alpha$ values close to $1.0$, although not as smooth as when
using $\alpha$ values near $0.1$. This can be observed in Figure \ref{fig:comparison_alpha_values}.
We believe that the averaging approach of the ensemble method results in a more robust acquisition 
function with a smaller number of local maxima (as a beneficial side effect) in the noiseless setting.
Appendix \ref{SEC:AP5} shows a similar analysis in the noisy setting, where the effect of local maxima is less pronounced, since 
the presence of noise does not result in almost zero predictive variances at each $\mathbf{x}_s^\star$.

\begin{table}[htb!]
	\centering
	\caption{Average number of local maxima for each method over 100 repetitions.}
	\label{tab:n_local_max}
	\begin{tabular}{lr@{$\pm$}l}
		\hline
		{\bf Method} & \multicolumn{2}{c}{\bf \# of Local Maxima}\\
		\hline
		{\bf JES }     & 17.60 & 0.56\\
		{\bf Ensemble} & 12.53 & 0.49\\
		\hline
	\end{tabular}
\end{table}

Finally, we note that the computational cost of AES for a particular $\alpha$ is the same as 
the one of JES because both methods use the same approximation to calculate the conditional predictive
distribution $p(y | \mathcal{D}_{t-1}, \mathbf{x}, \{\mathbf{x}^\star,y^\star\})$. Therefore, the cost of evaluating the ensemble acquisition function
is $O(|\Gamma|S)$, where $|\Gamma|$ is the number of $\alpha$ values considered 
(eleven in our setting) and $S$ is the number of optimal samples of $\{\mathbf{x}^\star,y^\star\}$ generated.
The ensemble method has to add to this cost the over-head of having to optimize each individual AES acquisition
function, for each value of $\alpha$, to obtain the individual maxima described in (\ref{EQ:ensemble}).
One may argue that since the ensemble method is $|\Gamma|$ times more expensive than JES, one could increase the number
of samples $S$ in JES, at the same computational cost, to obtain better results. However, our experiments show
that, in a noiseless evaluation setting, increasing $S$ in JES does not lead to better results.


\section{Related Work} \label{SEC:RELATEDWORK}

After introducing our proposed method, AES, and the associated ensemble method, we describe in 
this section related techniques from the literature and highlight the main differences of AES with respect 
to them. We particularly focus on information-based BO acquisition functions since our method falls in that category. Moreover,
there is empirical evidence that information-based BO acquisition functions provide, in several problems, better results than 
other classical acquisition functions such as Expected Improvement or Upper Confidence Bound 
\cite{hennig2012entropy,hernandez2014predictive,hvarfner2022joint}.

The Entropy Search (ES) strategy was considered first in \cite{villemonteix2009informational}, where the direct reduction of the 
current entropy of $\mathbf{x}^\star$ given the observed data was targeted. Since the estimation of the entropy of $\mathbf{x}^\star$ is 
intractable, an expensive sampling based strategy based on discretizing the input space using a grid was considered. This method
is practical, but makes difficult the direct optimization of the acquisition function. A set of candidate points has to be used.
Furthermore, it is based on considering only the entropy of $\mathbf{x}^\star$, completely ignoring $y^\star$.  Additionally, it does not consider 
the fact that the acquisition function is the mutual information between $y$ and $\mathbf{x}^\star$ and does not swap these two variables 
to simply its expression, resulting in a complicated acquisition function.

In \cite{hennig2012entropy} an alternative approximation of the ES acquisition function was considered. Instead of
relying on a discretizing and sampling approach, as in \cite{villemonteix2009informational}, the expectation propagation (EP) algorithm was proposed
to estimate the conditional entropy of $\mathbf{x}^\star$ after performing an evaluation at a new candidate point. This resulted in an approximation of 
the acquisition that is complicated and difficult to evaluate, although it provided gradients, unlike the approximation in \cite{villemonteix2009informational}.
This enables the use of gradient based optimization algorithms to maximize the acquisition.
In any case, the method still considered only $\mathbf{x}^\star$ and ignored $y^\star$. Furthermore, it did not simplify the 
acquisition using the mutual information trick we consider here, which allows to swap $\mathbf{x}^\star$ and $y$.

In \cite{hernandez2014predictive} it is proposed to simplify the expression for the acquisition function of ES
by swapping  $\mathbf{x}^\star$ and $y$, as a consequence of the symmetry of the mutual information.
The resulting acquisition function is known as Predictive Entropy Search (PES). PES results in a much simpler expression
for the acquisition function than the one in \cite{hennig2012entropy}. The difficulty is still found in approximating the conditional predictive 
distribution given a sample of $\mathbf{x}^\star$. The samples of $\mathbf{x}^\star$ are obtained by optimizing functions generated
from the GP posterior. For this, a random feature approximation of the GP is used. To approximate the conditional predictive 
distribution, again, the EP algorithm is employed. EP is expensive and consists in approximating
with Gaussians several non-Gaussian factors introduced to guarantee compatibility with the sample 
of $\mathbf{x}^\star$. PES ignored the value of $y^\star$. Empirical evidence shows that PES performs much better than ES.

In \cite{hoffman2015output,wang2017max}, it is considered the entropy of the optimum
in the output space, $y^\star$, instead of the of the optimum in the input space,
$\mathbf{x}^\star$. Specifically, it is suggested to select the next evaluation as
the one that minimizes the expected entropy of $y^\star$ the most.
This method is known as Max-value Entropy Search (MES). As in PES, the acquisition function is simplified using
the mutual information trick. The main advantage of MES is that the conditional predictive distribution has to be computed w.r.t. $y^\star$, instead 
of $\mathbf{x}^\star$, which significantly simplifies the evaluation of the acquisition. Specifically, given a sample of $y^\star$, such a conditional 
distribution can be approximated using a truncated Gaussian distribution. MES gives similar to or better results than those 
of PES and performs better w.r.t. the number of samples of $y^\star$ \cite{wang2017max}.

MES is computationally expensive due to the fact that it requires sampling the optimum of the optimization 
problem at each step using a random features approximation of the GP. Furthermore, such an approximation is only valid for particular GP 
covariance functions. The Fast Information-theoretic Bayesian Optimization (FITBO) \cite{ru2018fast} 
is an alternative approach that avoids to sample the global optimum. For this, extra approximations are introduced, based
on expressing the unknown objective function in a parabolic form. This introduces an extra hyper-parameter, but circumvents 
the process of sampling the global minimum. After more approximations, the entropy reduction of FITBO is purely analytical, 
being computationally fast. FITBO also considers the expected reduction in the entropy of $y^\star$, as MES.

There are some methods proposed to alleviate some limitations of PES and MES. For example, 
the Trusted Maximizers Entropy Search (TES) method \cite{nguyen2021trusted} reduces the 
over-all cost of PES. TES selects the next point to evaluate from a finite set of trusted maximizers. 
These trusted maximizers are inputs optimizing functions that are sampled from the GP posterior. 
Evaluating TES requires either only a stochastic approximation 
with sampling, or a deterministic approximation with EP. TES provides similar or better results than those of PES
at a smaller computational cost. 

We believe that the described improvements of MES and PES are orthogonal to our work and they 
could in principle be also used in AES. 

Another improvement is the rectified version of MES (RMES) \cite{nguyen2022rectified},
which solves known issues of MES in the noisy setting, where the conditional predictive distribution is a
truncated Gaussian random variable that, when contaminated by Gaussian noise, lacks a closed-form density.
RMES addresses this issue obtaining a closed-form density for the conditional distribution using the 
reparametrization trick. RMES gives similar or better results than MES. In AES, we employ a similar correction
in the noisy setting to the one introduced in \cite{hvarfner2022joint}, when there is observational noise in the 
objective. Such a correction is also incorporated in the implementation of MES that we use in our experiments.

Recently, joint entropy search (JES) has been proposed as an improvement over MES and PES, presenting state-of-the-art optimization 
performance \cite{hvarfner2022joint,tu2022joint}. Concretely, this approach considers the entropy over the joint distribution of both 
the global optimum in the input space and the output space. Namely, $\{\mathbf{x}^\star,y^\star\}$. The acquisition function simply chooses the 
next point to reduce the most the expected entropy of that random variable. The expression for such an acquisition is simplified
using the mutual information trick, by swapping $\{\mathbf{x}^\star,y^\star\}$ and $y$, and the conditional predictive distribution 
is approximated by a Gaussian with the same moments of a truncated Gaussian distribution contaminated by Gaussian noise (in the noisy 
setting). JES gives similar or better results than those of MES and PES \cite{hvarfner2022joint}. 

All the strategies described so far in this section result in an acquisition function $a(\mathbf{x})$ that estimates 
the mutual information between the solution of the optimization problem (\emph{i.e.}, $\mathbf{x}^\star$, $y^\star$ or $\{\mathbf{x}^\star,y^\star\}$) 
and $y$, the observation at $\mathbf{x}$. As described in Section \ref{SEC:RES}, this can be shown to be equivalent to
the Kullback-Leibler (KL) divergence between two probability distributions. By contrast, the proposed strategy AES, and the 
associated ensemble method, replace this divergence with the more general $\alpha$-divergence \cite{amari1985}, which includes
a parameter $\alpha$. The $\alpha$-divergence generalizes the KL-divergence. Specifically, the $\alpha$ parameter can be used to give 
higher importance to particular differences among the probability distributions, and when $\alpha \rightarrow 1$, it results in the 
regular the KL-divergence used in ES methods.

We are not aware of the use of $\alpha$-divergences in the context of BO. However, the use of Shannon's entropy in ES has 
been generalized in \cite{neiswanger2022generalizing} to a broader class of loss functions that result in other acquisition 
functions from the literature such as Knowledge Gradient or Expected Improvement. For this, it is suggested the use of decision-theoretic 
entropies parameterized by a problem-specific action set $\mathcal{A}$ and a loss function $\ell$. In the case of ES, the loss is the 
negative logarithm, and the action set contains the GP posterior distribution. In spite of this, there is not a clear and direct 
way of using the framework of \cite{neiswanger2022generalizing} to obtain the acquisition function of AES nor the ensemble method
we propose here.

In the literature, there are several works that have used $\alpha$-divergences for approximate Bayesian inference.
These works also consider values of $\alpha$ in the interval $(0,1)$. In particular, in \cite{hernandez2016black} 
the approximate minimization of $\alpha$-divergences is used to approximate the posterior distribution of Bayesian neural networks (NN) with a 
Gaussian distribution. The method is known as Black-box alpha. The results obtained show that intermediate values of $\alpha$, \emph{e.g.}, $\alpha=0.5$, leads 
to better results than other values of $\alpha$ that result in the approximate minimization of the KL-divergence.
The minimization of $\alpha$-divergences in the context of dropout for 
approximate inference in Bayesian NN is explored in \cite{li2017dropout}.
A generalization of Black-box alpha is considered in \cite{rodriguez2022adversarial}, where 
the approximate distribution is a flexible implicit distribution obtained by letting some 
noise go through a deep NN. The results obtained show that, in general, larger values of $\alpha$ may lead to better predictive
distributions in the case of regression problems, and smaller values of $\alpha$ may lead to better test mean squared error.
The use of $\alpha$-divergences to approximate the posterior distribution has also been studied in the case of GPs for regression
and binary classification \cite{bui2017unifying}, GPs for multi-class classification \cite{villacampa2020alpha} and deep 
GPs \cite{villacampa2022alpha}. In such a setting, one often observes that $\alpha\approx 1$ tends to give better predictive 
distributions, while $\alpha \approx 0$ tends to minimize the test mean squared error. The use of $\alpha$-divergences for 
approximate inference in generative models has also been explored in \cite{bui2016,midgleySSSH23}, with superior results to those obtained by 
the KL-divergence.

Finally, the use of several acquisition functions to solve an optimization problem has been considered before.
In \cite{hoffman2011portfolio} the authors propose a strategy to sample from a pool of acquisitions the 
acquisition function to use at each iteration of the BO algorithm. The idea is to favor those acquisition functions that lead to 
better results at each iteration and to penalize those that do not. The resulting method is called GP-hedge. A limitation is, however,
that GP-hedge assumes there is an optimal acquisition function in the pool of acquisition functions considered.
Specifically, GP-hedge does not combine the acquisition functions. We evaluated GP-hedge via preliminary experiments, and compared 
its performance w.r.t. our ensemble method based on AES. The ensemble method performed better, probably as a consequence 
of combining the acquisition functions instead of simply choosing one among them at each iteration according to some weights.

\section{Experiments} \label{SEC:EXP}


In this section, we compare, across several optimization problems,
AES and the ensemble method with other strategies for BO. 
Namely, we compare results with random search, and the acquisition functions 
based on the KL-divergence described in Section \ref{SEC:RELATEDWORK}, \emph{i.e.}, JES, 
MES, and PES.  Random search simply chooses randomly the next point to evaluate. 
We do not compare results with other BO methods 
such as Expected Improvement or Upper Confidence Bound since several works from the literature already 
compare them with PES, MES and JES, showing better results in several optimization 
problems \cite{hernandez2014predictive,wang2017max,hvarfner2022joint}.  
In the ensemble method, we consider eleven values for $\alpha$, \emph{i.e.}, $\{0.001, 0.1, 0.2, \ldots, 0.9, 0.999\}$.
Our implementation of AES and the ensemble method are available at
\url{https://github.com/fernandezdaniel/alphaES}. For the other acquisition functions,
PES, MES, and JES, we simply used the implementation provided in BOTorch \cite{balandat2020botorch}. 
In all problems, the goal is to maximize the objective. Minimization can be simply achieved by optimizing $-f(\mathbf{x})$.

In our experiments, we use $S=32$ samples of the problem's solution to estimate the acquisition of
AES, PES and MES. These samples are generated using a random feature approximation of the GP, 
as described in \cite{rahimi2007random,hernandez2014predictive}. We use a Mat\'ern $5/2$ covariance
function with ARD and fit the GP via maximum marginal-likelihood. These are standard choices in BOTorch.  
In each optimization problem, we use a set of 25 initial observations randomly chosen. This number of initial 
observations is expected to guarantee that the maximum marginal-likelihood approach used to fit the GP does not result 
in over-fitting. To maximize each acquisition, we used L-BFGS-B with $1$ restart and $200$ points to generate 
the initial conditions from which the starting point of the optimization is randomly chosen. 
See \cite{balandat2020botorch} for further details. Preliminary experiments in which we increase
the number of restarts and the number of points used to generate the initial conditions give similar results. See 
Appendix \ref{SEC:AP6}. We report average results across 100 repetitions of the experiments with different random seeds and
show the corresponding error bars. At each iteration, the BO method recommends the best observation, in 
the noiseless setting. In the noisy setting, we recommend the observation with the best predictive mean. This is done 
to remove the observational noise. In general, this gives better results than optimizing the GP mean 
to make a recommendation. 

\subsection{Quality of the Approximation of the Acquisition Function} \label{SUB:QUAACQ}

We investigate in this section the accuracy of the approximation of the AES acquisition function suggested in (\ref{EQ:aES_acq_approx}).
For this, we compare in a 1-dimensional toy problem the AES acquisition function with the exact acquisition function
it is targeting, estimated using a more accurate Monte Carlo method. The $\alpha$ values considered for AES and the exact method are same for 
comparison.  Since the problem is one-dimensional, the calculation of 
the exact acquisition is feasible. Specifically, for each sample of $\{y^\star, \mathbf{x}^\star\}$, 
$\{y_s^\star, \mathbf{x}_s^\star\}$, 
the conditional density 
$p(y|\mathcal{D}_{t-1},\mathbf{x},\{y_s^\star, \mathbf{x}_s^\star\})$
is estimated using a kernel density estimator on samples from the 
GP posterior at $\mathbf{x}$ that are compatible with $\{y_s^\star, \mathbf{x}_s^\star\}$. 
The integral w.r.t. $y$ in (\ref{EQ:aES_acq}) is estimated using quadrature, since it is one-dimensional.
We consider a large number of samples $S=6000$ of $\{y^\star, \mathbf{x}^\star\}$ to 
approximate the expectation in (\ref{EQ:aES_acq}). No significant changes are observed 
above that number of samples. Note that these operations are significantly more costly than 
the evaluation of AES via (\ref{EQ:aES_acq_approx}) for any value of $\alpha$, and become intractable 
in general. However, they are expected to give a good estimate of the AES acquisition function in this
simple problem. We also compare the ensemble acquisition function described in (\ref{EQ:ensemble}) with the 
exact ensemble acquisition function, estimated by a weighted average of the exact AES acquisition for 
the 11 values of $\alpha$ described above. For simplicity, we assume a noiseless evaluation setting.

\begin{figure*}[tbh!]
	\begin{tabular}{cc}
		\includegraphics[width=0.49\textwidth]{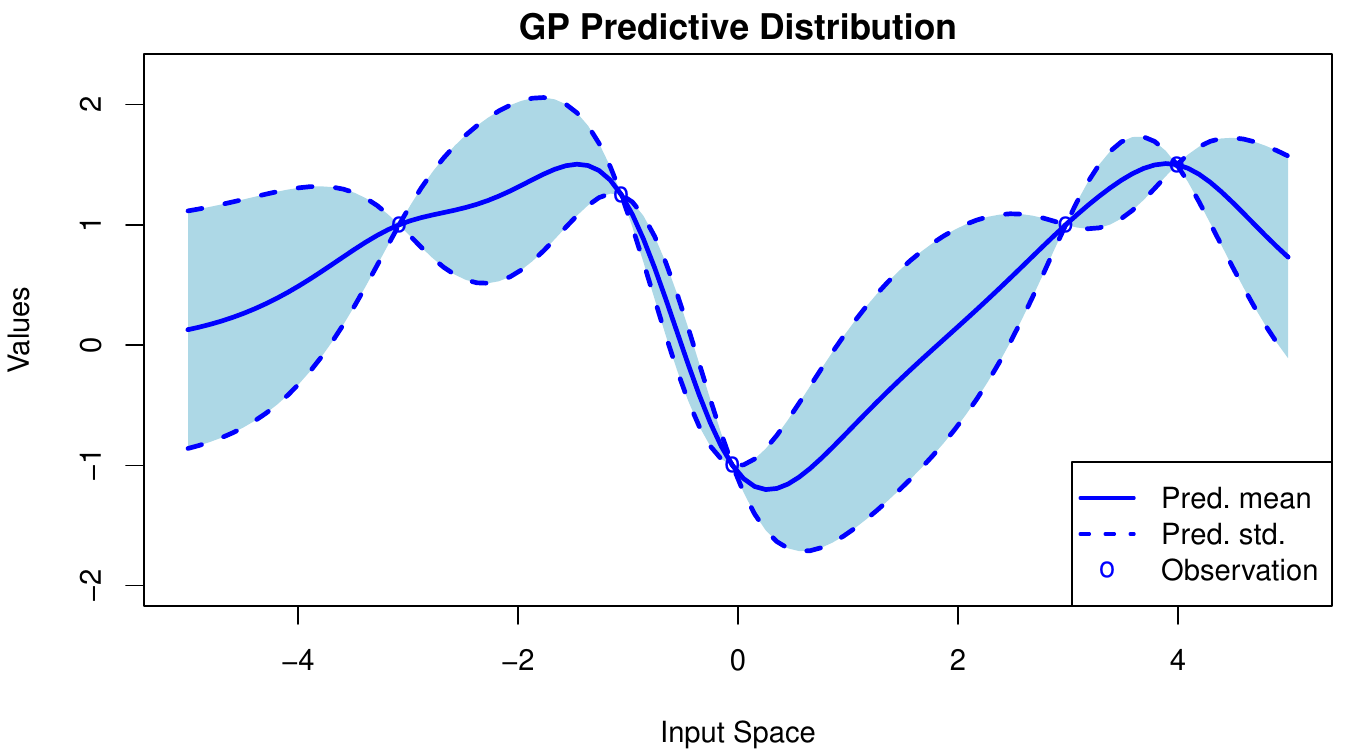}
		\includegraphics[width=0.49\textwidth]{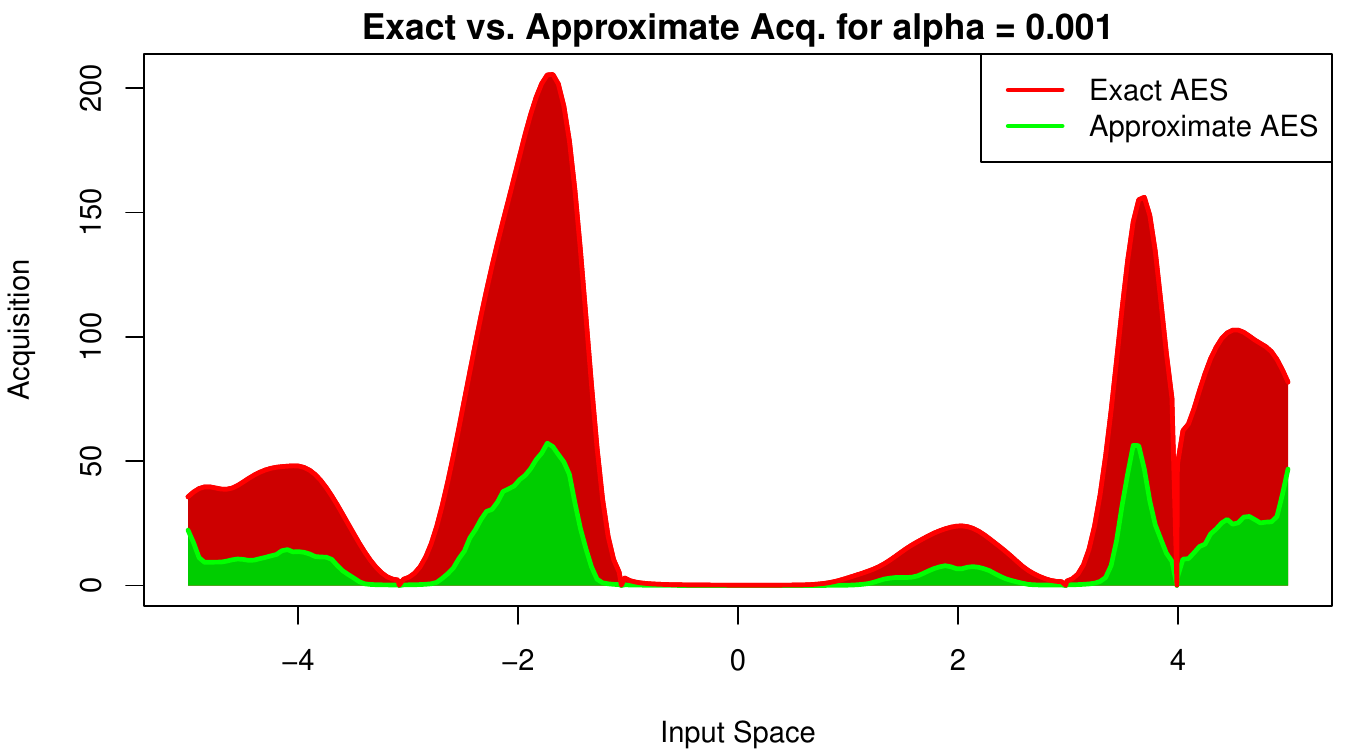} \\
		\includegraphics[width=0.49\textwidth]{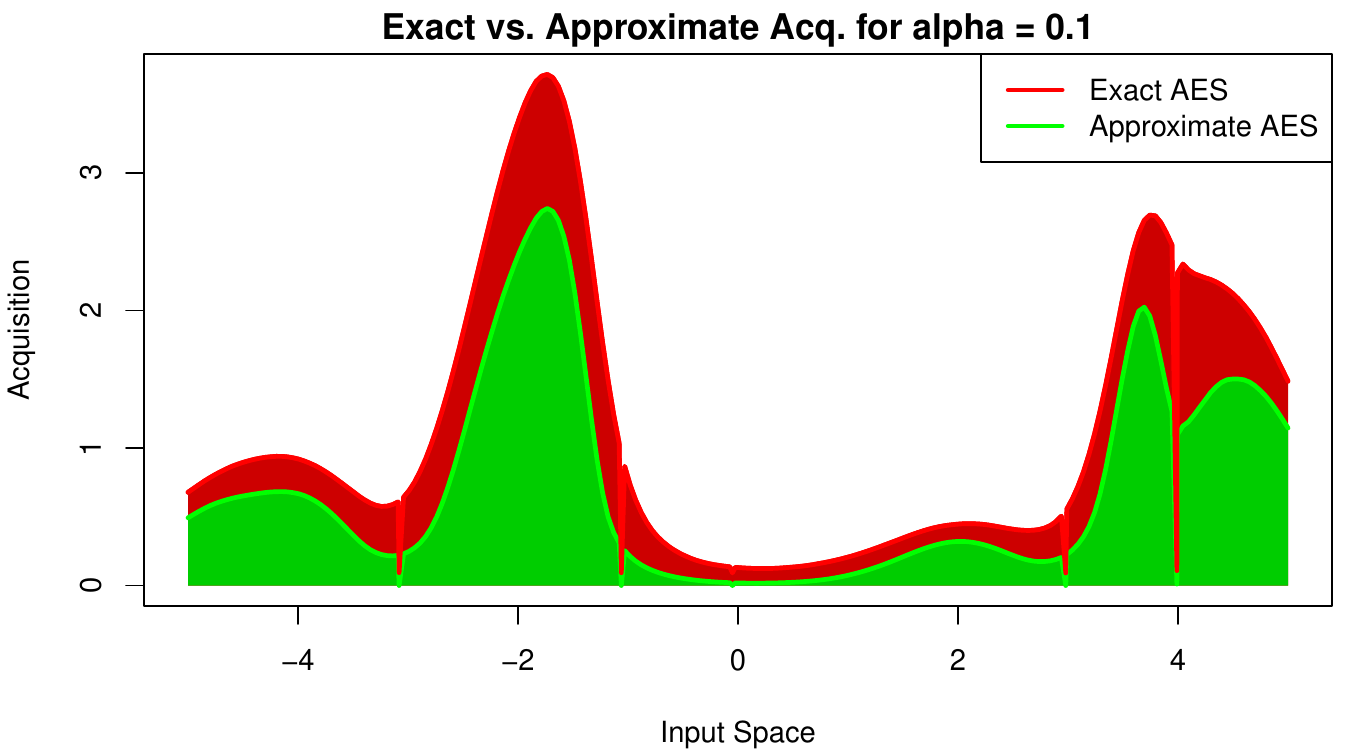}
		\includegraphics[width=0.49\textwidth]{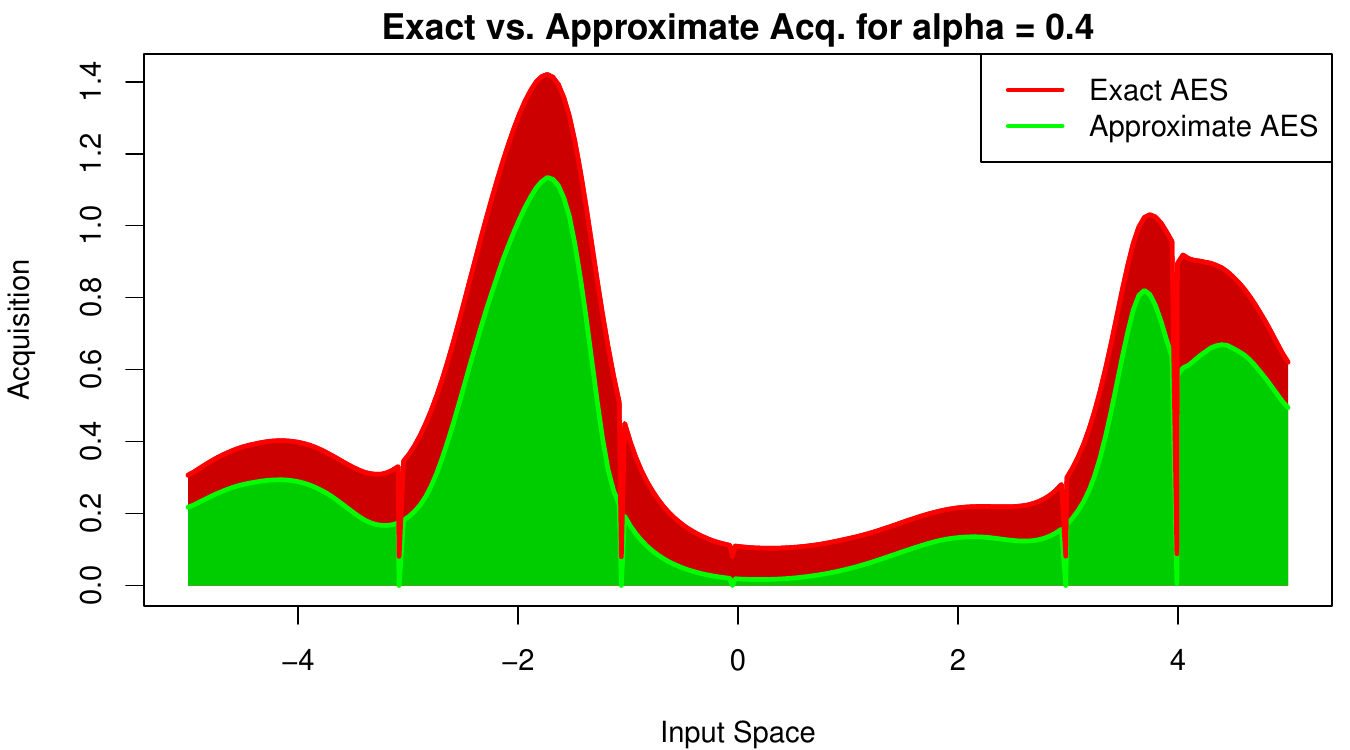} \\
		\includegraphics[width=0.49\textwidth]{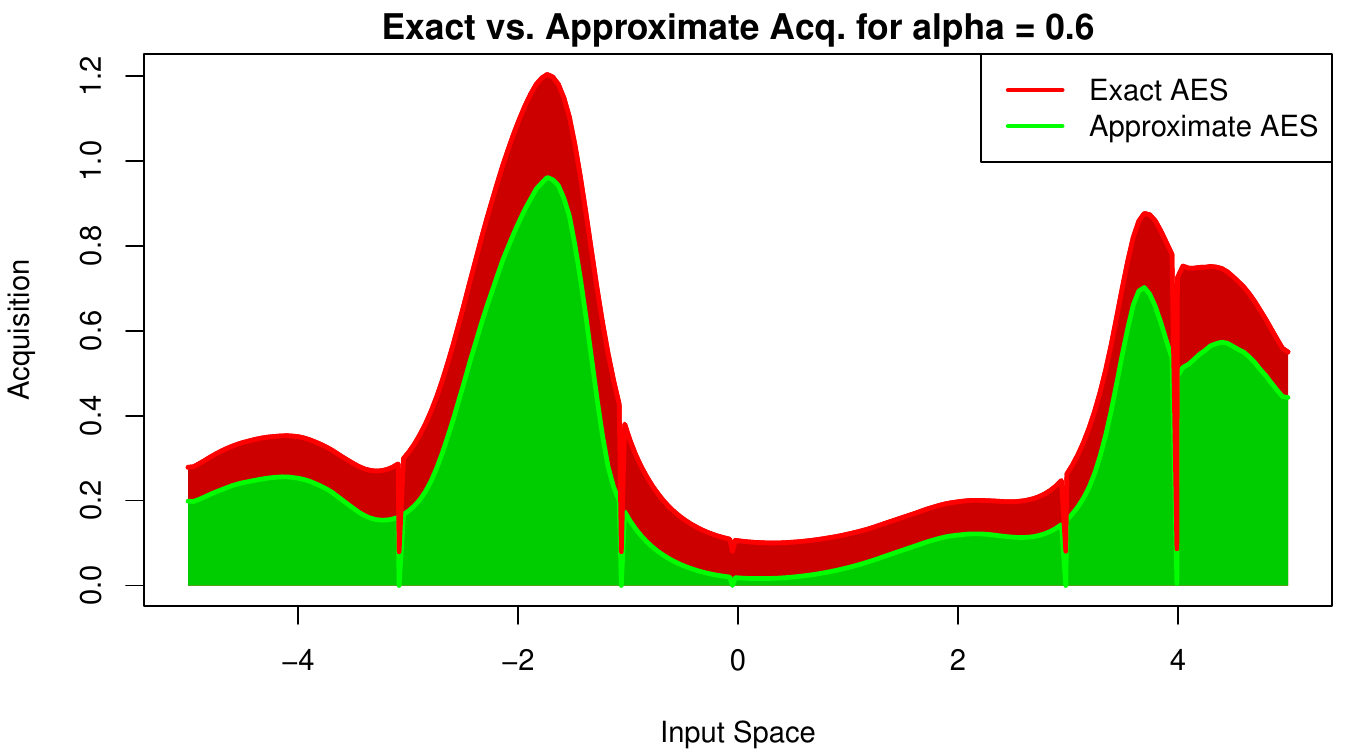} 
		\includegraphics[width=0.49\textwidth]{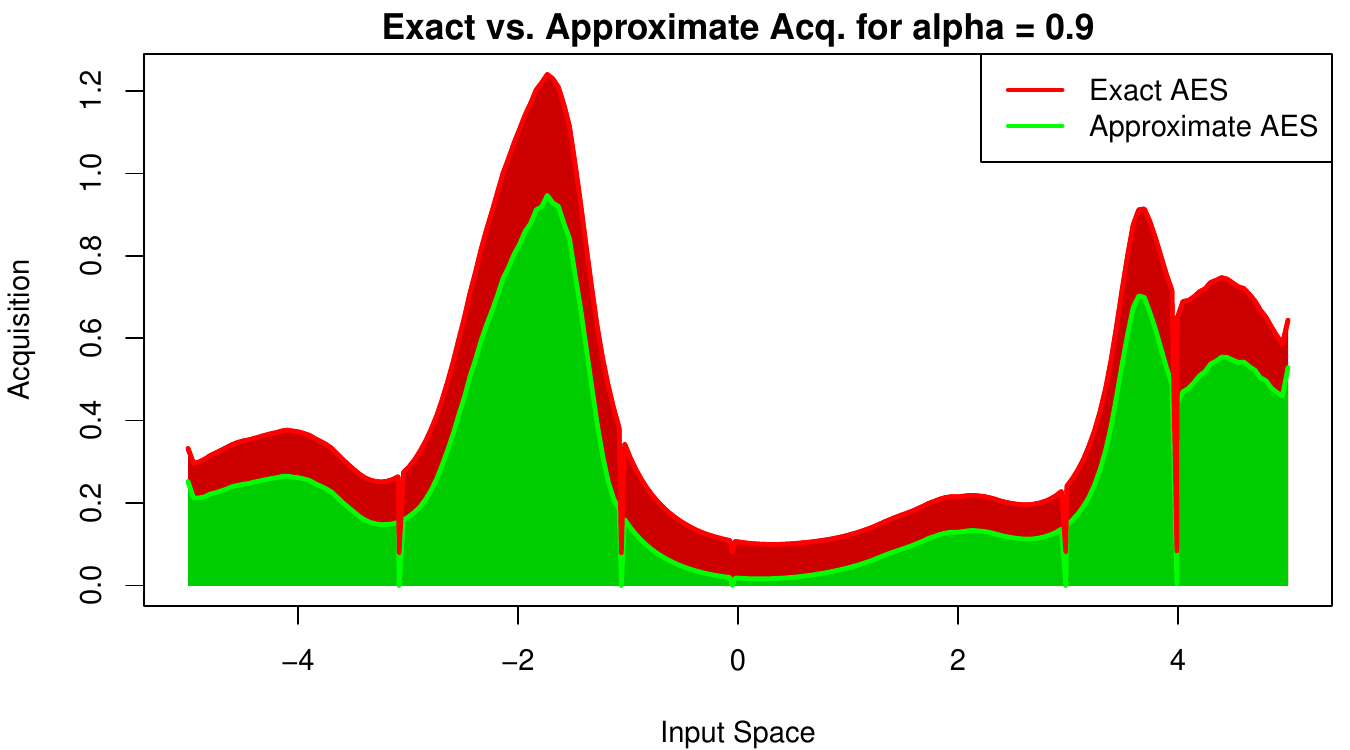} \\
		\includegraphics[width=0.49\textwidth]{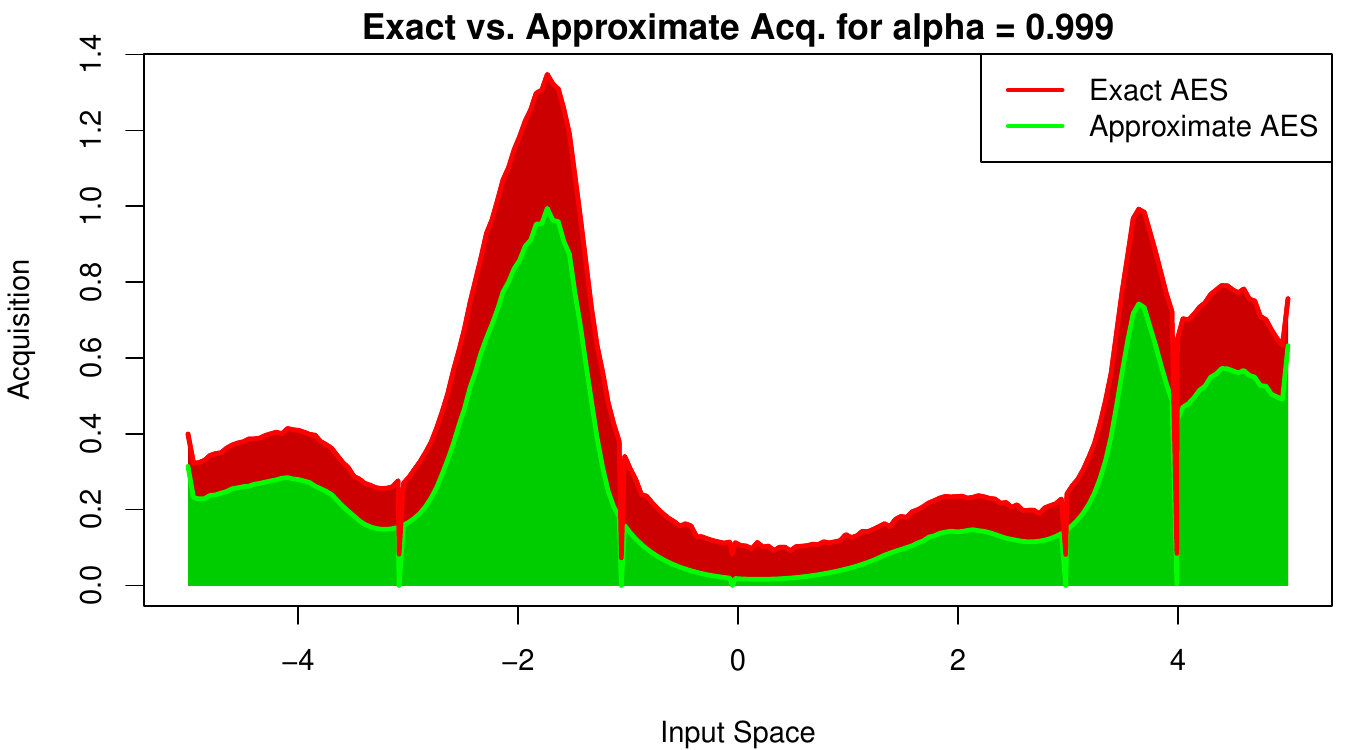}
		\includegraphics[width=0.49\textwidth]{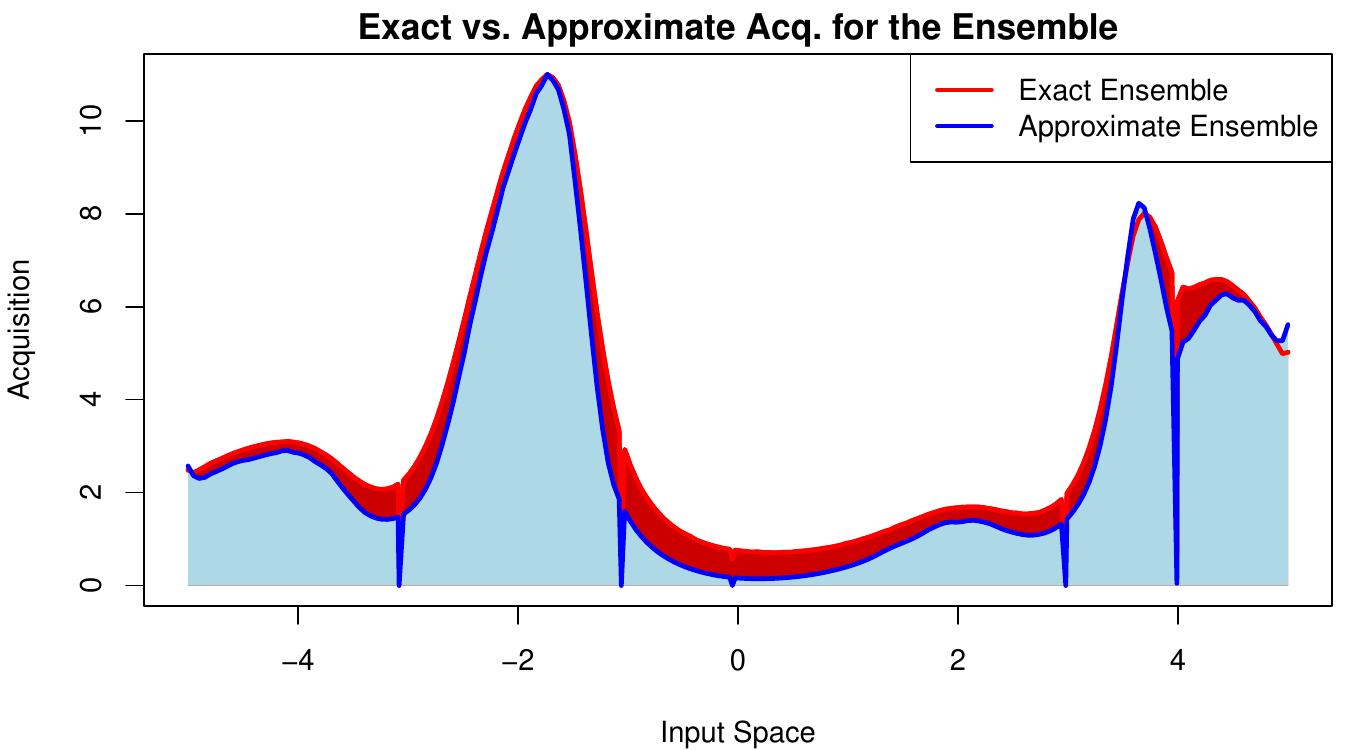}
	\end{tabular}
	\caption{
(top-left) GP predictive distribution for the objective.
From (top-right) to (bottom-left) acquisition function AES, when using the proposed approximation, and using a method that is expected to give the exact acquisition.
We report results for a representative set of $\alpha$ values. (bottom-right) Acquisition function for the ensemble method using the 
proposed approximation and a method that is expected to give the exact acquisition. Best viewed in color.
 }
	\label{FIG:EXPSYNTHE}
\end{figure*}

Figure \ref{FIG:EXPSYNTHE} (top-left) shows 
the GP predictive distribution for the objective and the observed data considered.
From (top-right) to (bottom-left) 
we compare, for a representative set of values for $\alpha$, our proposed approximation of 
the AES acquisition function with the exact acquisition estimated as described above.
We observe that the exact method and the proposed approximation are very similar and
have the local maxima near the same locations.
However, it is possible to observe that the proposed approximation tends to 
underestimate the exact acquisition function and that the approximation is worse
for values of $\alpha$ closer to zero. The under-estimation of the exact acquisition function 
has also been observed in other information-based strategies for BO such as PES \cite{hernandez2015predictive,hernandez2016predictive}.
We also observe that changing the value of $\alpha$ has an impact in the shape 
of the acquisition function in particular
regions of the input space (\emph{i.e.}, when $x > 4$).
Figure \ref{FIG:EXPSYNTHE} (bottom-right) compares the proposed approximation for the
ensemble acquisition function with the exact acquisition function.
We observe that the exact method and the proposed approximation are almost identical in this case, 
sharing the same local maxima and minima, with only small differences at particular points of the input 
space.

\subsection{Synthetic Experiments} \label{SUB:SYNEXP}

We carried out several synthetic experiments in which the objective is sampled from a GP and hence
there is no model bias. We consider 4 experiments with a different number of input dimensions.
Namely, 4, 6, 8, and 12 dimensions. In each experiment, we consider two scenarios: one with 
noiseless evaluations and another with evaluations contaminated by standard Gaussian noise 
with a variance of $0.1$. We consider $100$ repetitions of the 
experiments and report average results with the associated error bars.
We assess the performance of each method by measuring the relative difference
(on a logarithmic scale) between the recommendation's value in the noiseless
objective function and the optimal value, relative to the number of evaluations performed.
The optimal value for each problem is found via gradient optimization using a grid of size $D \times 10,000$ 
to choose the starting point, where $D$ is the dimension of the problem.

\begin{figure*}[tbh!]
	\begin{tabular}{cc}
		\includegraphics[width=0.49\textwidth]{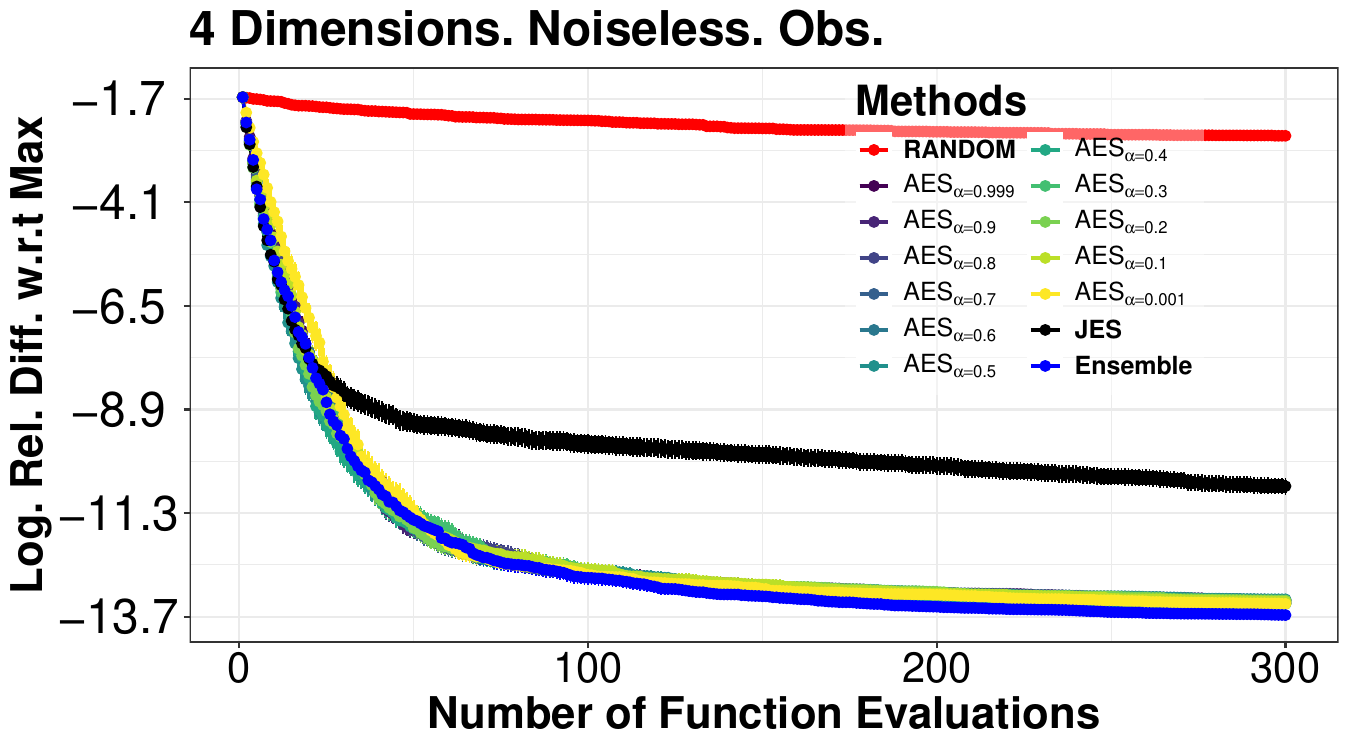}
		\includegraphics[width=0.49\textwidth]{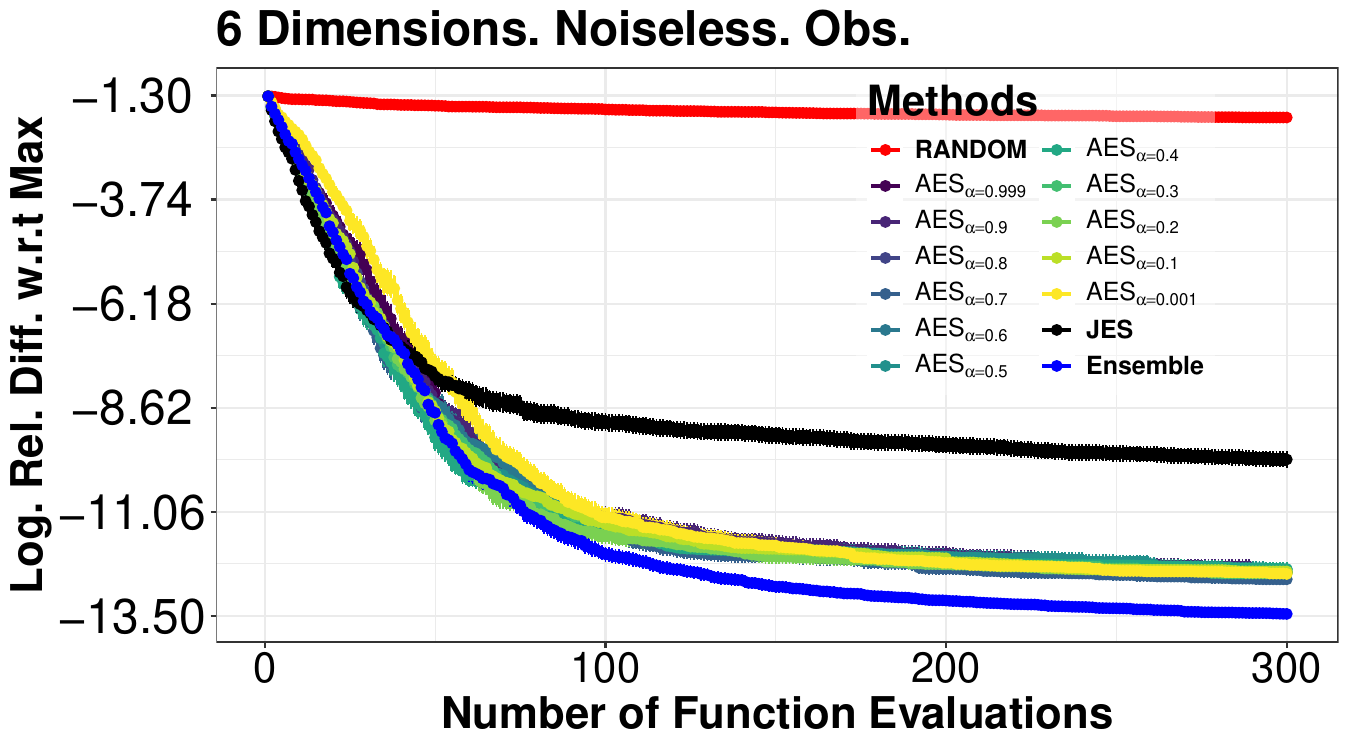} \\
		\includegraphics[width=0.49\textwidth]{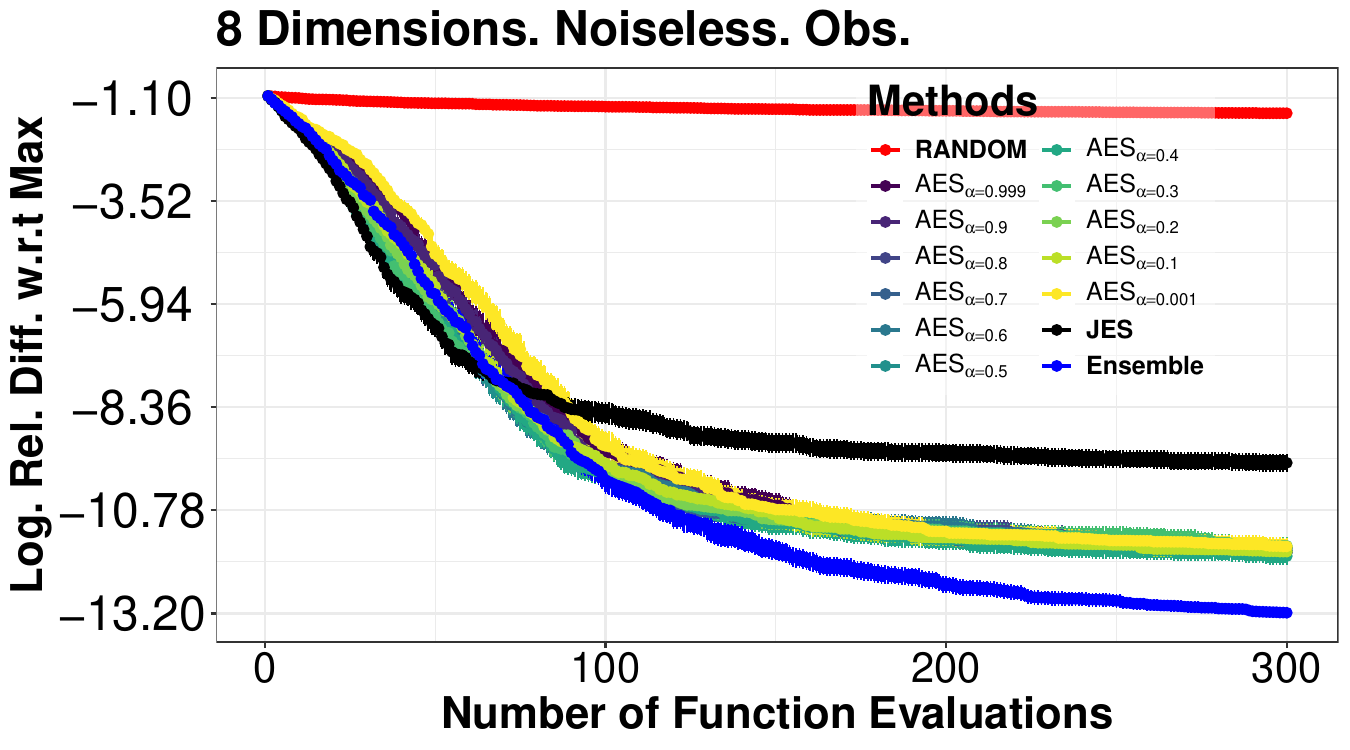}
		\includegraphics[width=0.49\textwidth]{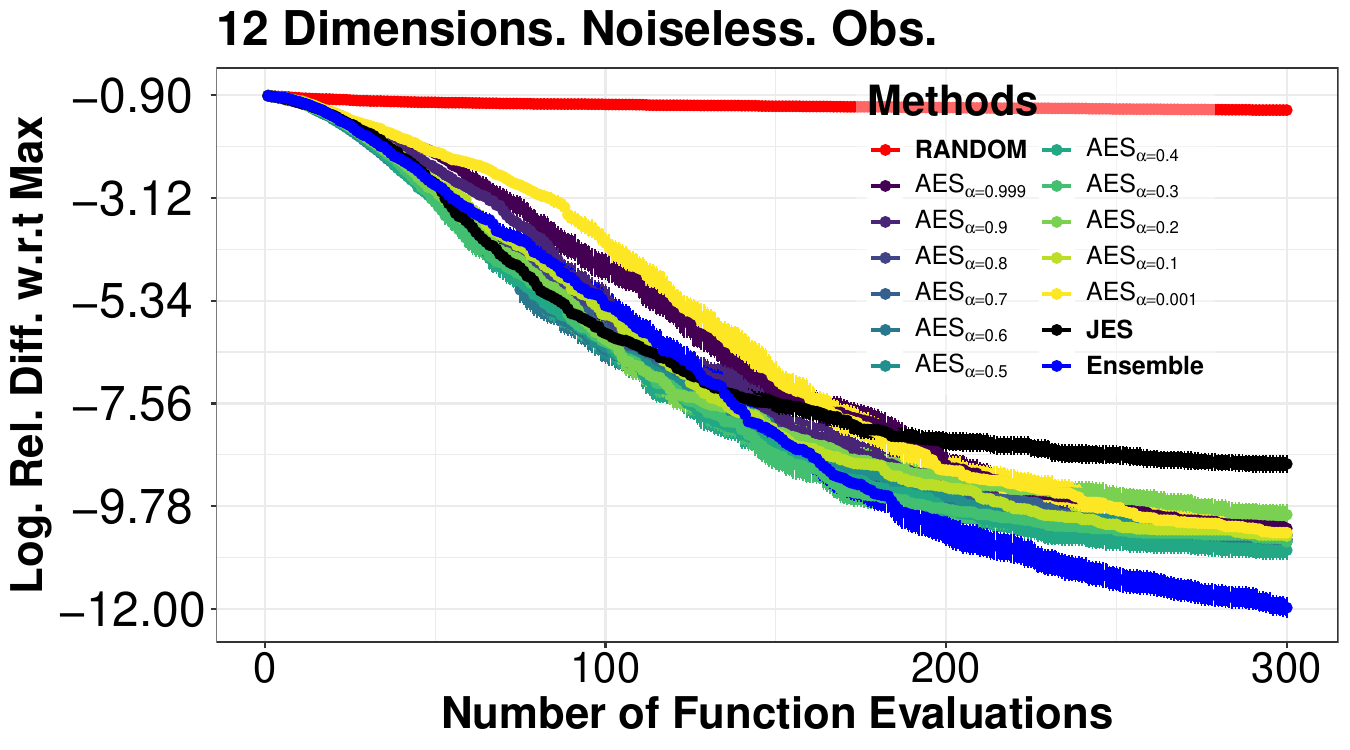}
	\end{tabular}
	\caption{ 
        Average logarithm relative difference between the objective at each method's
        recommendation and the objective at the global maximum, with respect to
        the number of evaluations. Results are shown for the 4, 6, 8,
        and 12 dimensional problems. Observations are noiseless. Best viewed in color.}
	\label{FIG:SYNNOISELESS}
\end{figure*}

Figure \ref{FIG:SYNNOISELESS} shows the results obtained by AES, for each value of $\alpha$ considered, and 
the ensemble method on the noiseless synthetic problems, for each number of inputs dimensions in 4, 6, 8, and 12.
We also report the results of JES and the random search strategy.
Among the BO methods, the ensemble method consistently
gives the best performance w.r.t. the number of evaluations performed, followed by AES
for different values of $\alpha$. By contrast, JES gives slightly worst 
results followed by the random search strategy, which is the worst over-all method. 
The fact that the ensemble method outperforms AES for
all values of $\alpha$ shows the benefits of the ensemble strategy. Specifically, combining the 
different acquisition functions for a range of values of $\alpha$ is better than using a 
single $\alpha$ value. We observe that in the 4-dimensional experiment, 
the difference in performance between the ensemble
method and the other AES variants is smaller. However, when the number of dimensions grows, these 
differences increase. We believe that the differences 
are small in the first problem because all methods 
reach the optimal solution. In the experiment with 6 input dimensions,
the differences w.r.t. the ensemble method become more significant.
Here, AES variants also outperform JES, but they convergence after 
approximately $300$ evaluations with minimal further improvement. 
By contrast, the ensemble method still keeps giving better solutions, on average,
with extra evaluations. In the experiment with 8 dimensions, a similar behavior is observed,
but the differences w.r.t. the ensemble method become larger.
Additionally, here, intermediate values of $\alpha$ within the range $0.2 \leq \alpha \leq 0.5$ perform
slightly better at the beginning than other AES variants.
Finally, in the 12-dimensional experiment, similar results are observed. The performance of 
AES with intermediate values of $\alpha$ gives better results at the beginning of the
optimization process and outperform JES. However, the ensemble method 
obtains over-all better results. In this problem all methods need more
evaluations to reach closer solutions to the global maximum.

\begin{figure*}[tbh!]
	\begin{tabular}{cc}
		\includegraphics[width=0.49\textwidth]{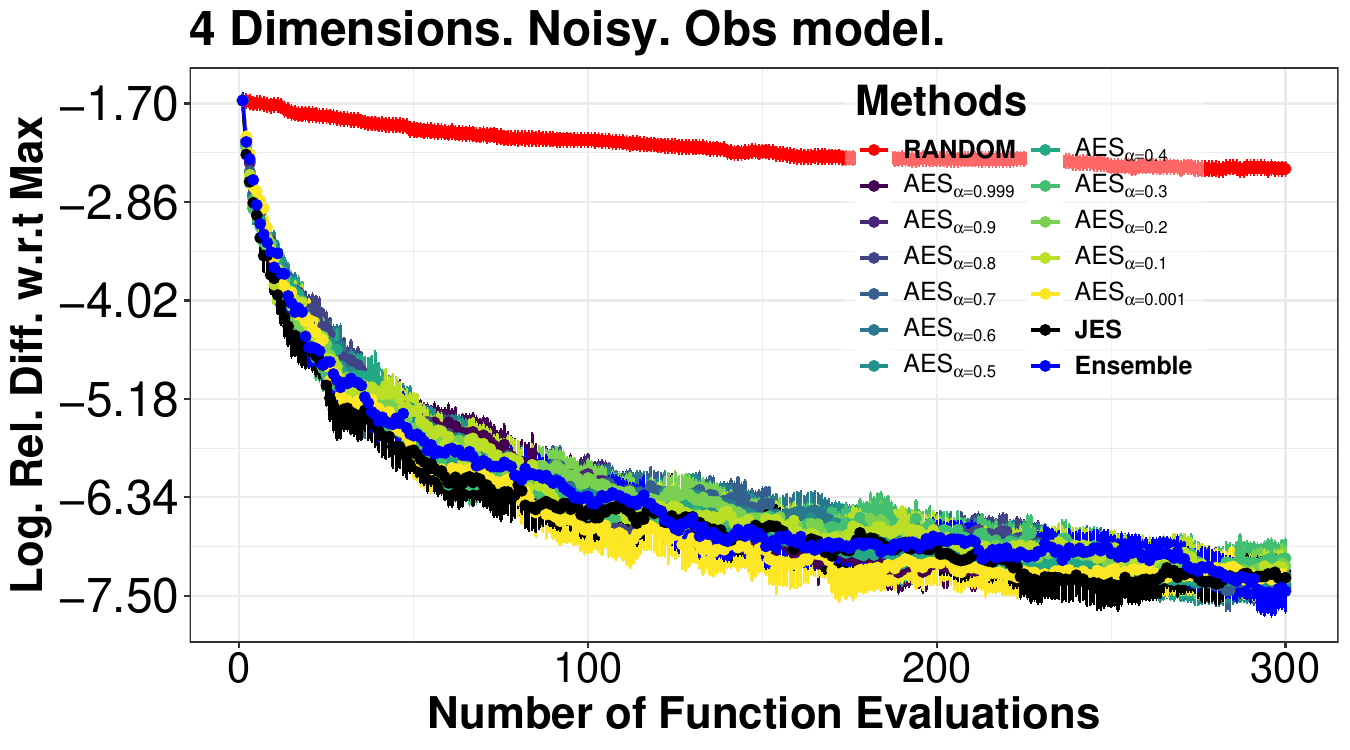}
		\includegraphics[width=0.49\textwidth]{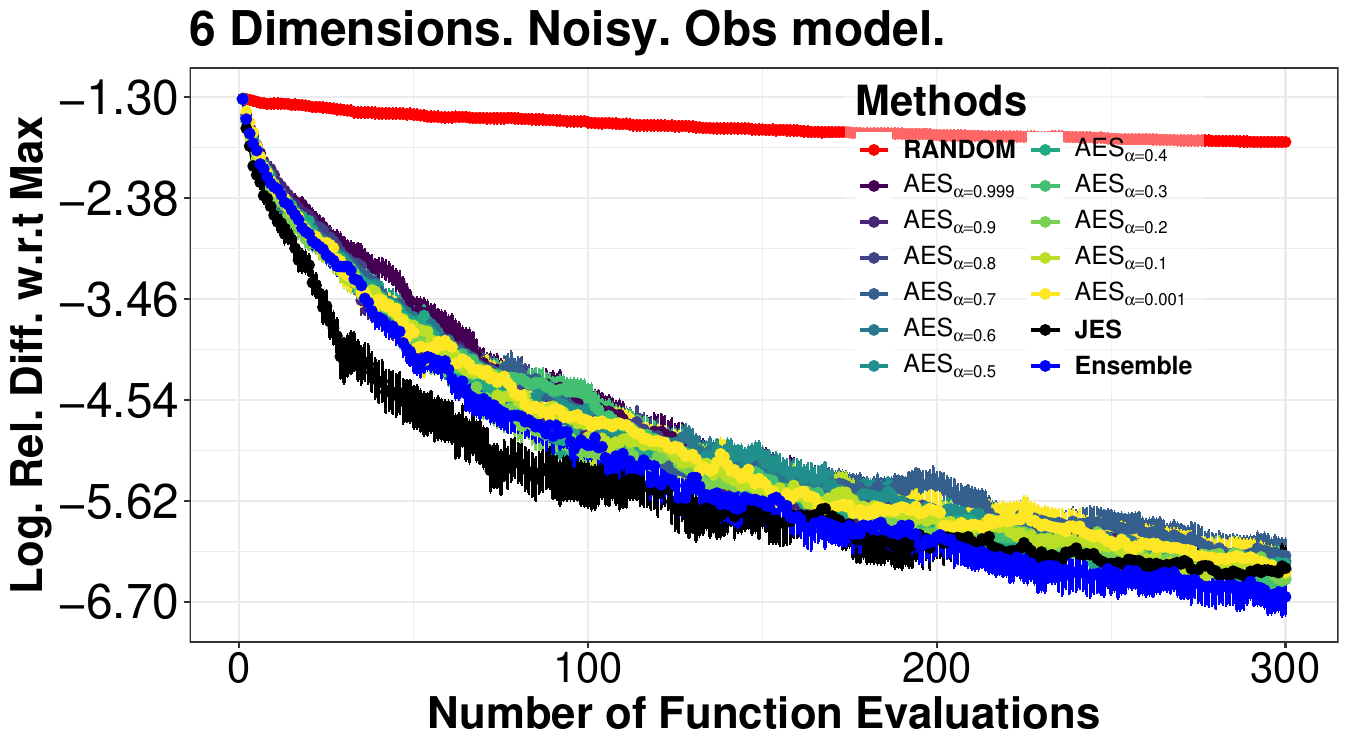} \\
		\includegraphics[width=0.49\textwidth]{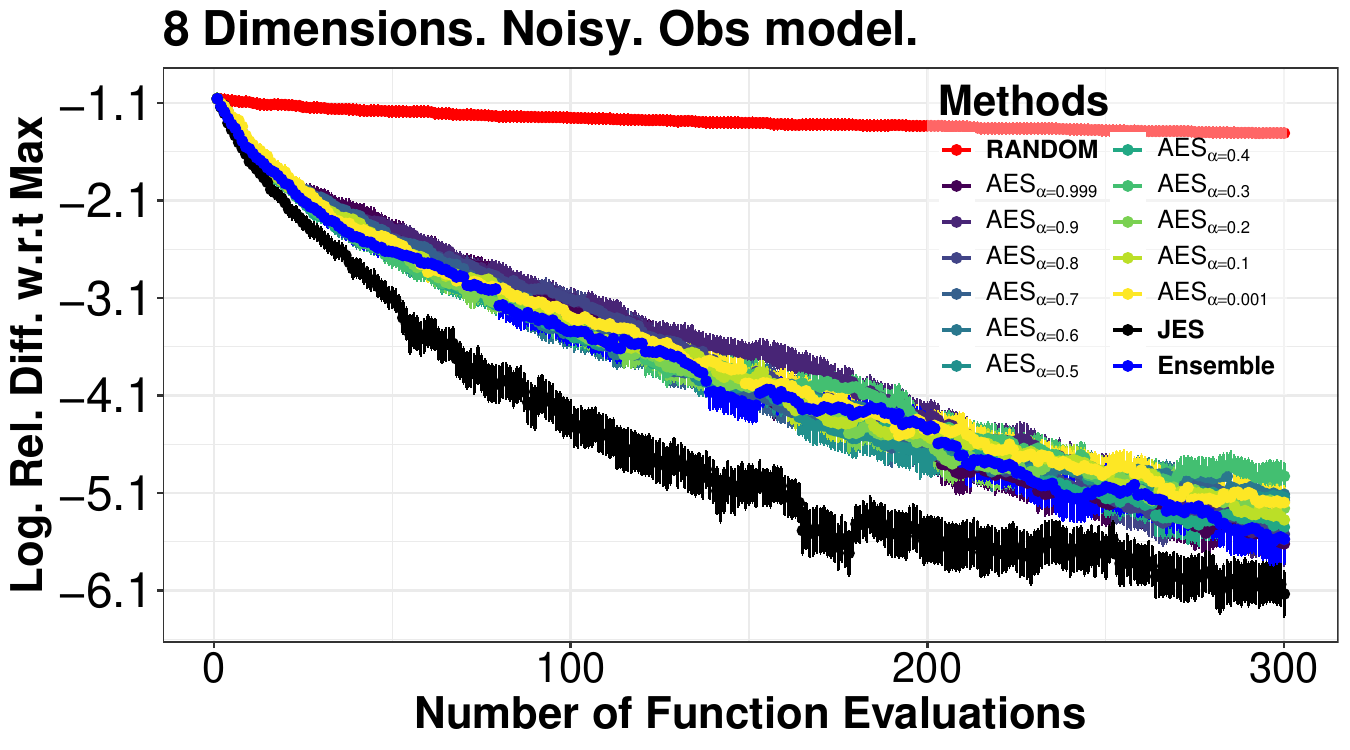}
		\includegraphics[width=0.49\textwidth]{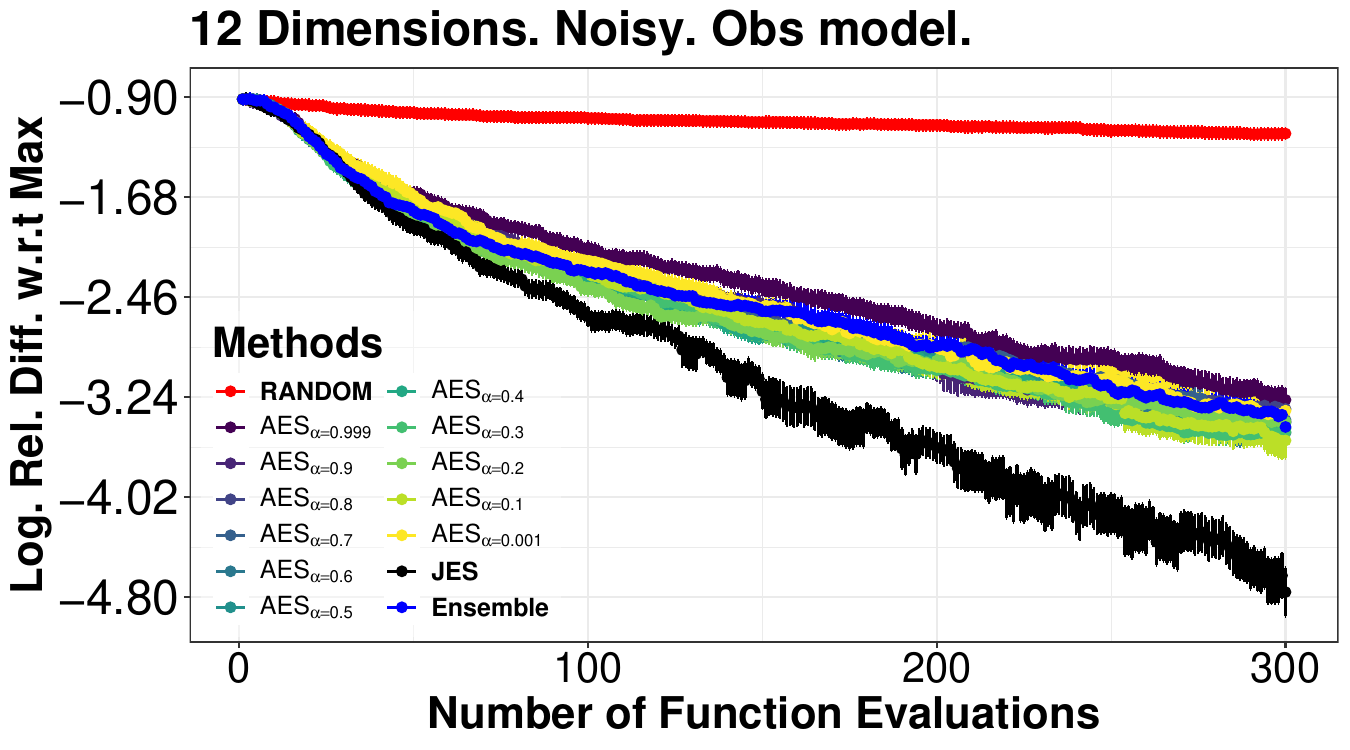}
	\end{tabular}
	\caption{ 
        Average logarithm relative difference between the objective at each method's
        recommendation and the objective at the global maximum, with respect to
        the number of evaluations. Results are shown for the 4, 6, 8,
        and 12 dimensional problems. Observations are noisy. Best viewed in color.}
	\label{FIG:SYNNOISY}
\end{figure*}

Figure \ref{FIG:SYNNOISY} shows the results of each method on the noisy evaluation setting, for 
each input dimensionality, \emph{i.e.}, 4, 6, 8, and 12 dimensions.
Again, random search exhibits the worst performance. Among the BO methods,
there are no significant differences in the problems with 4 and 6 dimensions. 
However, as the number of dimensions increases, JES performs better than the other strategies. 
Additionally, no single $\alpha$ value in AES is consistently better, and the ensemble method
performs similarly to the other $\alpha$-divergence based methods. We believe that these results could 
be explained because in the noisy setting, the beneficial properties of the ensemble method are
smaller and the performance of JES is not impaired by local maxima in the acquisition function.
See Appendix \ref{SEC:AP5} for further details.

These synthetic experiments illustrate the beneficial properties of the ensemble method 
which considers a range of values for $\alpha$ when compared to the AES method that exclusively
considers a particular value of $\alpha$.
Specifically, the ensemble method always performs similar or better than AES. 
Therefore, in the remaining experiments, we will consider 
exclusively the ensemble method.

\subsection{Impact of the Number of Samples}

We investigate the impact of the number of samples $S$ considered in the performance of our 
ensemble method when approximating the expectation 
in (\ref{EQ:aES_acq_approx}). For this, we consider the 4-dimensional and the 
8-dimensional synthetic problems described before. 
We report the performance of the ensemble method for increasing values of $S$, from $1$ to $64$. Recall that in such 
a method the generated samples are re-used for each value of $\alpha$ considered. That is, we only
generate $S$ samples once, instead of $S$ samples for each value of $\alpha$. 
We compare results with JES for the same number of generated samples $S$.
We consider both a noiseless and a noisy evaluation scenario.

\begin{figure*}[tbh!]
	\begin{tabular}{cc}
		\includegraphics[width=0.49\textwidth]{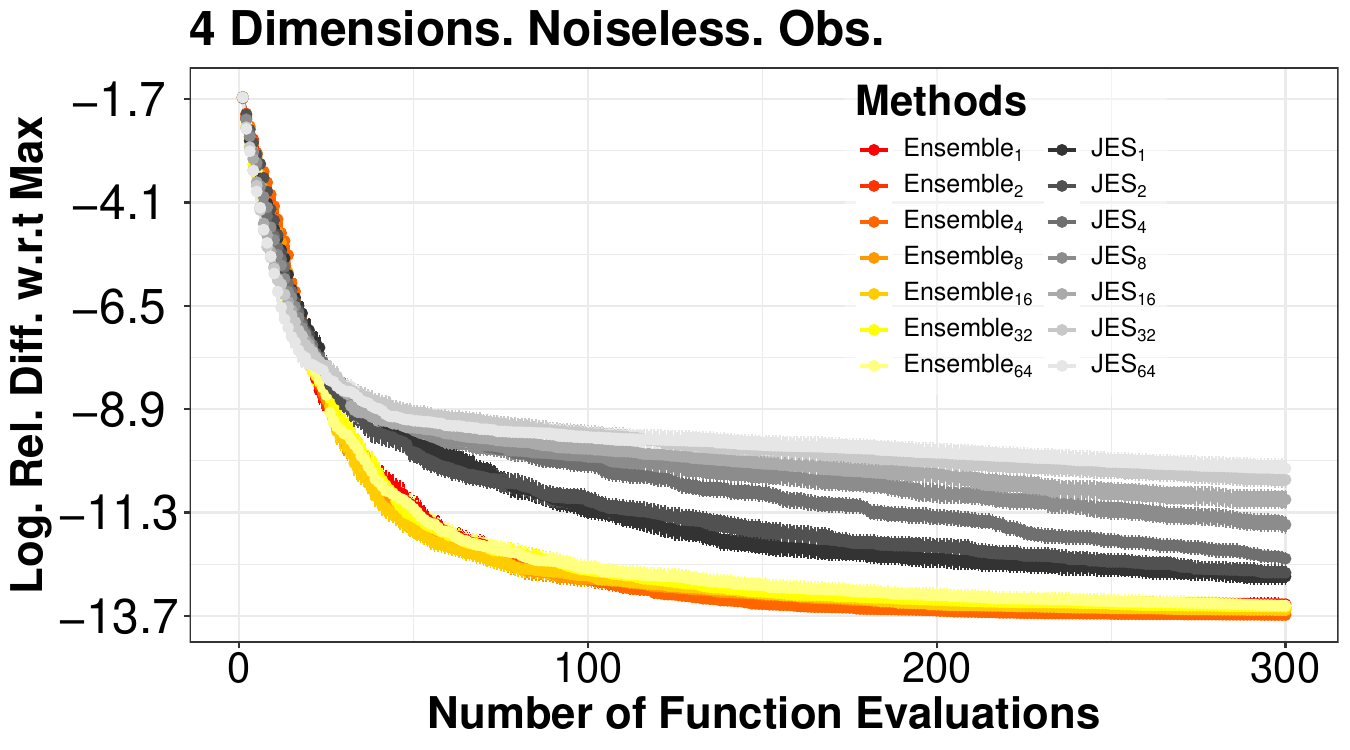}
		\includegraphics[width=0.49\textwidth]{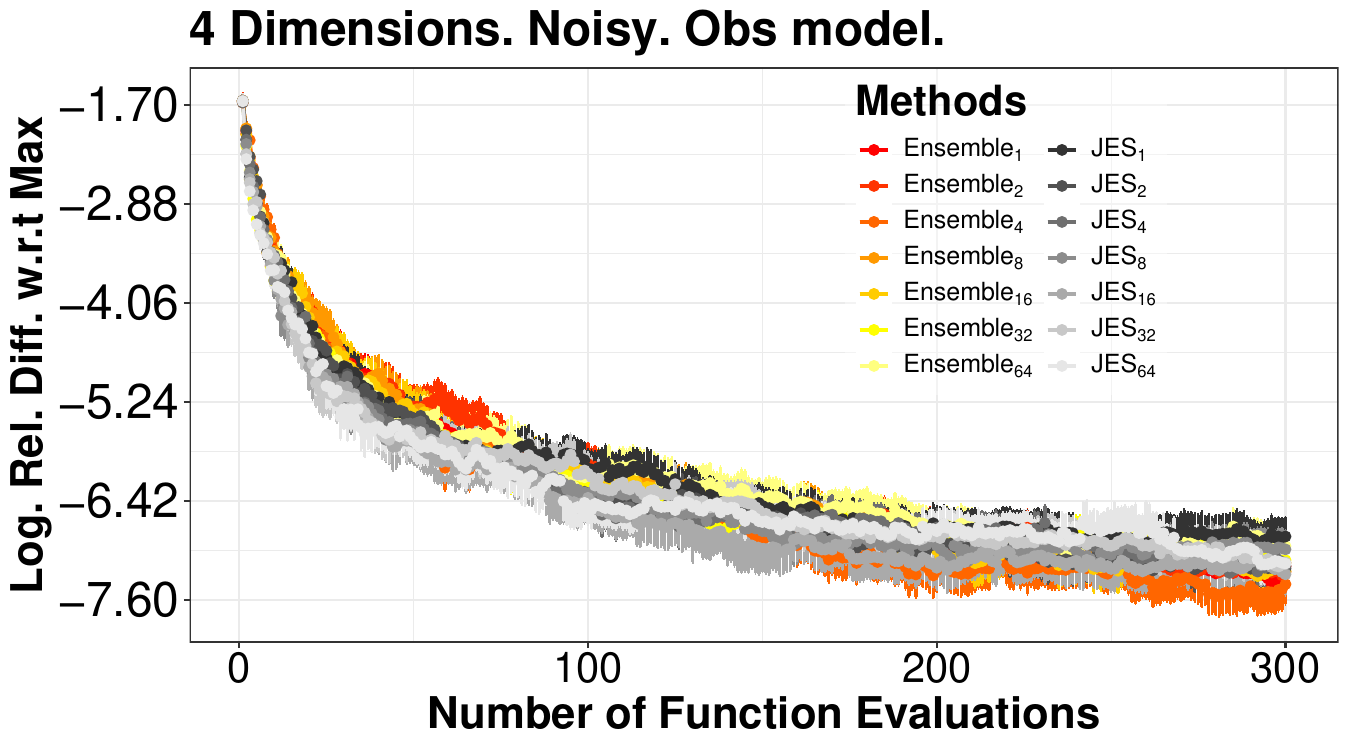} \\
		\includegraphics[width=0.49\textwidth]{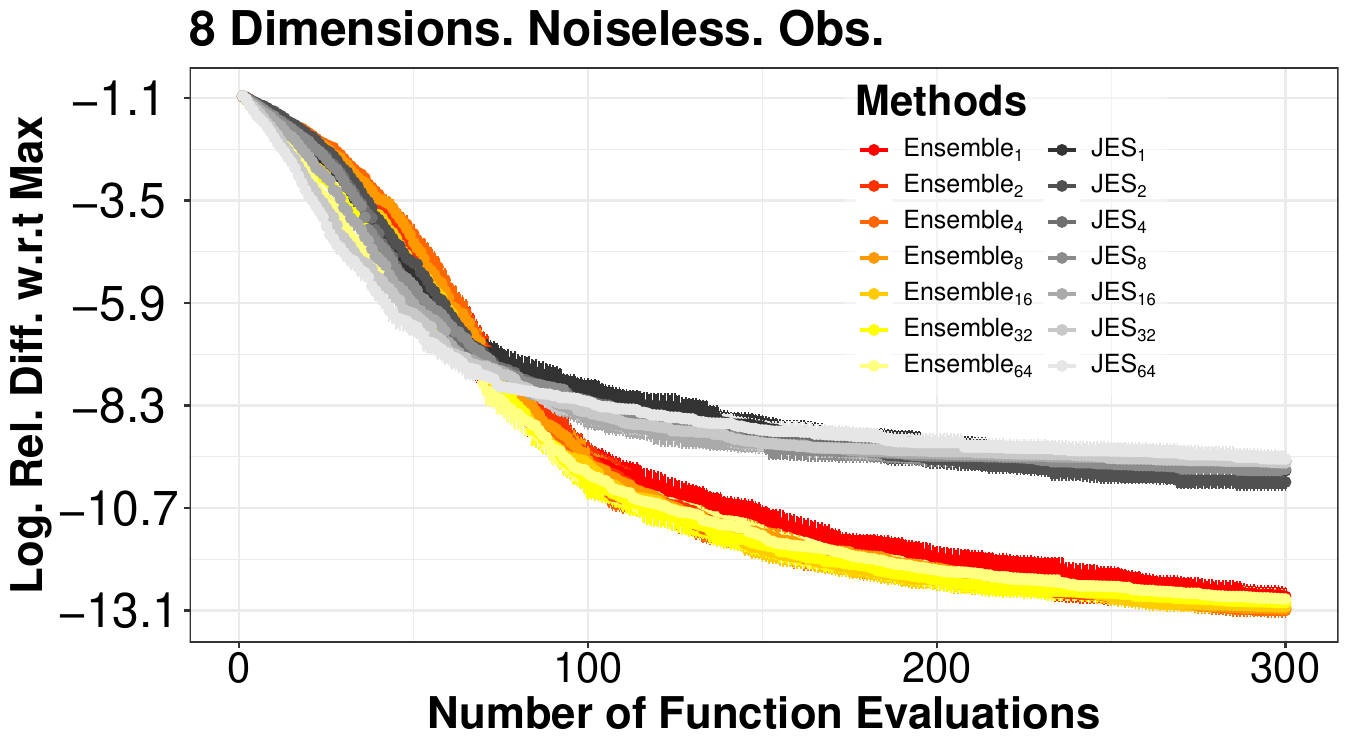}
		\includegraphics[width=0.49\textwidth]{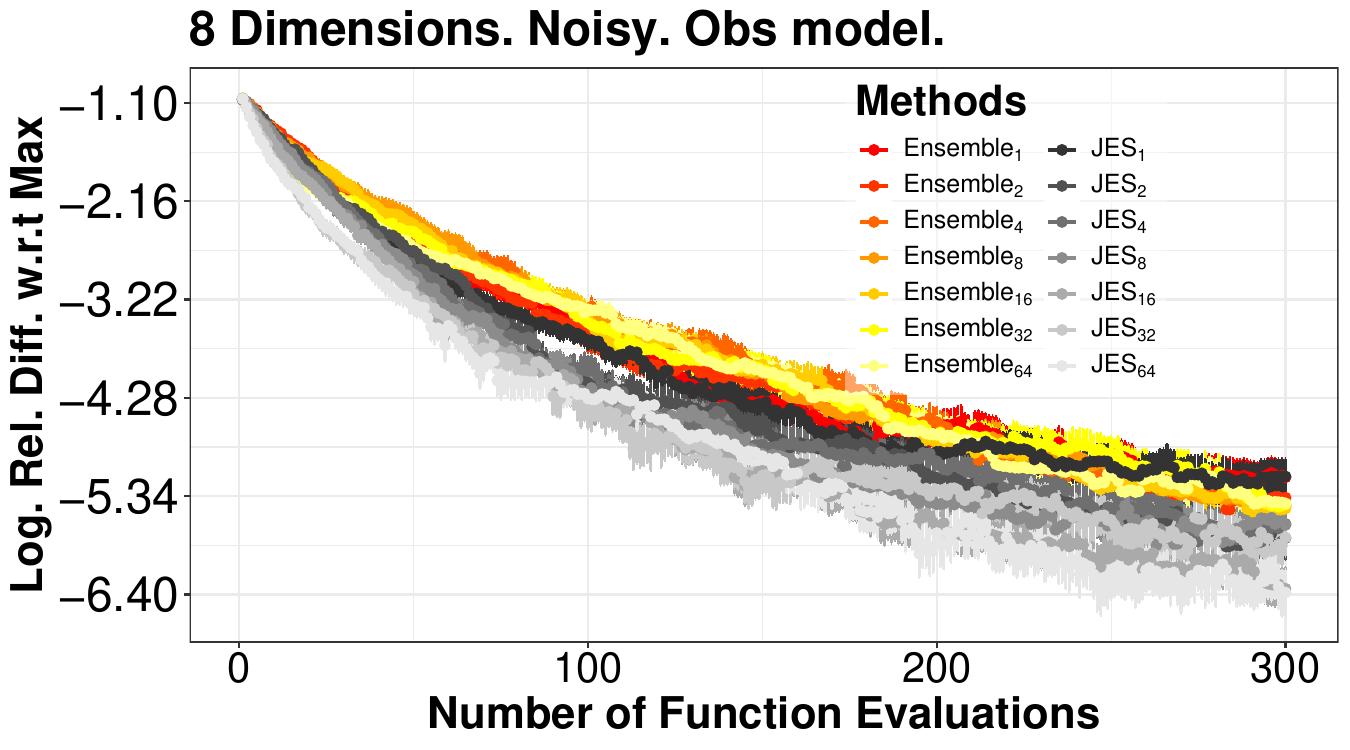} 
  
	\end{tabular}
	\caption{ 
        Average logarithm relative difference between the objective at each method's
        recommendation and the objective at the global maximum, with respect to
        the number of evaluations. Results are shown for the 4, and 8
        dimensional problems. The number of samples $S$ ranges from $1$ to $64$.
	Best viewed in color.}
	\label{FIG:SAMPLES}
\end{figure*}

Figure \ref{FIG:SAMPLES} shows the results obtained for each method and 
problem considered.  Remarkably, in the 4-dimensional problem and the noiseless setting, the performance of 
JES deteriorates as the number of samples increases, whereas the
performance of the ensemble method remains very similar, independently of the number 
of samples considered. This behavior of JES is mitigated in the 8-dimensional problem, 
where varying the number of samples yields similar results for all values of $S$. 
In the noisy evaluation setting, the performance of the ensemble method is also little dependent on the number
of samples considered. However, the behavior of JES is a bit different and increasing $S$ 
improves the results. 

We believe that the phenomenon in which the performance of JES deteriorates as the number of samples $S$ increases,
is related to the large number of local minima in the JES acquisition, as described in Section \ref{SUB:RESACQ}, and illustrated in
Figure \ref{fig:comparison_alpha_values}. Specifically, as $S$ increases, JES generates more local maxima in the acquisition 
function in the noiseless setting.  A large number of local maxima makes more likely that the optimization
of the acquisition function is trapped in a sub-optimal solution. This will enforce JES to explore more 
extensively the input space. This behavior is good in high-dimensional problems, where extensive exploration 
is necessary. However, in low-dimensional problems, this excess exploration prevents JES from adequately
exploiting solutions in promising regions. By contrast, increasing the number of samples
does not have this detrimental effect on the ensemble method. Since the number of local minima in the acquisition 
of the ensemble method is smaller in the noiseless setting, its trade-off between exploration 
and exploitation is expected to be less affected by $S$.  In noisy problems, the conditional distribution 
$p(y|\mathcal{D}_{t-1},\mathbf{x},\{y^\star, \mathbf{x}^\star\})$ does not have zero variance at the sampled solutions.
Therefore, JES does not have that many local optima and the aforementioned behavior does not happen.

Summing up, in the ensemble method $S$ has little effect on the final performance. By contrast, in 
JES, increasing $S$ slightly deteriorates or gives similar results in the noiseless evaluation setting.
In the noisy evaluation setting, increasing $S$ slightly improves the results of JES 
in high-dimensional problems.

\subsection{Benchmark Experiments} \label{SUB:BENEXP}

In the previous sections, we considered that the objective function is generated
from a GP. This means that there is no model bias and the probabilistic model used 
in the BO loop can perfectly fit the objective. In general, the objective need not 
be generated from a prior GP and model bias can have an effect on the performance.
Therefore, in this section, we consider several benchmark functions that are often used
in optimization problems and that are not generated from a GP prior. Namely, 
Hartmann-3D, Hartmann-6D, Styblinski-Tang-4D, and Cosine-8D \cite{jamil2013literature, yang2010engineering}.
Note that each objective function is followed by the number of dimensions it depends on.

Using the aforementioned objectives, we carry out experiments and
compare the performance of the ensemble method with that of JES, PES, MES, and a 
random search strategy. We exclude AES for specific values of $\alpha$ because 
the ensemble method consistently achieved equal or better results in the synthetic 
experiments. As before, we consider both noiseless and noisy scenarios, contaminating
observations in the noisy setting with additive Gaussian noise with variance $0.1$.
Again, we measure the performance of each method in terms of the relative
difference (in a logarithmic scale) between the objective at the recommendation
and the global maximum. We report average results over $100$ repetitions of the 
experiments in which the initial observations differ.

\begin{figure*}[tbh!]
	\begin{tabular}{cc}
		\includegraphics[width=0.49\textwidth]{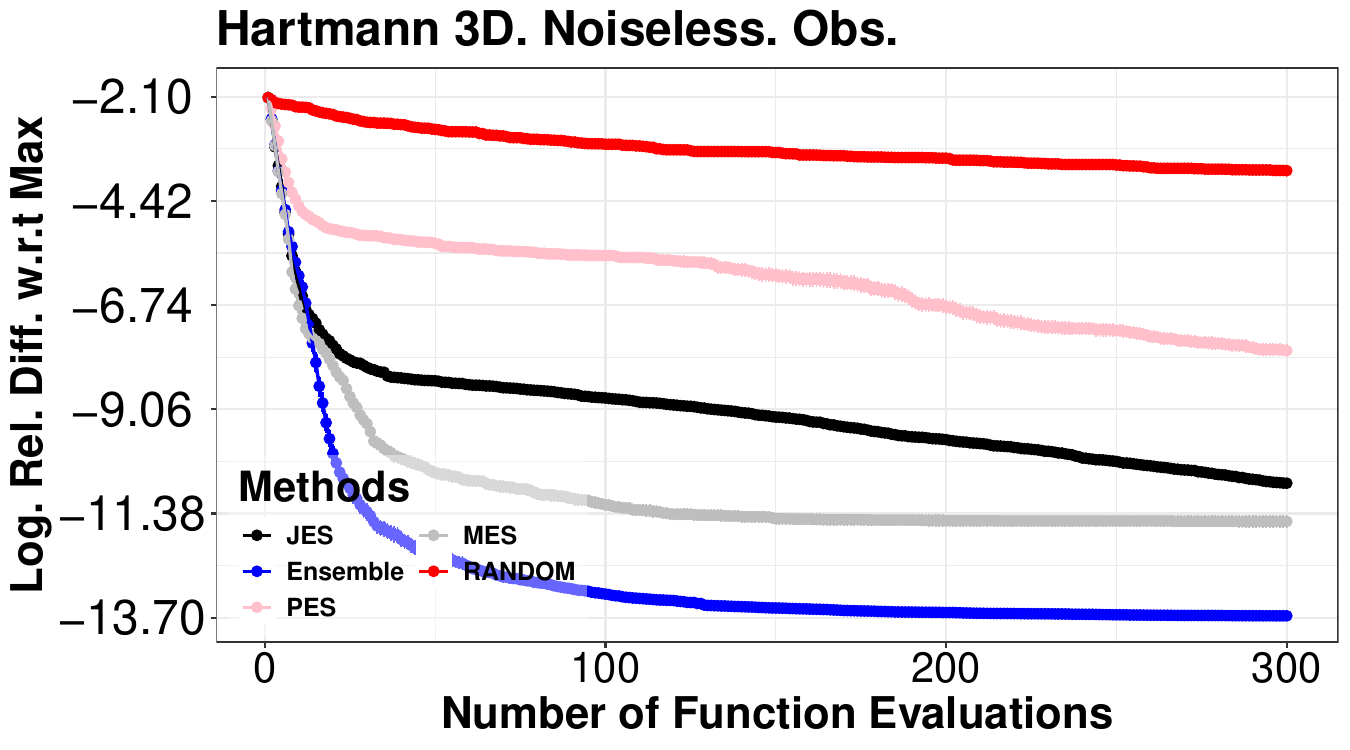}
		\includegraphics[width=0.49\textwidth]{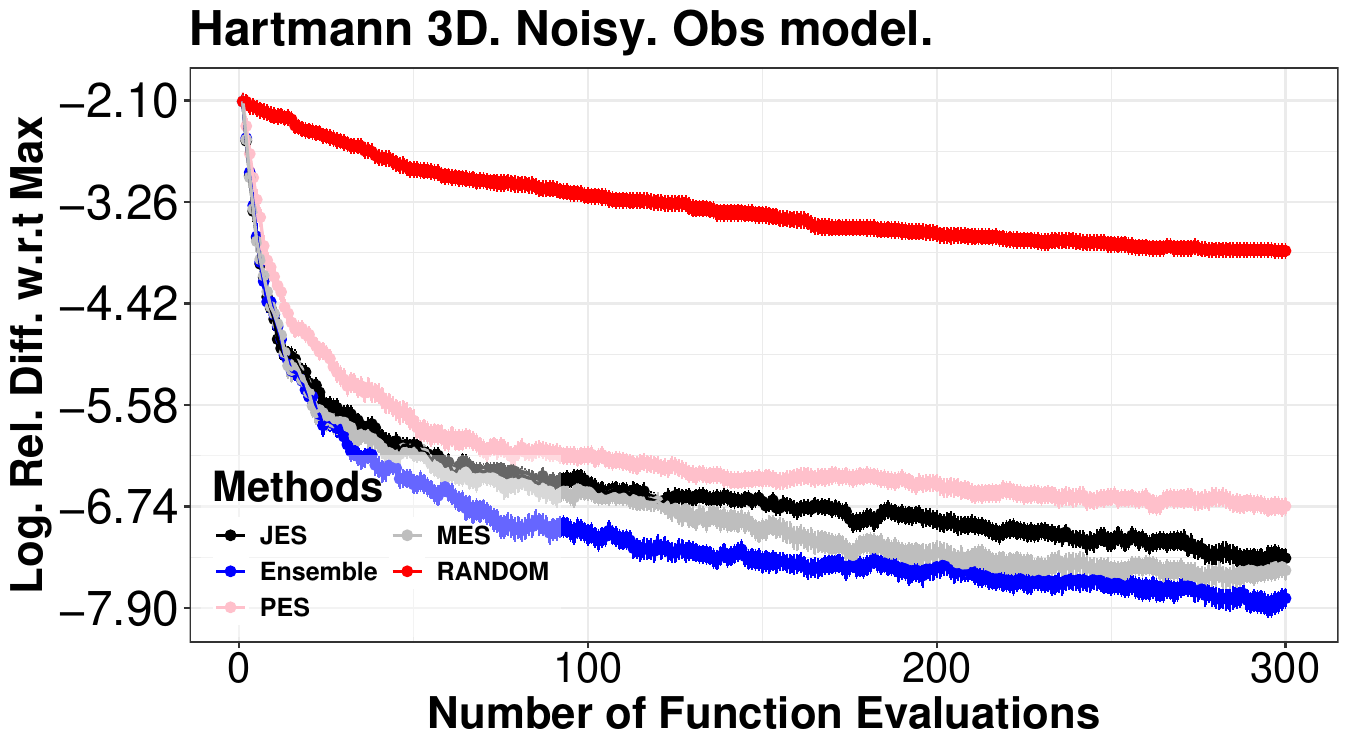} \\
		\includegraphics[width=0.49\textwidth]{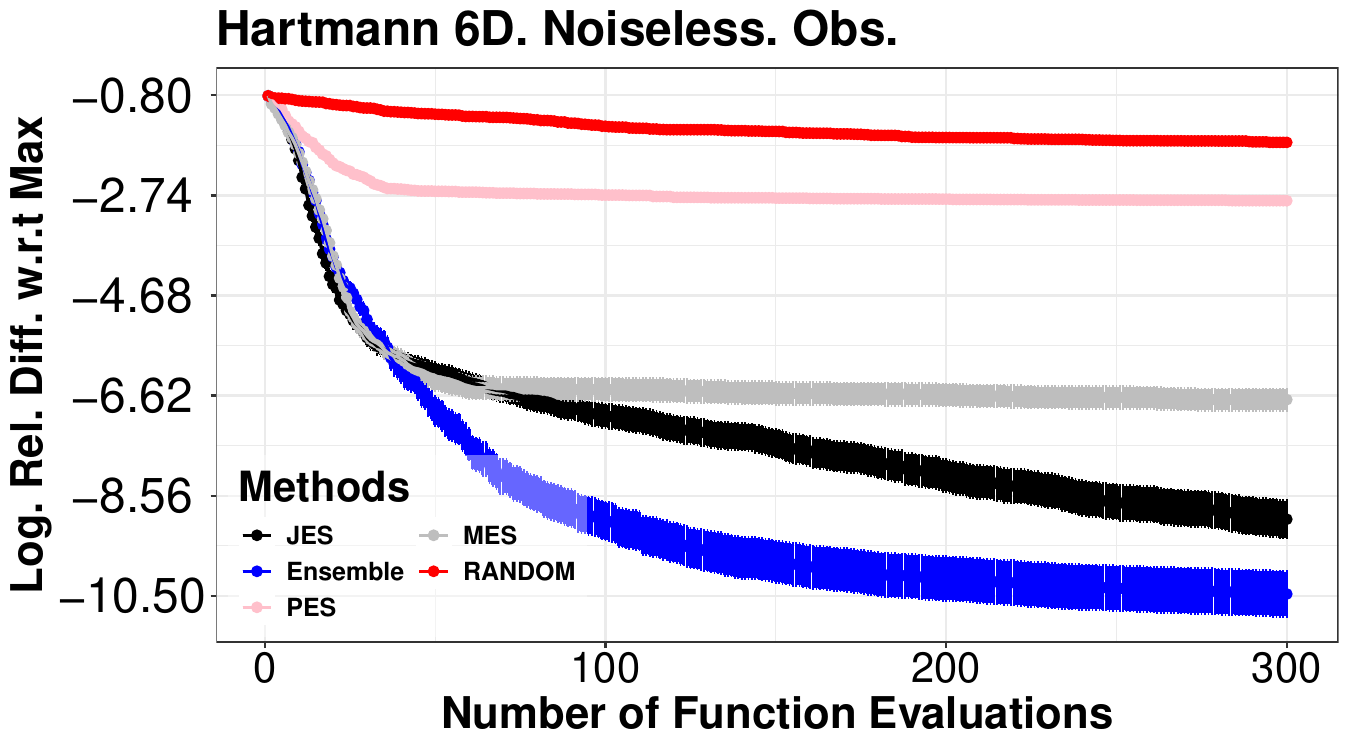}
		\includegraphics[width=0.49\textwidth]{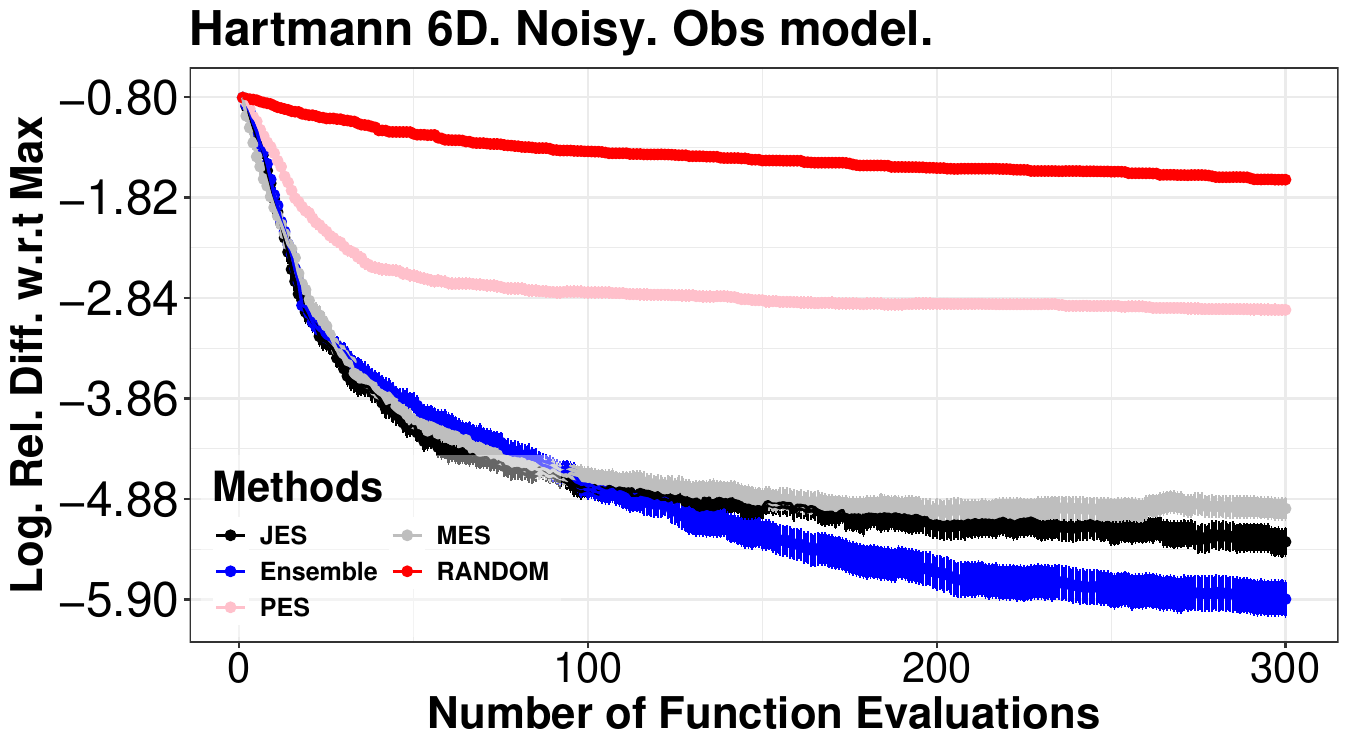} \\
		\includegraphics[width=0.49\textwidth]{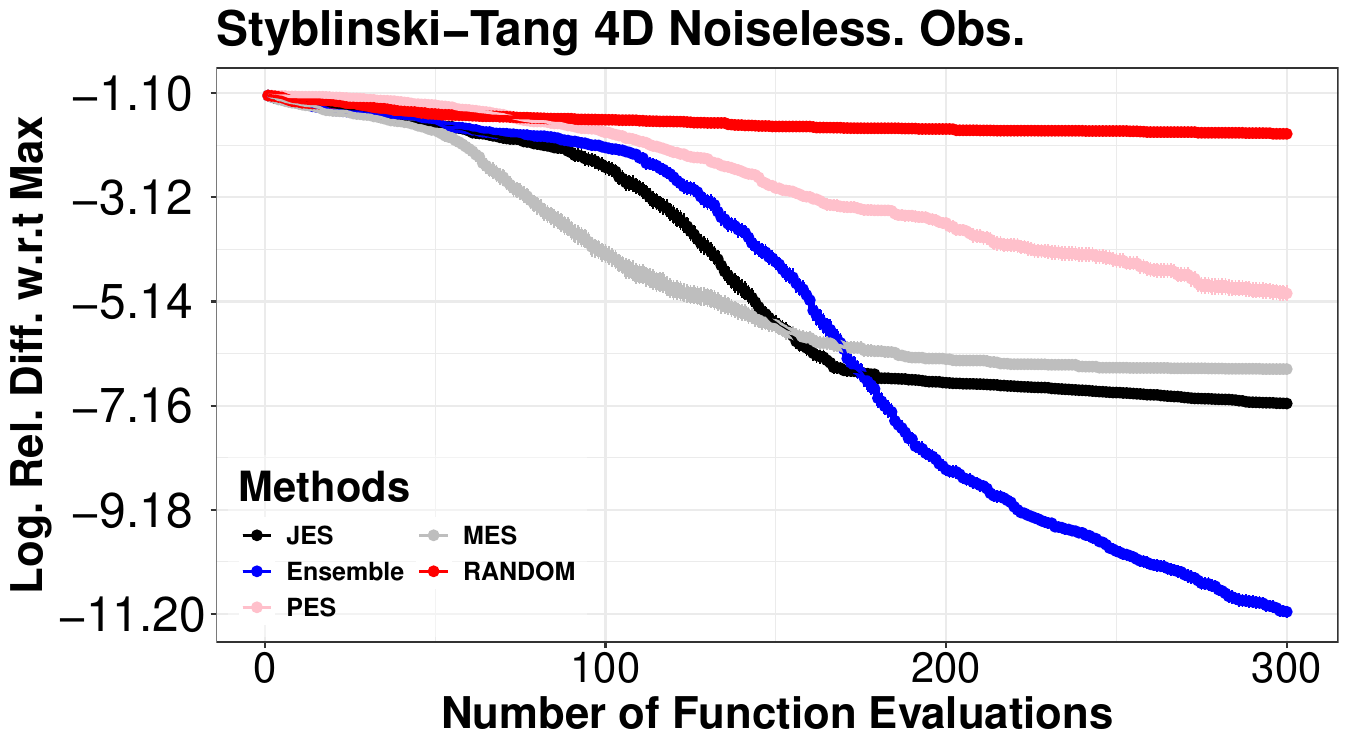}
		\includegraphics[width=0.49\textwidth]{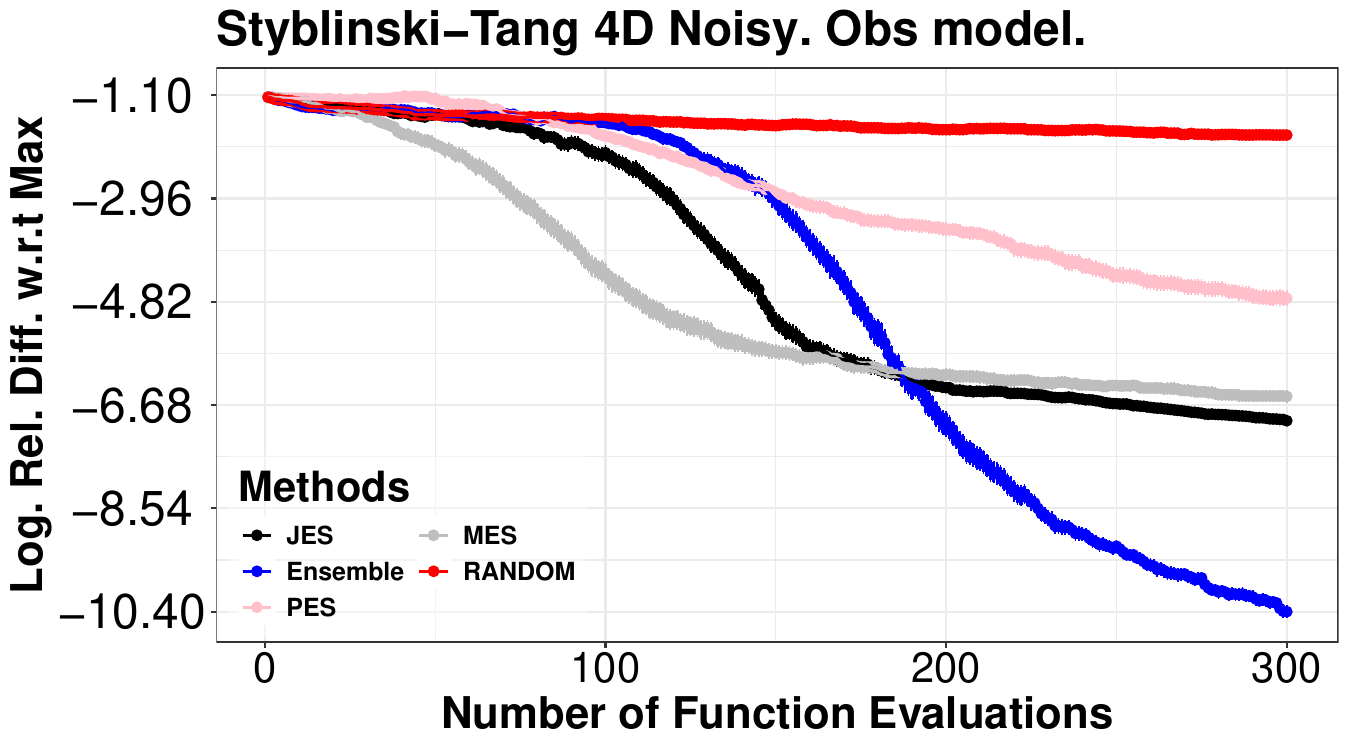} \\
		\includegraphics[width=0.49\textwidth]{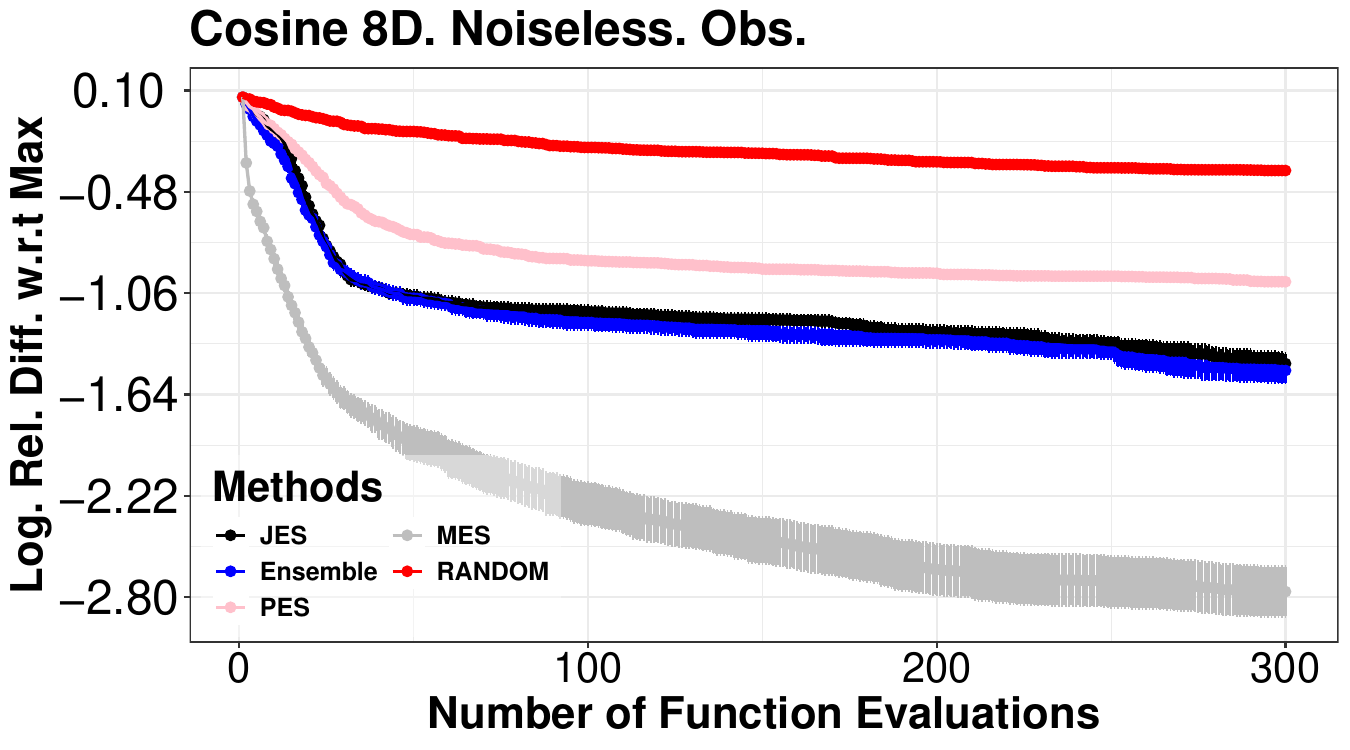}
		\includegraphics[width=0.49\textwidth]{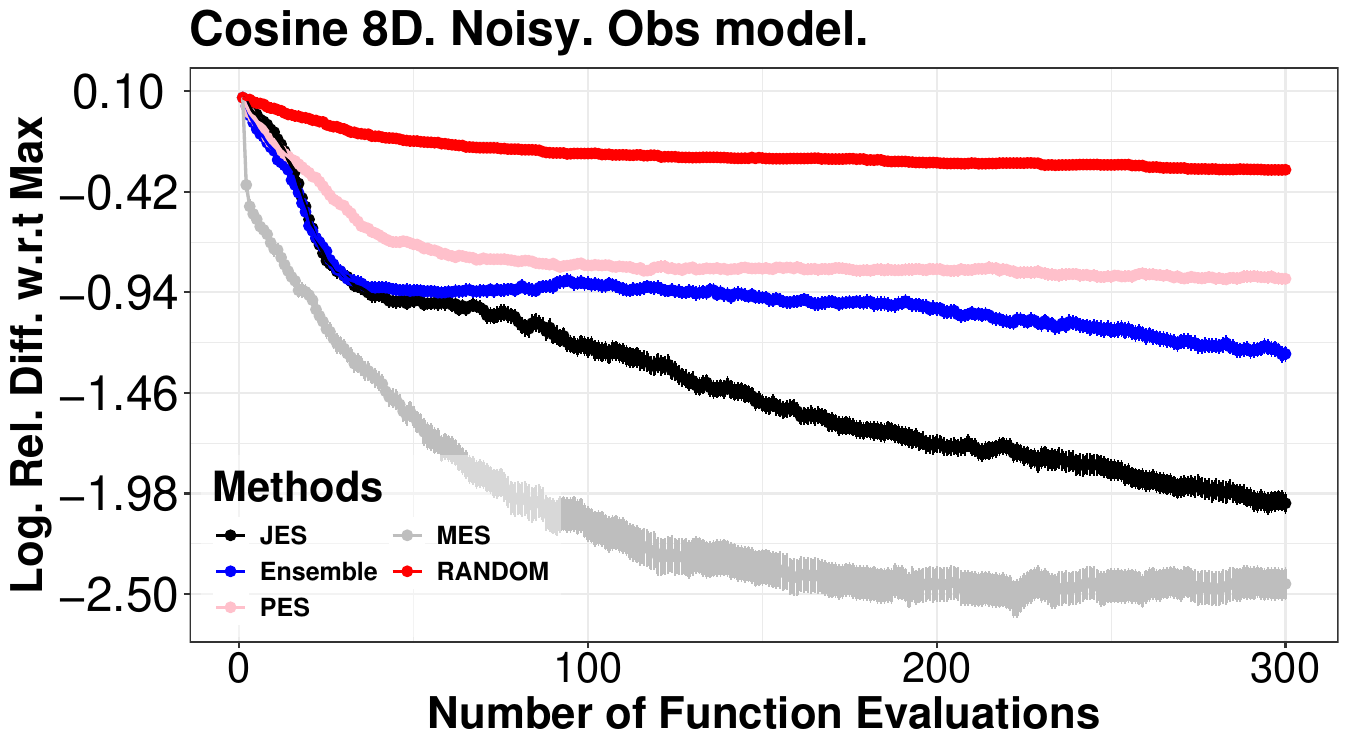}
	\end{tabular}
	\caption{ 
        Average log hyper-volume relative difference between
        the recommendation of each method and the maximum hyper-volume,
        with respect to the number of evaluations made. We show
        the results for the 3-dimensional Hartman problem,
        the 6-dimensional Hartman problem,
        the 4-dimensional Styblinski-Tang problem and
        the 8-dimensional Cosine problem.
        We consider noiseless (left-column) and noisy observations (right-column).
        Best seen in color. }
	\label{FIG:BENCHMARKS}
\end{figure*}

The results obtained are displayed in Figure \ref{FIG:BENCHMARKS}. The figure 
shows that the ensemble method achieves the best performance in 3 of the 4 objectives
considered. In the noiseless setting, the ensemble method is always equal or better 
than JES. In the noisy setting, JES performs better than the ensemble method in Cosine-8D.
In this problem the GP model does not perform very well since the obtained solutions
are the furthest away from the optimum among the 4 problems considered.
Here, MES is the best performing method both in the noiseless and the noisy setting.
Random is the over-all worst performing method in each problem and setting, followed
by PES. The summary of these experiments is that the ensemble method performs very well
in the noiseless evaluation setting and is competitive with state-of-the-art methods
for BO based on information theory. In the noisy evaluation setting the benefits
of using the ensemble method are smaller. These results are compatible with the ones
observed in the previous sections.

\subsection{Real world Experiments} \label{SUB:MLPEXP}

We evaluate the performance of the ensemble method, JES, PES, MES and random search in real-world 
experiments. For this, we consider tuning five hyper-parameters of a neural network for classification with 
two hidden layers. The hyper-parameters considered are the number of hidden units in each layer, 
the batch size, the amount of $\ell_2$ regularization, the learning rate, and the number of 
training epochs. Again, we do compare results with AES for specific values of $\alpha$ since the ensemble method
performs similarly or better in synthetic experiments. The network's accuracy is estimated using 5-fold 
cross-validation, and the optimization process is run for 200 iterations. As in the previous experiments,
we report the performance of each method as the relative difference (in a logarithmic scale) between the objective 
at the recommendation, and the best objective value observed (across each method). We report average 
results across 100 repetitions of the experiments. We consider 6 different classification datasets 
extracted from the UCI repository \cite{ucirepository}. Namely, Pima, Image, Defects, Liver, Australian, and Ionosphere. 
Table \ref{tab:datasets} shows the characteristics of these datasets.

\begin{table}[!htb]
	\begin{center}
	\caption{Characteristics of the UCI datasets employed in the experiments. } 
	\label{tab:datasets}
	\begin{tabular}{lccc}
	\hline
		{\bf Dataset } & {\bf \# Instances} & {\bf \# Features} & {\bf \# Classes} \\
	\hline
		Pima & 768 & 8 & 2\\   
		Image & 2310 & 19 & 7 \\   
		Defects & 1109 & 21 & 2 \\   
		Liver & 583 & 10 & 2 \\   
		Australian & 690 & 14 & 2 \\   
		Ionosphere & 351 & 34 & 2 \\   
	\hline
	\end{tabular}
	\end{center}
\end{table}

\begin{figure*}[tbh!]
	\begin{tabular}{cc}
		\includegraphics[width=0.49\textwidth]{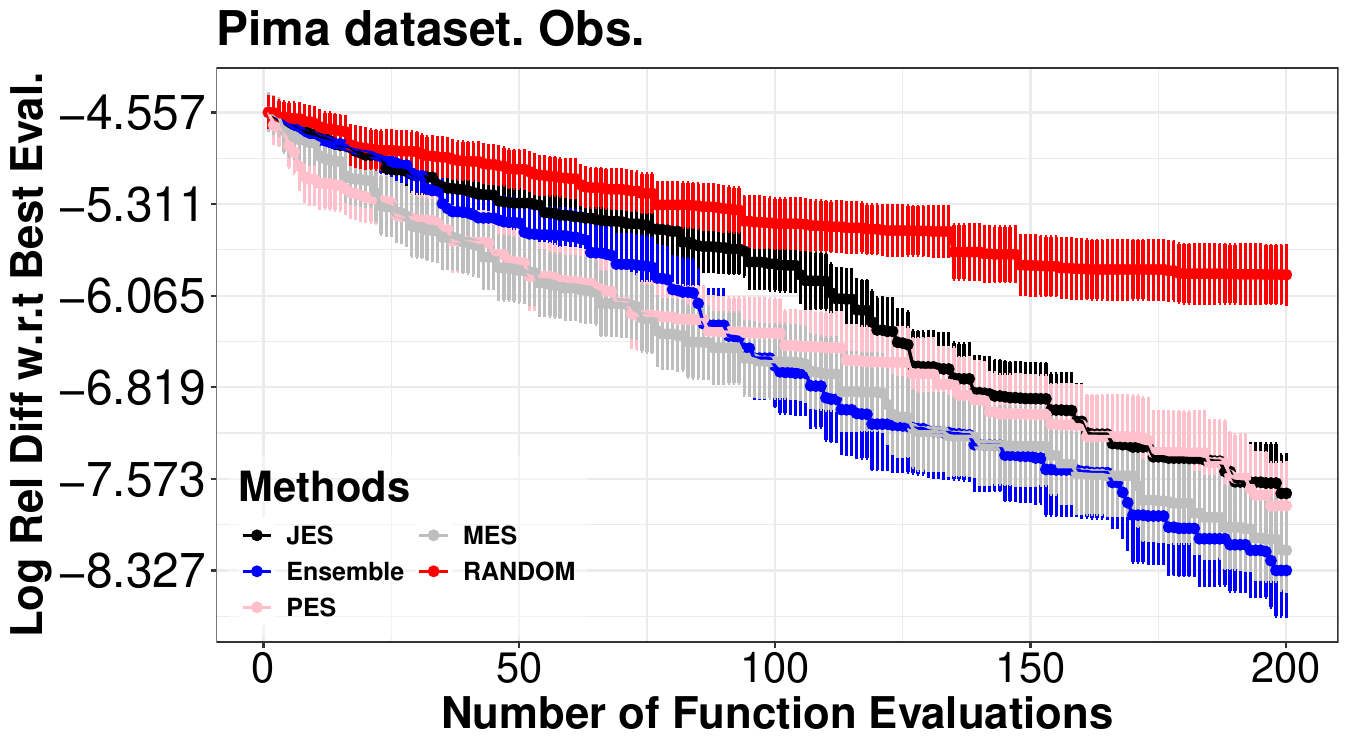}
		\includegraphics[width=0.49\textwidth]{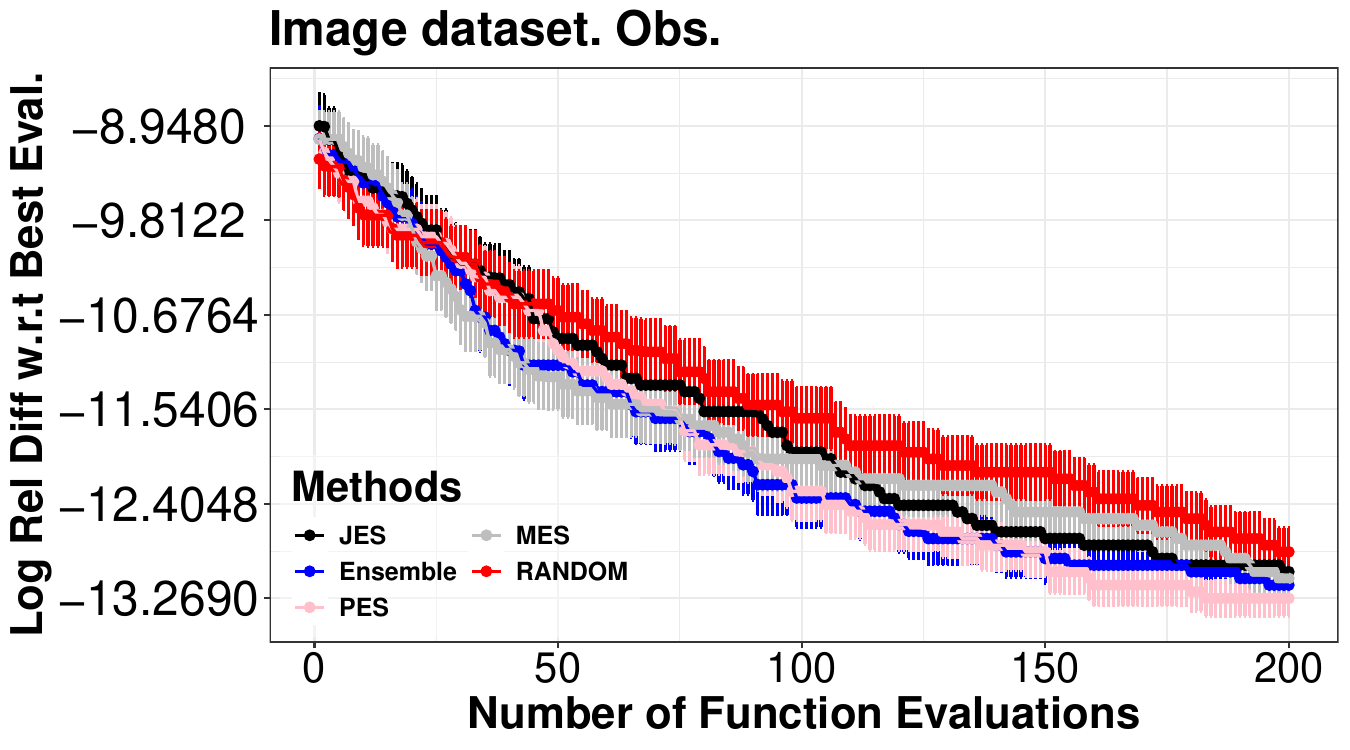} \\
		\includegraphics[width=0.49\textwidth]{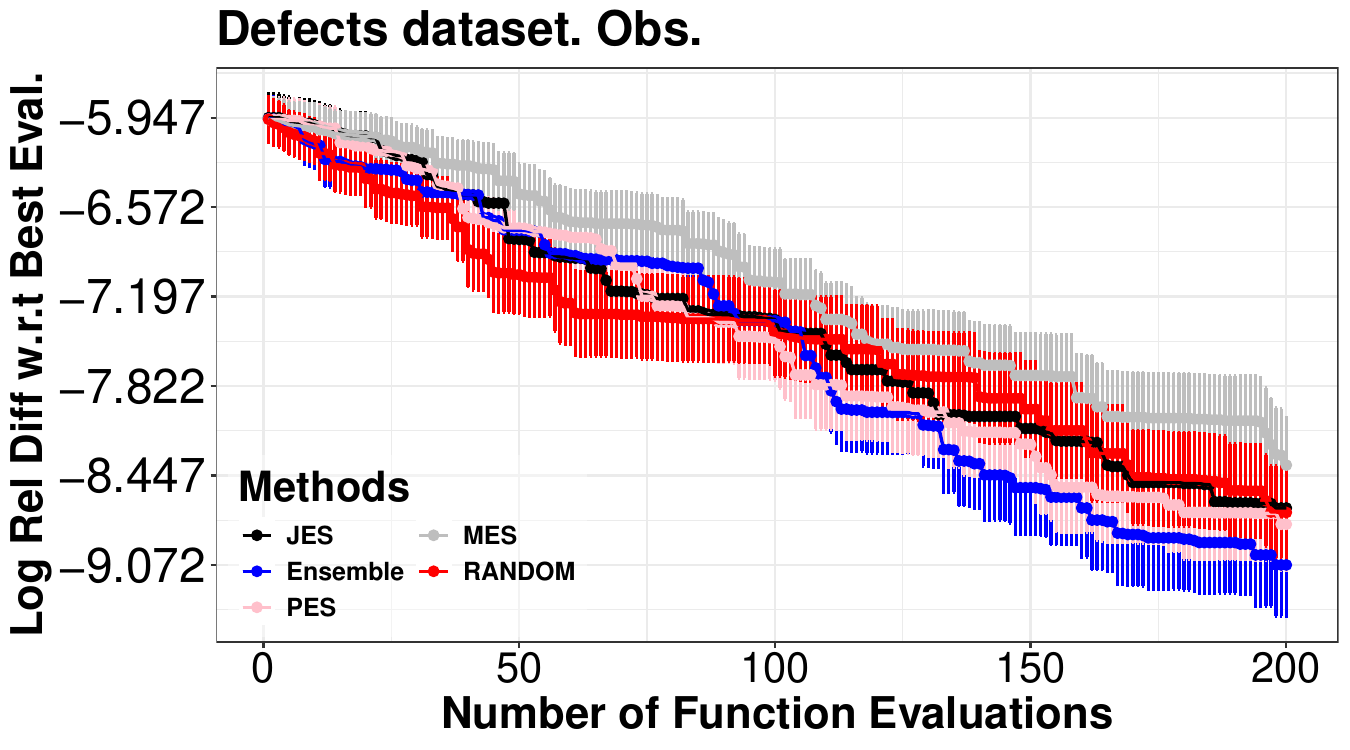}
		\includegraphics[width=0.49\textwidth]{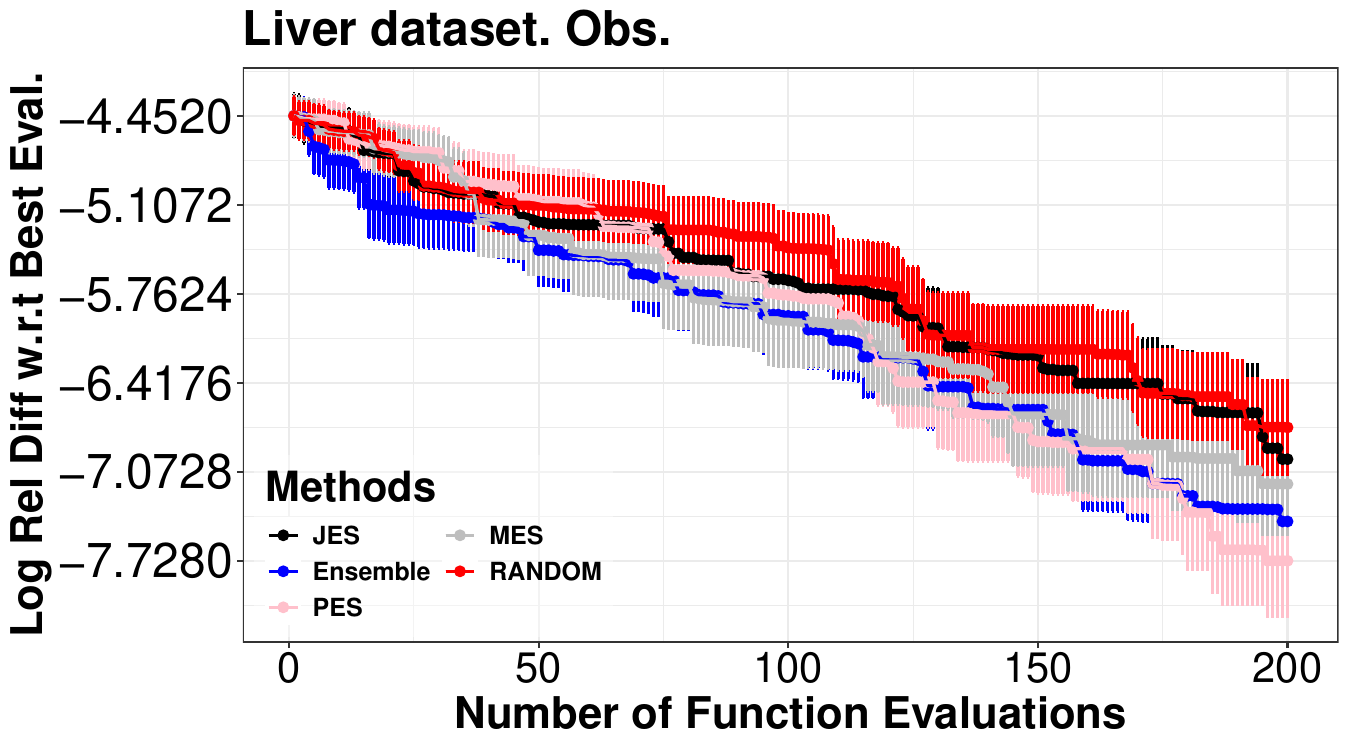} \\
		\includegraphics[width=0.49\textwidth]{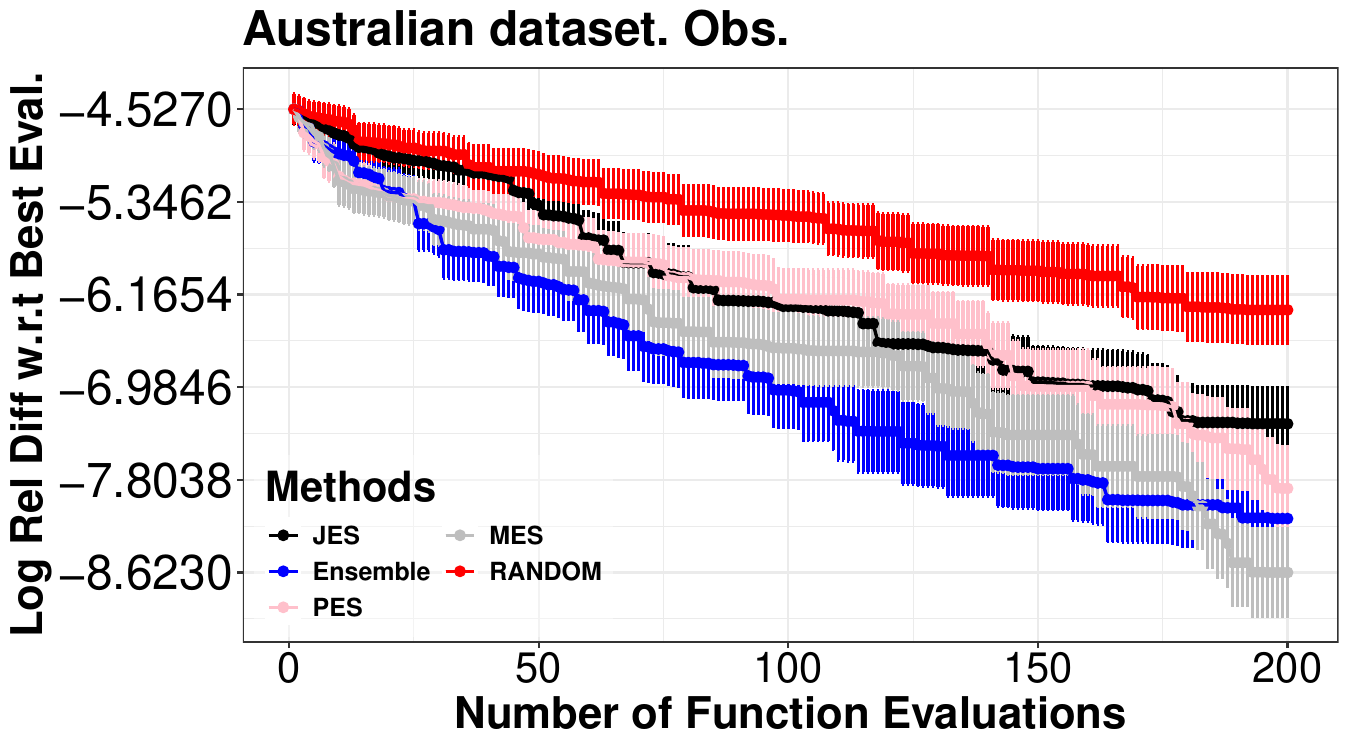}
		\includegraphics[width=0.49\textwidth]{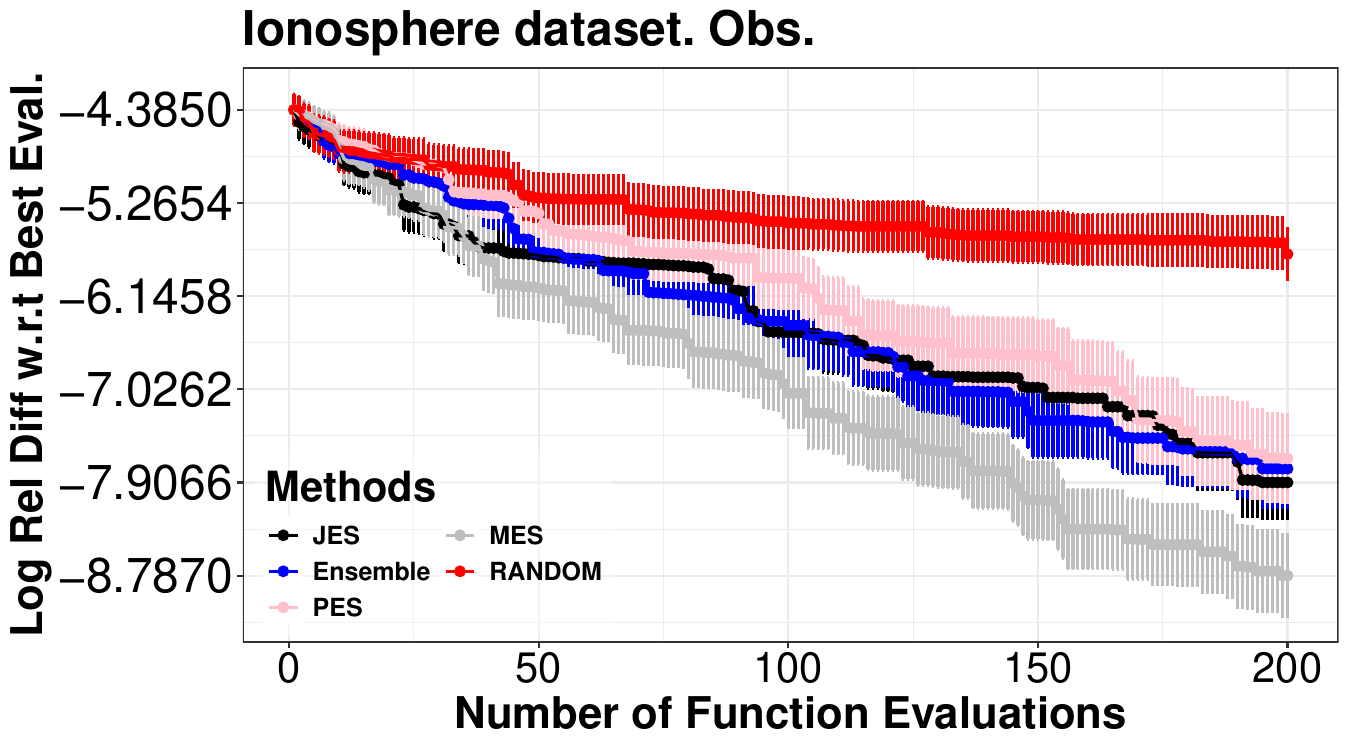}
	\end{tabular}
	\caption{ 
        Average log relative difference between the objective at each method's
        recommendation and the best objective value observed, with respect to
        the number of evaluations performed. We present results for optimizing
        the hyper-parameters of a neural network on the datasets 
	Pima, Liver, Image, Defects, Australian and Ionosphere.
        Best seen in color. }
	\label{FIG:EXPREAL}
\end{figure*}

Figure \ref{FIG:EXPREAL} shows the results obtained. Note that in these experiments the 
variability from one repetition to another is very large, which is translated in high error bars 
and smaller differences among methods. Furthermore, the performance of random search is very close 
to that of the compared methods in some datasets. This indicates that the GP could not be 
a very good model in these problems and that model bias may play an important role.
In spite of this, we observe that the ensemble method 
generally achieves good performance results.
 Specifically, it performs better than JES in Pima, Australian 
and Ionosphere, although the differences are small. In the other datasets it gives 
similar results to JES.  The ensemble method is also, in general, comparable or slightly better 
than the other methods. The only exception is the Ionosphere dataset, where MES performs better. 
However, MES performs poorly on the Defects dataset, where it is outperformed by 
random search.  JES performs well on the Image 
dataset, but encounters difficulties on Pima and Australian. PES consistently lags 
behind, being the worst-performing information-based BO method. Finally, random search 
performs the worst overall, especially in the Pima and Ionosphere 
datasets. Summing up, despite the noise in these experiments, the ensemble method 
attains good results in these problems, achieving results that are similar and sometimes
better than those of the state-of-the-art.

\section{Conclusions} \label{SEC:CON}

This paper has introduced Alpha Entropy Search (AES),
a method for BO whose acquisition function formulation is based 
on information theory. Specifically, AES generalizes previous methods that aim at
choosing the next evaluation point as the one that is expected
to minimize the most the entropy of the solution of the problem.
AES measures the level of dependency between the objective at
the candidate point to evaluate, $y$, and the problem's solution
$\{\mathbf{x}^\star, y^\star\}$ both in the input and the output space.
For this, the $\alpha$-divergence is used instead of the typical KL-divergence
of information-based methods. The $\alpha$-divergence has a parameter $\alpha$ that 
trades-off evaluating differences between each distribution at a single mode, 
and evaluating differences globally. We did not found a particular value of
$\alpha$ that generally provided best over-all results. Therefore, we considered
an ensemble method that simultaneously considers a range of values for $\alpha$.

Our experiments in synthetic, benchmark and real-world problems show 
that the ensemble method performs better than considering a single value of $\alpha$ and that 
it arxvicprovides competitive results with the state-of-the-art methods for information based BO.
Namely, JES, MES and PES. This is particularly the case in a noiseless evaluation setting. In a 
noisy evaluation setting, however, the differences among methods are smaller and our 
proposed method gives similar results to those of the state-of-the-art. 

The computational cost of the ensemble method is larger than that of the other 
information-based strategies (11 times more expensive). This may be seen as a disadvantage. 
However, in BO the bottle-neck is always the evaluation of the 
objective function, which is assumed to be significantly more expensive. 
Therefore, under this assumption, the larger computational cost
of the ensemble method w.r.t. the other strategies is negligible.

Finally, our method is very general and could be extended to other settings. 
Specifically, in our work we have exclusively considered $\{\mathbf{x}^\star, y^\star\}$ as 
the solution of the optimization problem. But one may also consider $\mathbf{x}^\star$ or $y^\star$, 
leading to AES generalizations of PES or MES, respectively.

\subsubsection*{Acknowledgements}
The authors acknowledge financial support from project PID2022-139856NB-I00, funded by MCIN and from the Autonomous Community of Madrid (ELLIS Unit Madrid). They also
acknowledge the use of the facilities of Centro de Computaci\'on Cient\'ifica, UAM.
This publication is also part of the R\&D\&i project Semi Automatic Meta Analysis helps with Reference PP2024\_32, funded by the Universidad Pontificia Comillas.

\bibliographystyle{abbrv}
\bibliography{references}

\appendix

\section{KL-divergence and Joint Entropy Search} \label{SEC:AP1}

Here, we show that 
the acquisition function of Joint Entropy Search (JES) is
given by the Kullback-Leibler divergence between the conditional distribution
$p(\{\mathbf{x}^\star, y^\star\},y|\mathcal{D}_{t-1},\mathbf{x})$
and the product of the marginal distributions $p(\{\mathbf{x}^\star,y^\star\}|\mathcal{D}_{t-1})$
and $p(y|\mathcal{D}_{t-1}, \mathbf{x})$:
\begin{align}
    a_{\text{JES}}(\mathbf{x}) &= \text{KL}(p(\{\mathbf{x}^\star, y^\star\},y|\mathcal{D}_{t-1},\mathbf{x})||p(\{\mathbf{x}^\star,y^\star\}|\mathcal{D}_{t-1})p(y|\mathcal{D}_{t-1}, \mathbf{x})) \nonumber \\
    &= \int p(\{\mathbf{x}^\star, y^\star\},y|\mathcal{D}_{t-1},\mathbf{x}) \log {\frac{p(\{\mathbf{x}^\star, y^\star\},y|\mathcal{D}_{t-1},\mathbf{x})}{p(\{\mathbf{x}^\star,y^\star\}|\mathcal{D}_{t-1})p(y|\mathcal{D}_{t-1}, \mathbf{x})}} d\{\mathbf{x}^\star,y^\star\} dy \nonumber \\
    &= \int p(\{\mathbf{x}^\star, y^\star\},y|\mathcal{D}_{t-1},\mathbf{x})
       \log {\frac{p(y|\{\mathbf{x}^\star, y^\star\},\mathcal{D}_{t-1},\mathbf{x})\cancel{p(\{\mathbf{x}^\star, y^\star\}|\mathcal{D}_{t-1})}}{\cancel{p(\{\mathbf{x}^\star,y^\star\}|\mathcal{D}_{t-1})}p(y|\mathcal{D}_{t-1}, \mathbf{x})}} d\{\mathbf{x}^\star,y^\star\} dy \nonumber \\
    &= \int p(\{\mathbf{x}^\star, y^\star\},y|\mathcal{D}_{t-1},\mathbf{x})
       \log {p(y|\{\mathbf{x}^\star, y^\star\},\mathcal{D}_{t-1},\mathbf{x})} d\{\mathbf{x}^\star,y^\star\} dy \nonumber \\
    & \qquad - \int p(\{\mathbf{x}^\star, y^\star\},y|\mathcal{D}_{t-1},\mathbf{x}) \log {p(y|\mathcal{D}_{t-1}, \mathbf{x})} d\{\mathbf{x}^\star,y^\star\} dy \nonumber \\
    &= \int p(\{\mathbf{x}^\star, y^\star\},y|\mathcal{D}_{t-1},\mathbf{x})
       \log {p(y|\{\mathbf{x}^\star, y^\star\},\mathcal{D}_{t-1},\mathbf{x})} d\{\mathbf{x}^\star,y^\star\} dy
    + H \left [ p(y|\mathcal{D}_{t-1}, \mathbf{x}) \right] \nonumber \\
    &= - \int p(\{\mathbf{x}^\star, y^\star\}|\mathcal{D}_{t-1})
       H \left [ p(y|\{\mathbf{x}^\star, y^\star\},\mathcal{D}_{t-1},\mathbf{x}) \right ]
       d\{\mathbf{x}^\star,y^\star\} dy 
    + H \left [ p(y|\mathcal{D}_{t-1}, \mathbf{x}) \right] \nonumber \\
    &= H \left [ p(y|\mathcal{D}_{t-1}, \mathbf{x}) \right]
	- \mathds{E}_{p(\{\mathbf{x}^\star, y^\star\}|\mathcal{D}_{t-1})} \left [
        H \left [ p(y|\{\mathbf{x}^\star, y^\star\},\mathcal{D}_{t-1},\mathbf{x}) \right ]
       \right ]
    \,, 
    \label{EQ:KL_PYX_PYPX_appendix}
\end{align}
where we have used the product rule of probability and 
that $p(\{\mathbf{x}^\star, y^\star\}|\mathcal{D}_{t-1})$ does not depend on $\mathbf{x}$.
As in the main document, $H \left [ p(y|\mathcal{D}_{t-1}, \mathbf{x}) \right]$
is the entropy of the predictive distribution of the GP at $\mathbf{x}$, given the data already observed $\mathcal{D}_{t-1}$,
and $H \left [ p(y|\{\mathbf{x}^\star, y^\star\},\mathcal{D}_{t-1}, \mathbf{x}) \right]$
is the entropy of the conditional predictive distribution, at $\mathbf{x}$, given the data already observed $\mathcal{D}_{t-1}$
and that the solution of the optimization problem is $\{\mathbf{x}^\star, y^\star\}$.

\section{Derivation of Alpha Entropy Search} \label{SEC:AP2}

In the main document, we propose replacing the KL-divergence in JES
with a more general divergence, Amari's $\alpha$-divergence.
This divergence includes the parameter $\alpha$, which allows us to vary the
weight given to discrepancies between distributions across different regions.
Specifically, by adjusting $\alpha$, we can amplify or down-weight differences
across various areas of the input space. This substitution results in the
following acquisition function:
\begin{align}
	a_{\text{AES}}(\mathbf{x})
	&= D_{\alpha}(p(y,\{y^\star, \mathbf{x}^\star\}|\mathcal{D}_{t-1}, \mathbf{x})||p(\{y^\star, \mathbf{x}^\star\}|\mathcal{D}_{t-1}, \mathbf{x})p(y|\mathcal{D}_{t-1}, \mathbf{x})) \nonumber \\
    &= \textstyle 
    \frac{1}{(1 - \alpha) \alpha}
    \left( 
	1 - \int
                 \left (
                     p(\{y^\star, \mathbf{x}^\star\}|\mathcal{D}_{t-1}) 
                     p(y|\mathcal{D}_{t-1}, \mathbf{x})
                 \right )^{1-\alpha}
                 p(\{\mathbf{x}^\star, y^\star\},y|\mathcal{D}_{t-1},\mathbf{x})^{\alpha}
            d\{\mathbf{x}^\star,y^\star\} dy
    \right) \nonumber \\
    &= \textstyle 
    \frac{1}{(1 - \alpha) \alpha}
    \left( 
	1 - \int
                 p(\{y^\star, \mathbf{x}^\star\}|\mathcal{D}_{t-1})
                 p(y|\mathcal{D}_{t-1}, \mathbf{x})
                 \left (
                     \frac{
                         p(\{\mathbf{x}^\star, y^\star\},y|\mathcal{D}_{t-1},\mathbf{x})
                     }{
                         p(\{y^\star, \mathbf{x}^\star\}|\mathcal{D}_{t-1})
                         p(y|\mathcal{D}_{t-1}, \mathbf{x})
                     }
                \right)^\alpha
                 d\{\mathbf{x}^\star,y^\star\} dy
    \right) \nonumber \\
    &= \textstyle
    \frac{1}{(1 - \alpha) \alpha}
    \left( 
	1 - \mathds{E}_{p(\{y^\star, \mathbf{x}^\star\}|\mathcal{D}_{t-1})}
            \left[
                \int 
		p(y|\mathcal{D}_{t-1}, \mathbf{x})
                \left(
		    \frac{
                    p(\{\mathbf{x}^\star, y^\star\},y|\mathcal{D}_{t-1},\mathbf{x})
                }{
                    p(\{y^\star, \mathbf{x}^\star\}|\mathcal{D}_{t-1})
                    p(y|\mathcal{D}_{t-1}, \mathbf{x})
                }
                \right)^\alpha dy 
            \right]
    \right) \nonumber \\
     &= \textstyle
    \frac{1}{(1 - \alpha) \alpha}
    \left( 
	1 - \mathds{E}_{p(\{y^\star, \mathbf{x}^\star\}|\mathcal{D}_{t-1})}
            \left[
                \int 
		p(y|\mathcal{D}_{t-1}, \mathbf{x})
                \left(
		    \frac{
                    p(y|\{\mathbf{x}^\star, y^\star\},\mathcal{D}_{t-1},\mathbf{x})\cancel{p(\{\mathbf{x}^\star, y^\star\}|\mathcal{D}_{t-1})}
                }{
                    \cancel{p(\{\mathbf{x}^\star,y^\star\}|\mathcal{D}_{t-1})}p(y|\mathcal{D}_{t-1}, \mathbf{x})
                }
                \right)^\alpha dy 
            \right]
    \right) \nonumber \\
    &= \textstyle
    \frac{1}{(1 - \alpha) \alpha}
    \left( 
	1 - \mathds{E}_{p(\{y^\star, \mathbf{x}^\star\}|\mathcal{D}_{t-1})}
            \left[
                \int 
		p(y|\mathcal{D}_{t-1}, \mathbf{x})
                \left(
		    \frac{p(y|\mathcal{D}_{t-1},\mathbf{x},\{y^\star, \mathbf{x}^\star\})}{p(y|\mathcal{D}_{t-1}, \mathbf{x})}
                \right)^\alpha dy 
            \right]
    \right)\,.
\label{EQ:aES_acq_appendix}
\end{align} 
As with other information-based BO methods, this expression is analytically
intractable and requires approximation. In particular, neither the expectation
in \eqref{EQ:aES_acq_appendix} nor the conditional distribution
$p(y|\mathcal{D}_{t-1},\mathbf{x},\{y^\star, \mathbf{x}^\star\})$ can be computed in closed form.  
The approximation of the conditional distribution is discussed in Section \ref{SUB:COMPDIV} of the main
document, and in Appendix \ref{SEC:AP3}, we detail how to evaluate the integral in \eqref{EQ:aES_acq_appendix}.
Again, we have used the product rule of probability and 
that $p(\{\mathbf{x}^\star, y^\star\}|\mathcal{D}_{t-1})$ does not depend on $\mathbf{x}$.

\section{Evaluating the Integral in AES w.r.t $y$} \label{SEC:AP3}

In this appendix we show how to evaluate the integral:
\begin{align}
	\int 
		p(y|\mathcal{D}_{t-1}, \mathbf{x})
                \left(
		    \frac{p(y|\mathcal{D}_{t-1},\mathbf{x},\{y^\star, \mathbf{x}^\star\})}{p(y|\mathcal{D}_{t-1}, \mathbf{x})}
                \right)^\alpha dy \,,
		\label{eq:integral}
\end{align} 
where $p(y|\mathcal{D}_{t-1},\mathbf{x},\{y^\star, \mathbf{x}^\star\})$ is approximated using a truncated Gaussian distribution,
as described in the main manuscript. This integral involves a product and a ratio between the predictive distribution
conditioned to the problem's solution and the unconditioned distribution to the power of $\alpha$.
Given that both distributions are Gaussian (after the approximations described in the main manuscript) and that 
the Gaussian belongs to the exponential family of distributions, we can evaluate
the integral in closed form using the exponential form of the Gaussian distribution.

First, recall that the density of a Gaussian distribution with mean $\mu$ and variance $\sigma^2$ can be expressed using a natural 
parameters representation in terms of natural parameters $\boldsymbol{\eta}$, the vector of sufficient statistics $\mathbf{u}(x)$, and a log-partition 
function $g(\boldsymbol{\eta})$.
Namely,
\begin{align}
	f(x \mid \mu, \sigma^2) = \frac{1}{\sqrt{2 \pi \sigma^2}} \exp \left( -\frac{(x - \mu)^2}{2 \sigma^2} \right) = 
	\exp\left( \boldsymbol{\eta}^\top \mathbf{u}(x) - g(\boldsymbol{\eta}) \right)\,,
\end{align}
where $\boldsymbol{\eta} = \left(\eta_1 = \frac{\mu}{\sigma^2},\eta_2 = \frac{1}{\sigma^2}\right)^\top$, sufficient statistics 
are $\mathbf{u}(x) = \left( x, -0.5x^2 \right)^\top$ and the log-partition function is $g(\boldsymbol{\eta}) = \frac{\eta_1^2}{2\eta_2} + \frac{1}{2}\log(2\pi)-\frac{1}{2}\log (\eta_2)$. 
This notation greatly simplifies the evaluation of the aforementioned integral.

Using the previous representation of the Gaussian distribution we can now focus on Eq. (\ref{eq:integral}),
which involves two different Gaussians, where each of them can be converted into its natural parameter 
representation in the following way:
\begin{align}
	p(y|\mathcal{D}_{t-1}, \mathbf{x}) & = \exp\left( \bm{\eta}^\top u(\mathbf{x}) - g(\bm{\eta}) \right)\,, &
	p(y|\mathcal{D}_{t-1}, \mathbf{x}, \{ y^\star, \mathbf{x}^\star \}) & = \exp\left( \bm{\eta}^\star{}^\top u(\mathbf{x}) - g(\bm{\eta}^\star) \right)\,,
\end{align}
where $\bm{\eta}$ are the natural parameter of the unconditioned predictive distribution of the GP at $\mathbf{x}$, 
and where $\bm{\eta}^\star$ are the natural parameters of the approximate conditional predictive distribution at $\mathbf{x}$, 
given that $\{ y^\star, \mathbf{x}^\star \}$ is the solution of the optimization problem. Now, we can compute the ratio inside the integral as:
\begin{align}
    \left( \frac{p(y|\mathcal{D}_{t-1}, \mathbf{x}, \{ y^\star, \mathbf{x}^\star \})}{p(y|\mathcal{D}_{t-1}, \mathbf{x})} \right)^\alpha
	&= \left( \frac{ \exp\left( \bm{\eta}^\star{}^\top \mathbf{u}(y) - g(\bm{\eta}^\star) \right) }{ \exp\left( \bm{\eta}^\top u(\mathbf{x}) - g(\bm{\eta}) \right) } \right)^\alpha \\
	&= \exp\left( \alpha \left[ (\bm{\eta}^\star - \bm{\eta})^\top \mathbf{u}(y) - (g(\bm{\eta}^\star) - g(\bm{\eta})) \right] \right),
\end{align}
and substitute inside the integral also substituting the first factor 
$p(y|\mathcal{D}_{t-1}, \mathbf{x})$ by $\exp\left( \bm{\eta}^\top \mathbf{u}(y) - g(\bm{\eta}) \right)$, obtaining:
\begin{align}
	& \int \exp\left( \bm{\eta}^\top \mathbf{u}(y) - g(\bm{\eta}) \right) \exp\left( 
	\alpha \left[ (\bm{\eta}^\star - \bm{\eta})^\top \mathbf{u}(y) - \left( g(\bm{\eta}^\star) - g(\bm{\eta}) \right) \right] \right) dy, \nonumber \\
	=\, & \int \exp\Big( \bm{\eta}^\top \mathbf{u}(y) - g(\bm{\eta}) + \alpha (\bm{\eta}^\star - \bm{\eta})^\top \mathbf{u}(y) - \alpha \left( g(\bm{\eta}^\star) - g(\bm{\eta}) \right) \Big) dy \nonumber \\
	=\, & \int \exp\Big( \bm{\eta}^\top \mathbf{u}(y) - g(\bm{\eta}) + \alpha \bm{\eta}^\star{}^\top \mathbf{u}(y) - \alpha \bm{\eta}^\top \mathbf{u}(y) - \alpha g(\bm{\eta}^\star) + \alpha g(\bm{\eta}) \Big) dy \nonumber \\
	=\, & \int \exp\Big( \left( \bm{\eta}^\top \mathbf{u}(y) - \alpha \bm{\eta}^\top \mathbf{u}(y) \right) + \alpha \bm{\eta}^\star{}^\top \mathbf{u}(y) + \left( - g(\bm{\eta}) + \alpha g(\bm{\eta}) \right) - \alpha g(\bm{\eta}^\star) \Big) dy \nonumber \\
	=\, & \int \exp\Big( \left( (1 - \alpha) \bm{\eta}^\top \mathbf{u}(y) + \alpha \bm{\eta}^\star{}^\top \mathbf{u}(y) \right) + (\alpha - 1) g(\bm{\eta}) - \alpha g(\bm{\eta}^\star) \Big) dy \nonumber \\
	=\, & \exp\left( (\alpha - 1) g(\bm{\eta}) - \alpha g(\bm{\eta}^\star) \right) \int \exp\left( \left( (1 - \alpha) \bm{\eta} + \alpha \bm{\eta}^\star \right)^\top \mathbf{u}(y) \right) dy \nonumber \\
=\, & \exp\left( (\alpha - 1) g(\bm{\eta}) - \alpha g(\bm{\eta}^\star) \right) \exp\left( g\left( (1 - \alpha) \bm{\eta} + \alpha \bm{\eta}^\star \right) \right) \nonumber \\
=\, & \exp\left( (\alpha - 1) g(\bm{\eta}) -\alpha g(\bm{\eta}^\star) + g\left( (1 - \alpha)\bm{\eta} + \alpha \bm{\eta}^\star \right) \right),
\end{align}
where the last integral is simply given by the exponential of the log-normalizer of a Gaussian, $g(\cdot)$ with natural parameters $(1 - \alpha) \bm{\eta} + \alpha \bm{\eta}^\star$. 
Summing up, we have obtained that:
\begin{align}
	\int
		p(y|\mathcal{D}_{t-1}, \mathbf{x})
                \left(
		    \frac{p(y|\mathcal{D}_{t-1},\mathbf{x},\{y^\star, \mathbf{x}^\star\})}{p(y|\mathcal{D}_{t-1}, \mathbf{x})}
                \right)^\alpha dy
	& =
	\exp\left\{ (\alpha - 1) g(\bm{\eta}) -\alpha g(\bm{\eta}^\star)\right. \nonumber \\
	& \quad \left. + g((1 - \alpha)\bm{\eta} + \alpha \bm{\eta}^\star) \right\}\,,
\end{align}
where $g(\bm{\eta})$ is the log-normalizer of a Gaussian with natural parameters $\bm{\eta}$,
$\bm{\eta}$ are the natural parameters of $p(y|\mathcal{D}_{t-1}, \mathbf{x})$, and
$\bm{\eta}^\star$ are the natural parameters of the Gaussian approximation of $p(y|\mathcal{D}_{t-1},\mathbf{x},\{y^\star, \mathbf{x}^\star\})$.
Specifically,
\begin{align}
	g({\bm \eta}) & = 0.5 \log ( 2 \pi) - 0.5 \log \eta_2 + 0.5 \frac{\eta_1^2}{\eta_2}\,, &
	\bm{\eta} & = \left(\frac{m(\mathbf{x})}{v(\mathbf{x})+\sigma^2}, \frac{1}{v(\mathbf{x}) + \sigma^2} \right)^\top\,, \nonumber \\
	\bm{\eta}^\star & = \left(\frac{m_\text{tr}(\mathbf{x})}{v_\text{tr}(\mathbf{x})+\sigma^2}, \frac{1}{v_\text{tr}(\mathbf{x}) + \sigma^2} \right)^\top\,,
\end{align}
where $m(\mathbf{x})$, $v(\mathbf{x})$, $m_\text{tr}(\mathbf{x})$ and $v_\text{tr}(\mathbf{x})$ are respectively the mean and variances
of the unconditional and conditional predictive distribution for $f(\mathbf{x})$, and $\sigma^2$ is the variance of the noise.

\section{AES and JES Approximations when $\alpha \rightarrow 1$} \label{SEC:AP4}

As explained in the main document, even when $S \rightarrow \infty$ and $\alpha \rightarrow 1$ $\tilde{a}_{\text{AES}}(\mathbf{x}) \nrightarrow  \tilde{a}_\text{JES}(\mathbf{x})$, although this might not be obvious. In this appendix, we provide the details for this result.

As shown in Appendix \ref{SEC:AP1}, we can express the JES acquisition 
as the KL-divergence between $p(\{\mathbf{x}^\star, y^\star\},y|\mathcal{D}_{t-1},\mathbf{x})$ and the product of 
the marginals $p(\{\mathbf{x}^\star,y^\star\}|\mathcal{D}_{t-1})$ and $p(y|\mathcal{D}_{t-1}, \mathbf{x}))$:
\begin{align}
    a_{\text{JES}}(\mathbf{x}) &= \text{KL}(p(\{\mathbf{x}^\star, y^\star\},y|\mathcal{D}_{t-1},\mathbf{x})||p(\{\mathbf{x}^\star,y^\star\}|\mathcal{D}_{t-1})p(y|\mathcal{D}_{t-1}, \mathbf{x})) \nonumber \\
    &= \int p(\{\mathbf{x}^\star, y^\star\},y|\mathcal{D}_{t-1},\mathbf{x}) \log {\frac{p(\{\mathbf{x}^\star, y^\star\},y|\mathcal{D}_{t-1},\mathbf{x})}{p(\{\mathbf{x}^\star,y^\star\}|\mathcal{D}_{t-1})p(y|\mathcal{D}_{t-1}, \mathbf{x})}} d\{\mathbf{x}^\star,y^\star\} dy \nonumber \\
    &= \int p(\{\mathbf{x}^\star, y^\star\},y|\mathcal{D}_{t-1},\mathbf{x})
       \log {\frac{p(y|\{\mathbf{x}^\star, y^\star\},\mathcal{D}_{t-1},\mathbf{x})\cancel{p(\{\mathbf{x}^\star, y^\star\}|\mathcal{D}_{t-1})}}{\cancel{p(\{\mathbf{x}^\star,y^\star\}|\mathcal{D}_{t-1})}p(y|\mathcal{D}_{t-1}, \mathbf{x})}} d\{\mathbf{x}^\star,y^\star\} dy \nonumber \\
    &= \int p(\{\mathbf{x}^\star, y^\star\},y|\mathcal{D}_{t-1},\mathbf{x})
       \log {p(y|\{\mathbf{x}^\star, y^\star\},\mathcal{D}_{t-1},\mathbf{x})} d\{\mathbf{x}^\star,y^\star\} dy \nonumber \\
    & \qquad - \int p(\{\mathbf{x}^\star, y^\star\},y|\mathcal{D}_{t-1},\mathbf{x}) \log {p(y|\mathcal{D}_{t-1}, \mathbf{x})} d\{\mathbf{x}^\star,y^\star\} dy 
    \nonumber \\
    &= H \left [ p(y|\mathcal{D}_{t-1}, \mathbf{x}) \right]
    - \mathds{E}_{p(\{y^\star, \mathbf{x}^\star\}|\mathcal{D}_{t-1})} \left [
        H \left [ p(y|\{\mathbf{x}^\star, y^\star\},\mathcal{D}_{t-1},\mathbf{x}) \right ]
       \right ]
    \,, 
    \label{EQ:aJES_part2}
\end{align}
where the expectation has to be approximated by Monte Carlo using $S$ samples of $\{\mathbf{x}^\star,y^\star\}$ 
and the conditional distribution $p(y|\{y^\star, \mathbf{x}^\star\},\mathcal{D}_{t-1}, \mathbf{x})$
is approximated using a truncated Gaussian, as described in the main document.

On the other hand, in Appendix \ref{SEC:AP2} we describe the AES acquisition 
as Amari's $\alpha$-divergence between $p(\{\mathbf{x}^\star, y^\star\},y|\mathcal{D}_{t-1},\mathbf{x})$
and the product of the marginals
$p(\{\mathbf{x}^\star,y^\star\}|\mathcal{D}_{t-1})$ and $p(y|\mathcal{D}_{t-1}, \mathbf{x}))$. That is,
\begin{align}
	a_{\text{AES}}(\mathbf{x})
	&= D_{\alpha}(p(y,\{y^\star, \mathbf{x}^\star\}|\mathcal{D}_{t-1}, \mathbf{x})||p(\{y^\star, \mathbf{x}^\star\}|\mathcal{D}_{t-1}, \mathbf{x})p(y|\mathcal{D}_{t-1}, \mathbf{x})) \nonumber \\
    &= \textstyle
    \frac{1}{(1 - \alpha) \alpha}
    \left( 
	1 - \mathds{E}_{p(\{y^\star, \mathbf{x}^\star\}|\mathcal{D}_{t-1})}
            \left[
                \int 
		p(y|\mathcal{D}_{t-1}, \mathbf{x})
                \left(
		    \frac{
                    p(\{\mathbf{x}^\star, y^\star\},y|\mathcal{D}_{t-1},\mathbf{x})
                }{
                    p(\{y^\star, \mathbf{x}^\star\}|\mathcal{D}_{t-1})
                    p(y|\mathcal{D}_{t-1}, \mathbf{x})
                }
                \right)^\alpha dy 
            \right]
    \right)
    \,. 
\label{EQ:aES_part1}
\end{align} 
Again, we cannot compute this expression analytically. The expectation is approximated via Monte Carlo using $S$ samples of $\{\mathbf{x}^\star,y^\star\}$,
and the conditional distribution $p(y|\{y^\star, \mathbf{x}^\star\},\mathcal{D}_{t-1}, \mathbf{x})$
is approximated using a truncated Gaussian, as described in the main document.

If the exact conditional distribution $p(y|\{y^\star, \mathbf{x}^\star\},\mathcal{D}_{t-1}, \mathbf{x})$ is used, 
then AES and JES give the same result when $S \rightarrow \infty$ and $\alpha \rightarrow 1$.  
However, consider now the approximation of the conditional 
distribution $p(y|\{y^\star, \mathbf{x}^\star\},\mathcal{D}_{t-1}, \mathbf{x})$
using a truncated Gaussian. Let that approximate distribution be $\tilde{p}(y|\{y^\star, \mathbf{x}^\star\},\mathcal{D}_{t-1}, \mathbf{x})$.
This step is common in both AES and JES. In the case of AES, the corresponding approximate  acquisition is:
\begin{align}
	\tilde{a}_{\text{AES}}(\mathbf{x}) &= \textstyle
    \frac{1}{(1 - \alpha) \alpha} \left( 1 - \mathds{E}_{p(\{y^\star, \mathbf{x}^\star\}|\mathcal{D}_{t-1})}
            \left[ \int p(y|\mathcal{D}_{t-1}, \mathbf{x})
		\left( \frac{ \tilde{p}(y |\{\mathbf{x}^\star, y^\star\},\mathcal{D}_{t-1},\mathbf{x}) }
		{ p(y|\mathcal{D}_{t-1}, \mathbf{x}) } \right)^\alpha dy 
            \right] \right)
	    \nonumber \\
	    & = 
\textstyle
    \frac{1}{(1 - \alpha) \alpha} \left( 1 - \mathds{E}_{p(\{y^\star, \mathbf{x}^\star\}|\mathcal{D}_{t-1})}
            \left[ \int p(y|\mathcal{D}_{t-1}, \mathbf{x})
		\left( \frac{ \tilde{p}(\{\mathbf{x}^\star, y^\star\},y|\mathcal{D}_{t-1},\mathbf{x}) }
		{ p(\{y^\star, \mathbf{x}^\star\}|\mathcal{D}_{t-1}) p(y|\mathcal{D}_{t-1}, \mathbf{x})
                } \right)^\alpha dy 
            \right] \right) \nonumber \\
	&= D_{\alpha}(\tilde{p}(y,\{y^\star, \mathbf{x}^\star\}|\mathcal{D}_{t-1}, \mathbf{x})||p(\{y^\star, \mathbf{x}^\star\}|\mathcal{D}_{t-1}, \mathbf{x})p(y|\mathcal{D}_{t-1}, \mathbf{x})) 
    \,, 
\end{align}
where 
\begin{align}
	\tilde{p}(\{\mathbf{x}^\star, y^\star\},y|\mathcal{D}_{t-1},\mathbf{x})  &=  \tilde{p}(y|\{y^\star, \mathbf{x}^\star\},\mathcal{D}_{t-1}, \mathbf{x})
	p(\{y^\star, \mathbf{x}^\star\}|\mathcal{D}_{t-1})\,,
	\label{eq:approx_joint}
\end{align}
is an approximate joint distribution. Thus, when $\alpha \rightarrow 1$ we have that:
\begin{align}
	\tilde{a}_{\text{AES}}(\mathbf{x}) \rightarrow 
	\text{KL}(\tilde{p}(\{\mathbf{x}^\star, y^\star\},y|\mathcal{D}_{t-1},\mathbf{x})||p(\{\mathbf{x}^\star,y^\star\}|\mathcal{D}_{t-1})p(y|\mathcal{D}_{t-1}, \mathbf{x}))\,.
\end{align}
By contrast, in the case of JES, the truncated Gaussian approximation gives the approximate acquisition:
\begin{align}
	\tilde{a}_{\text{JES}}(\mathbf{x}) &= H \left [ p(y|\mathcal{D}_{t-1}, \mathbf{x}) \right]
    - \mathds{E}_{p(\{y^\star, \mathbf{x}^\star\}|\mathcal{D}_{t-1})} \left [
	    H \left [ \tilde{p}(y|\{\mathbf{x}^\star, y^\star\},\mathcal{D}_{t-1},\mathbf{x}) \right ]
       \right ]
       \nonumber \\
    &= - \int p(\{\mathbf{x}^\star, y^\star\},y|\mathcal{D}_{t-1},\mathbf{x}) \log {p(y|\mathcal{D}_{t-1}, \mathbf{x})} d\{\mathbf{x}^\star,y^\star\} dy \nonumber \\
	& \quad  
	+ \int p(\{\mathbf{x}^\star, y^\star\},y|\mathcal{D}_{t-1},\mathbf{x})
	\log {\tilde{p}(y|\{\mathbf{x}^\star, y^\star\},\mathcal{D}_{t-1},\mathbf{x})} d\{\mathbf{x}^\star,y^\star\} dy \nonumber \\
	&= \int p(\{\mathbf{x}^\star, y^\star\},y|\mathcal{D}_{t-1},\mathbf{x}) 
	\log {\frac{\tilde{p}(\{\mathbf{x}^\star, y^\star\},y|\mathcal{D}_{t-1},\mathbf{x})}{p(\{\mathbf{x}^\star,y^\star\}|\mathcal{D}_{t-1})
	p(y|\mathcal{D}_{t-1}, \mathbf{x})}} d\{\mathbf{x}^\star,y^\star\} dy \nonumber \\
	& \neq \text{KL}(\tilde{p}(\{\mathbf{x}^\star, y^\star\},y|\mathcal{D}_{t-1},\mathbf{x})||p(\{\mathbf{x}^\star,y^\star\}|\mathcal{D}_{t-1})p(y|\mathcal{D}_{t-1}, \mathbf{x}))\,,
	\label{eq:jes_not_kl}
\end{align}
since the approximate joint distribution 
$\tilde{p}(\{\mathbf{x}^\star, y^\star\},y|\mathcal{D}_{t-1},\mathbf{x})$ appears only inside 
the log function. Outside the log function appears the exact joint 
distribution $p(\{\mathbf{x}^\star, y^\star\},y|\mathcal{D}_{t-1},\mathbf{x})$.
Note that in (\ref{eq:jes_not_kl}) we have also used (\ref{eq:approx_joint}). 
Therefore, JES uses the exact joint distribution $p(\{\mathbf{x}^\star, y^\star\},y|\mathcal{D}_{t-1},\mathbf{x})$
in one factor, but the approximate joint distribution $\tilde{p}(\{\mathbf{x}^\star, y^\star\},y|\mathcal{D}_{t-1},\mathbf{x})$ in the other. 
This explains why the final approximated expressions for JES 
and AES are different when $\alpha \rightarrow 1$. Summing up, JES and AES need 
not give the same results when $\alpha \rightarrow 1$.

\section{Comparison of AES, JES and the Ensemble Method in a Noisy Evaluation Setting} \label{SEC:AP5}

\begin{figure}[htb!]
  \centering
  \includegraphics[width=1.0\textwidth]{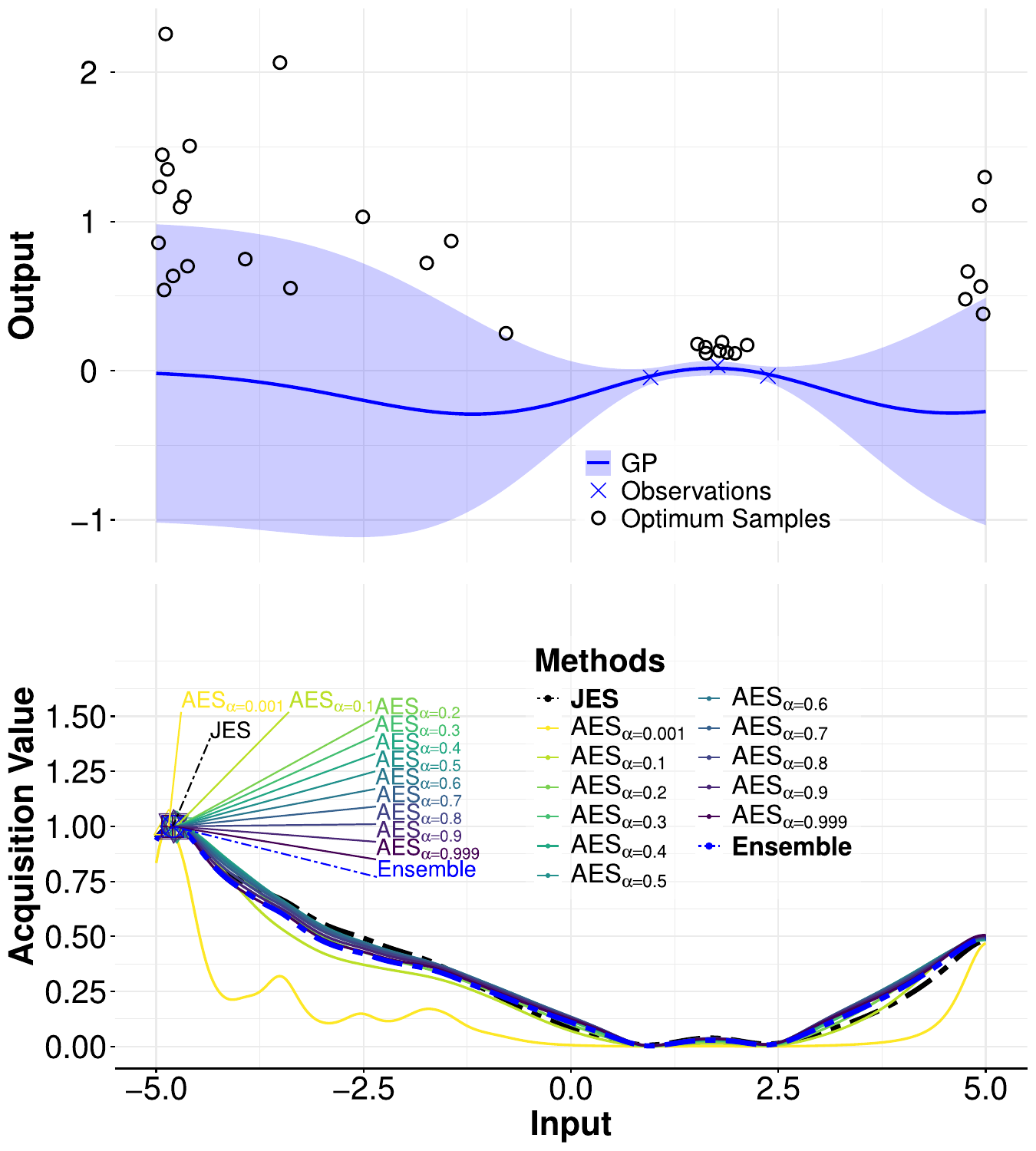}
	\caption{(bottom) Comparison of AES for different $\alpha$ values, JES and the ensemble acquisition function in a 1D noisy synthetic problem. We also display the maximum of each acquisition function.
	(top) Predictive distribution of the GP and generated samples of $\{\mathbf{x}^\star, y^\star\}$. The acquisition functions have been normalized so that the maximum is equal to one
	for a better visualization. Best viewed in color.} 
  \label{fig:comparison_noisy_alpha_values}
\end{figure}

In the main document, we compared JES, AES for different $\alpha$ values and the ensemble method on a 1D noiseless problem. 
We observed that averaging over $\alpha$, the ensemble acquisition function is less rugged than when using $\alpha$ values close to $1$.
In a noisy evaluation scenario, however, the number of peaks, \emph{i.e.}, local maxima of JES and AES, that appear at the 
sampled locations of $\{\mathbf{x}^\star, y^\star\}$ is lower than in a noiseless setting. 
This reduction occurs because the probabilistic model accounts for the presence of noise, so the variance of the 
conditional model does not decrease to zero at the sampled locations. 
This reduction in the number of peaks is shown in Table \ref{tab:n_local_max_noisy} where we display the average number of local 
maxima of JES and the ensemble method across $100$ repetitions on a 1D synthetic experiment under evaluation noise. In this setting, 
there is no statistically significant difference between the methods w.r.t. the number of local maxima in the acquisition. Furthermore, 
Figure \ref{fig:comparison_noisy_alpha_values} shows a plot of the different acquisition functions, for a particular repetition of the 
experiment, where this reduction of the local maxima can be visually observed. In that figure, all acquisition functions indicate that 
the input around $-4.8$ is expected to have the highest utility. Although it may appear that all the maxima are at the same input, 
they are actually very close but different. Specifically, they are in the range $-4.85$ to $-4.74$. Without the adverse 
effect of local maxima, the approach of averaging over different $\alpha$ values is expected to be less advantageous in 
a noisy setting. 

\begin{table}[htb!]
	\centering
	\caption{Average number of local maxima for each method over 100 repetitions in a 1D noisy synthetic problem.}
	\label{tab:n_local_max_noisy}
	\begin{tabular}{lr@{$\pm$}l}
		\hline
		{\bf Method} & \multicolumn{2}{c}{\bf \# of Local Maxima}\\
		\hline
		{\bf JES }     & 4.22 & 0.073\\
		{\bf Ensemble} & 4.21 & 0.071\\ 
		\hline
	\end{tabular}
\end{table}

\section{Impact of Number of Restarts and Candidate Points} \label{SEC:AP6}

\begin{figure*}[tbh!]
	\begin{tabular}{cc}
		\includegraphics[width=0.49\textwidth]{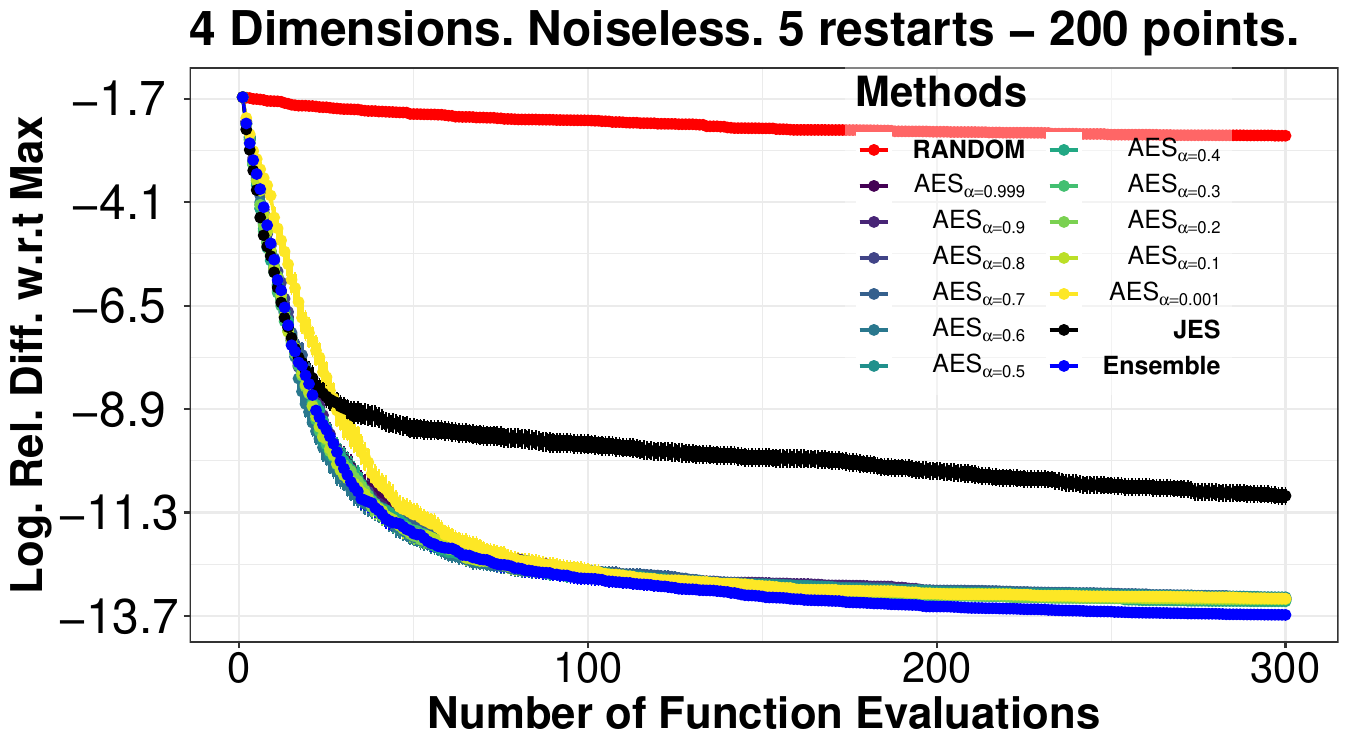}
		\includegraphics[width=0.49\textwidth]{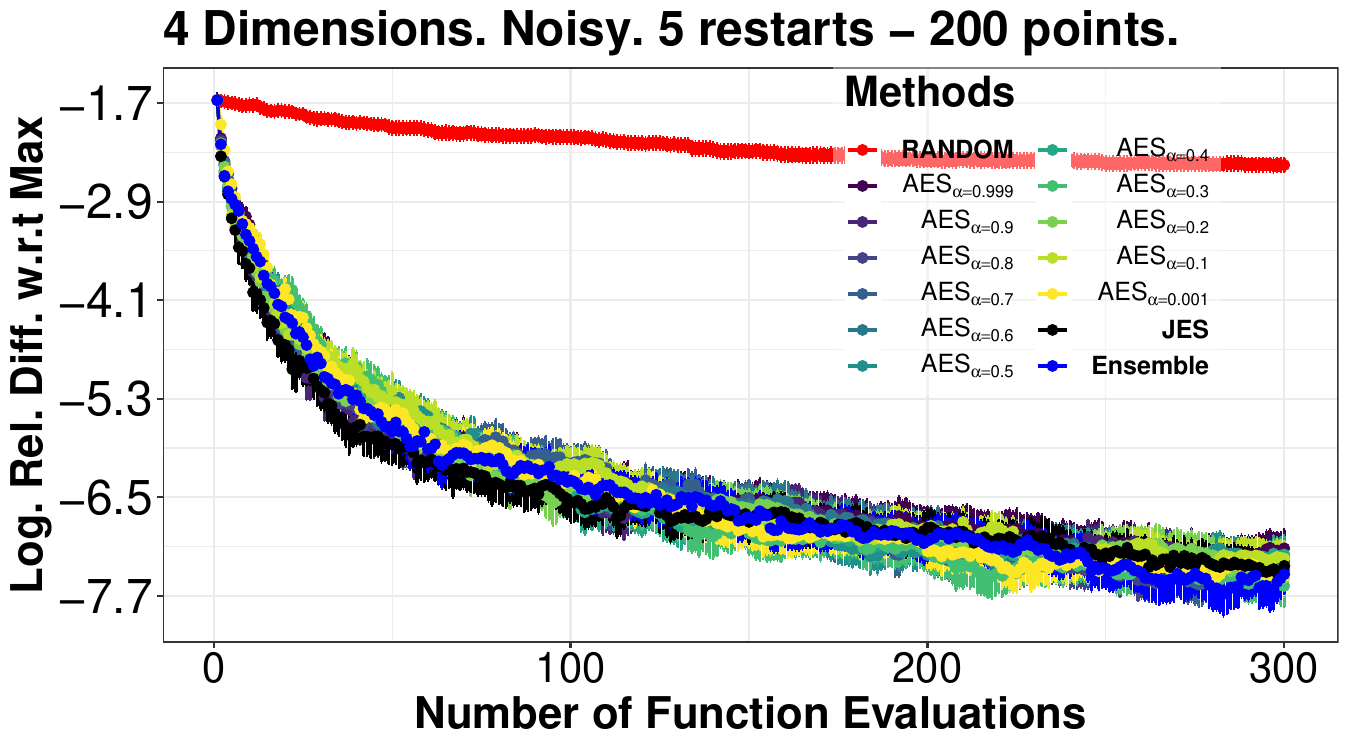}
	\end{tabular}
	\caption{ 
        Average logarithm relative difference between the objective at each method's
        recommendation and the objective at the global maximum, with respect to
        the number of evaluations. Results are shown for the 4,
        dimensional problem when the number of restarts is increased to $5$. Best viewed in color.}
	\label{fig:5_restarts}
\end{figure*}

\begin{figure*}[tbh!]
	\begin{tabular}{cc}
		\includegraphics[width=0.49\textwidth]{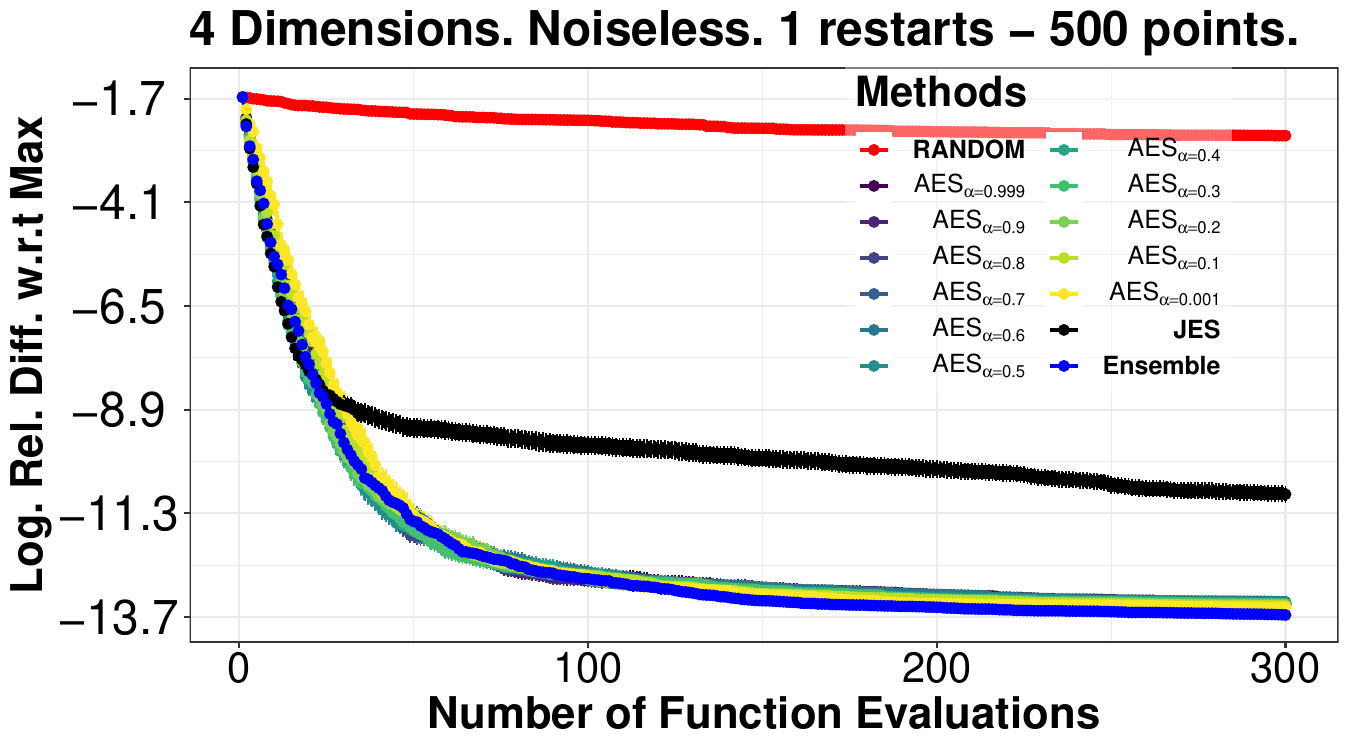}
		\includegraphics[width=0.49\textwidth]{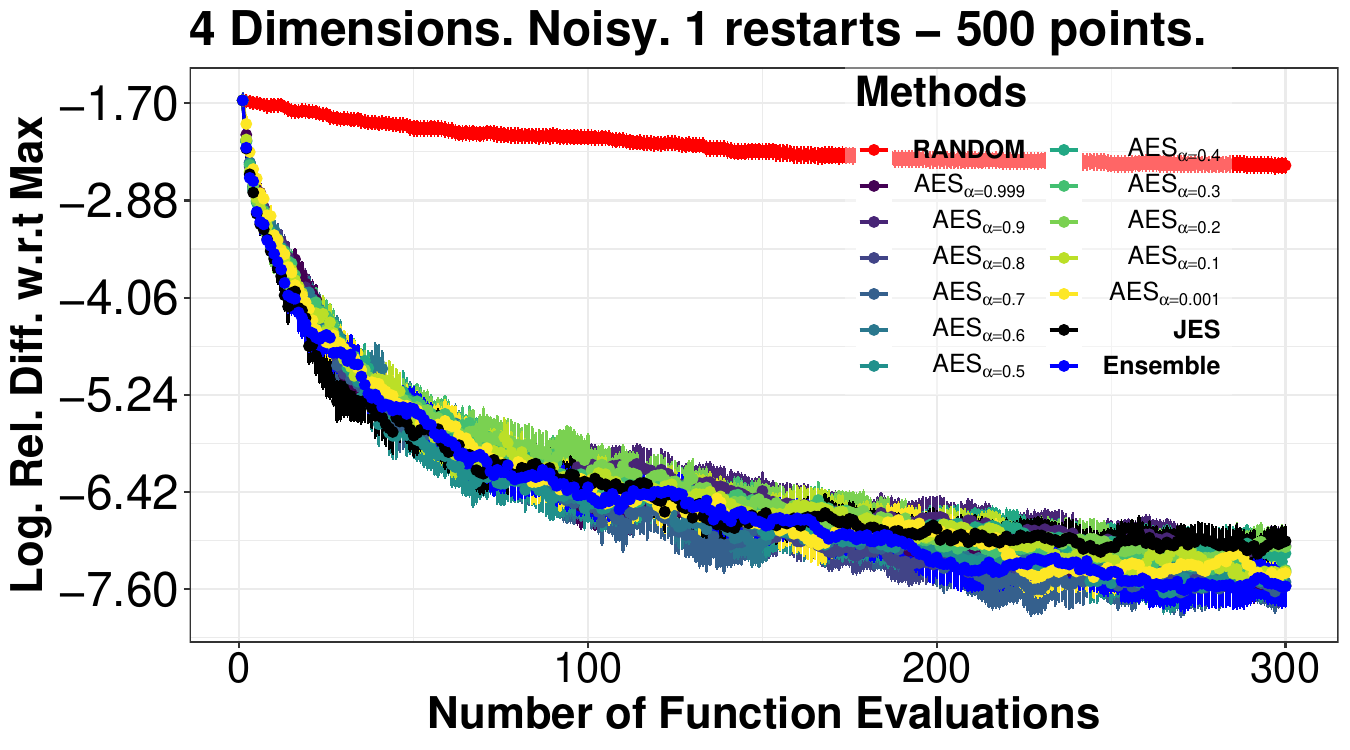}
	\end{tabular}
	\caption{ 
        Average logarithm relative difference between the objective at each method's
        recommendation and the objective at the global maximum, with respect to
        the number of evaluations. Results are shown for the 4,
        dimensional problem when the number of points to choose the starting point of the optimization 
	is increased to $500$. Best viewed in color.}
	\label{fig:500_points}
\end{figure*}

As described in the main document,  to maximize each acquisition function, we used 
L-BFGS-B with $1$ restart and $200$ candidate points, from which the 
starting point of the optimization is chosen. In this setting, BOTorch chooses randomly one 
point to start the optimization, from a random set of $200$ points, favoring the selection of points with high acquisition
\cite{balandat2020botorch}.
Here, we consider a different number of restarts and points to choose the starting 
point of the optimization process. Specifically, we consider increasing the number of 
restarts to $5$ while keeping the number of points equal to $200$. 
In this setting, the acquisition function is optimized $5$ times, from different starting points, chosen
from the initial set of $200$ points. BOTorch favors the selection of $5$ different points with 
high acquisition from the initial set of $200$ points. After optimization, the final point with the best acquisition 
is selected as the next evaluation point. The results obtained, in this setting, for the 4 dimensional synthetic problem, 
are displayed in Figure \ref{fig:5_restarts}, for the noiseless and the noisy evaluation scenario.
We observe that the results obtained are very similar to those 
reported in the main manuscript. Here, we also consider increasing the number of 
points to $500$ while keeping the number of restarts equal to $1$. 
In this setting, the acquisition function is optimized $1$ time, and the starting point is  chosen
from an initial set of $500$ random points. The results obtained, in this setting, for the 4 dimensional synthetic 
problem, are displayed in Figure \ref{fig:500_points}, for the noiseless and the noisy evaluation scenario.
Again, we observe that the results obtained are very similar to those reported in the main manuscript.

\end{document}